%% file: main.tex
\documentclass[10pt,twocolumn,letterpaper]{article}

\usepackage[pagenumbers]{cvpr} %

\newif\ifeccv
\eccvfalse

\usepackage{tikz}
\usetikzlibrary{spy}
\usepackage{multirow}
\usepackage{soul}
\usepackage[linesnumbered, boxed, ruled]{algorithm2e}
\usepackage{floatrow}
\usepackage[toc,page]{appendix}

\definecolor{cvprblue}{rgb}{0.21,0.49,0.74}
\usepackage[pagebackref,breaklinks,colorlinks,citecolor=cvprblue]{hyperref}

\def\paperID{*****} %
\def\confName{CVPR}
\def\confYear{2024}

\title{ReNoise: Real Image Inversion Through Iterative Noising \\[-8pt]}

\author{
        Daniel Garibi$^1$ \hspace{5mm}
        Or Patashnik$^1$ \hspace{5mm}
        Andrey Voynov$^2$ \hspace{5mm}
        Hadar Averbuch-Elor$^1$ \hspace{5mm}
        Daniel Cohen-Or$^1$
        \\[4pt]
        $^1$Tel Aviv University \hspace{10mm} $^2$Google Research
        \\
        \small\url{https://garibida.github.io/ReNoise-Inversion/}
         \\[-25pt]
}

\begin{document}

\input{figures/teaser}
\vspace{-5pt}
\input{0-abstract}
\input{1-intro}
\input{2-related}
\input{4-method}

\input{5-theory}
\input{6-experiments}

\input{7-conclusion}

{
    \small
    \bibliographystyle{ieeenat_fullname}
    \bibliography{main}
}

\clearpage
\begin{appendices}
    \input{8-sup}

\end{appendices}

\end{document}

%% file: figures/teaser.tex
\twocolumn[{%
    \renewcommand\twocolumn[1][]{#1}%
    \maketitle
    \begin{center}
        \includegraphics[width=1.0\textwidth]{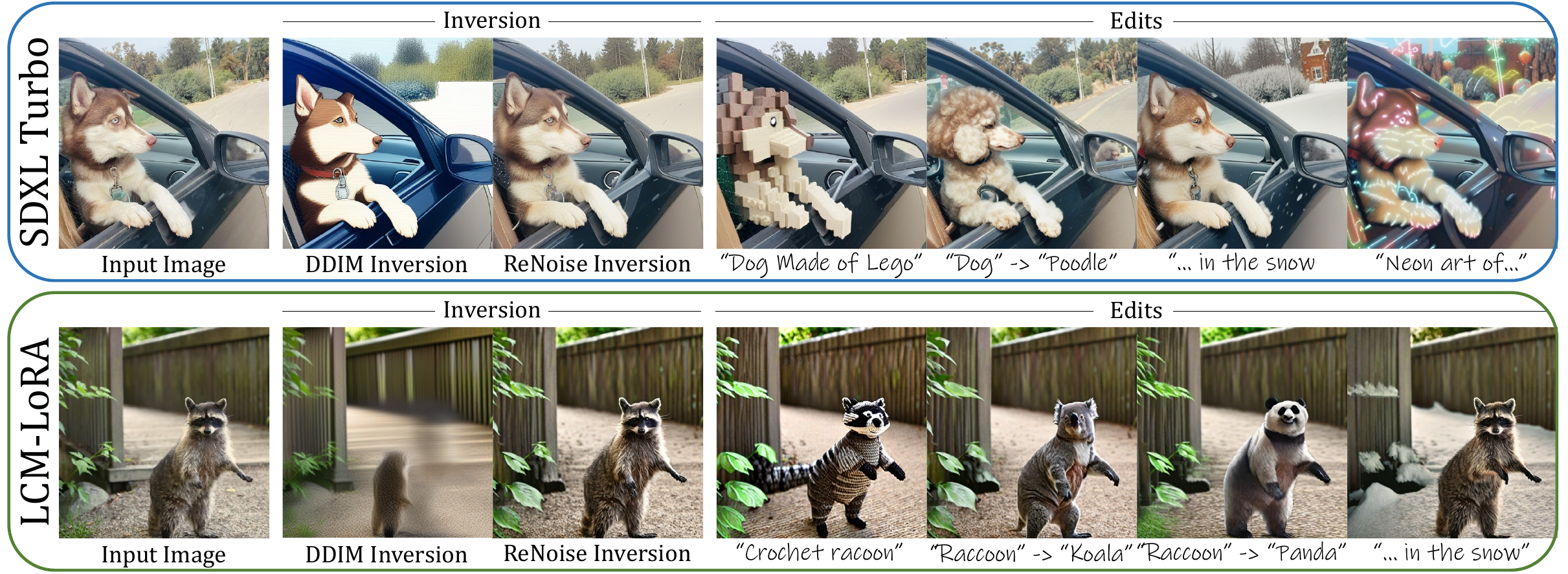}
        \vspace{-18pt}
        \captionof{figure}{
        Our ReNoise inversion technique can be applied to various diffusion models, including recent few-step ones. This figure illustrates the performance of our method with SDXL Turbo and LCM models, showing its effectiveness compared to DDIM inversion. 
        Additionally, we demonstrate that the quality of our inversions allows prompt-driven editing. 
        As illustrated on the right, our approach also allows for prompt-driven image edits.
        \ifeccv
        \else
        \fi
        }
    \vspace{-1pt}
    \label{fig:teaser}
    \end{center}
}] 

%% file: 0-abstract.tex
\ifeccv
\else
\fi

\begin{abstract}
\ifeccv
    \vspace{-10pt}
\else
    \vspace{-13pt}
\fi

Recent advancements in text-guided diffusion models have unlocked powerful image manipulation capabilities.
However, applying these methods to real images necessitates the inversion of the images into the domain of the pretrained diffusion model.
Achieving faithful inversion remains a challenge, particularly for more recent models trained to generate images with a small number of denoising steps.
In this work, we introduce an inversion method with a high quality-to-operation ratio, enhancing reconstruction accuracy without increasing the number of operations.
Building on reversing the diffusion sampling process, our method employs an iterative renoising mechanism at each inversion sampling step.
This mechanism refines the approximation of a predicted point along the forward diffusion trajectory, by iteratively applying the pretrained diffusion model, and averaging these predictions.
We evaluate the performance of our ReNoise technique using various sampling algorithms and models, including recent accelerated diffusion models. Through comprehensive evaluations and comparisons, we show its effectiveness in terms of both accuracy and speed. Furthermore, we confirm that our method preserves editability by demonstrating text-driven image editing on real images.

\ifeccv
    \keywords{Diffusion Models \and Image Editing \and Inversion}
\else
    \vspace{-14pt}
\fi
\end{abstract}

%% file: 1-intro.tex
\section{Introduction}
\ifeccv
    \vspace{-10pt}
\else
    \vspace{-5pt}
\fi

Large-scale text-to-image diffusion models have revolutionized the field of image synthesis~\cite{rombach2022highresolution, saharia2022photorealistic, ramesh2022hierarchical, ho2020denoising}. In particular, many works have shown that these models can be employed for various types of image manipulation~\cite{hertz2022prompttoprompt, parmar2023zeroshot, pnpDiffusion2022, meng2022sdedit, brooks2023instructpix2pix, avrahami2022blended, avrahami2022blended_latent, kawar2023imagic, ge2023expressive, patashnik2023localizing, alaluf2023crossimage}. To edit \emph{real} images, many of these techniques often require the inversion of the image into the domain of the diffusion model. That is, given a real image $z_0$, one has to find a Gaussian noise $z_T$, such that denoising $z_T$ with the pretrained diffusion model reconstructs the given real image $z_0$. The importance of this task for real image manipulation has prompted many efforts aimed at achieving accurate reconstruction ~\cite{mokady2022nulltext, miyake2023negativeprompt, han2023improving, hubermanspiegelglas2023edit}.

\ifeccv
\else
    \input{figures/trajectory_figure}
\fi

The diffusion process consists of a series of denoising steps $\{\epsilon_\theta(z_t, t)\}_{t=T}^1 $, which form a trajectory from the Gaussian noise to the model distribution (see Figure~\ref{fig:trajectory}). Each denoising step is computed by a trained network, typically implemented as a UNet, which predicts $z_{t-1}$ from $z_t$~\cite{ho2020denoising}. 
The output of the model at each step forms a \emph{direction} from $z_t$ to $z_{t-1}$~\cite{song2020score}. 
These steps are not invertible, in the sense that the model was not trained to predict $z_t$ from $z_{t-1}$.
Thus, the problem of inverting a given image is a challenge, and particularly for real images, as they are not necessarily in the model distribution (see Figure~\ref{fig:ddim_with_renoise}). 

In this paper, we present an inversion method with a high quality-to-operation ratio, which achieves superior reconstruction accuracy for the same number of UNet operations.
We build upon the commonly used approach of reversing the diffusion sampling process, which is based on the linearity assumption that the direction from $z_{t}$ to $z_{t+1}$ can be approximated by the negation of the direction from 
$z_t$ to $z_{t-1}$
~\cite{dhariwal2021diffusion, song2022denoising} (see Figure~\ref{fig:trajectory}). 
To enhance this approximation, we employ the fixed-point iteration methodology~\cite{burden2015numerical}. 
Specifically, given $z_t$, we begin by using the common approximation to get an initial estimate for $z_{t+1}$, denoted by $z_{t+1}^{(0)}$. Then, we iteratively \textit{renoise} $z_t$, following the direction implied by 
$z_{t+1}^{(k)}$ to obtain $z_{t+1}^{(k+1)}$.
After repeating this renoising process several times, we apply an averaging on $z_{t+1}^{(k)}$ to form a more accurate direction from $z_t$ to $z_{t+1}$.

We show that this approach enables longer strides along the inversion trajectory while improving image reconstruction. Therefore, our method can also be effective with diffusion models trained to generate images using a small number of denoising steps~\cite{luo2023lcm, sauer2023adversarial}. 
Furthermore, despite the need to repeatedly renoise in each step of the inversion process, the longer strides lead to a more favorable tradeoff of UNet operations for reconstruction quality.

Through extensive experiments, we demonstrate the effectiveness of our method in both image reconstruction and inversion speed. We validate the versatility of our approach across different samplers and models, including recent time-distilled diffusion models (e.g., SDXL-Turbo~\cite{sauer2023adversarial}). Importantly, we demonstrate that the editability of the inversion achieved by our method allows a wide range of text-driven image manipulations (see Figure~\ref{fig:teaser}).

\ifeccv
    \input{figures/mini_pages/intro_fig_eccv}
\else
    \input{figures/ddim_with_renoise}
\fi

%% file: figures/trajectory_figure.tex
\begin{figure}[t]
    \centering
    \input{figures/trajectory_figure_content}
\end{figure}

%% file: figures/trajectory_figure_content.tex
\includegraphics[width=1.0\linewidth]{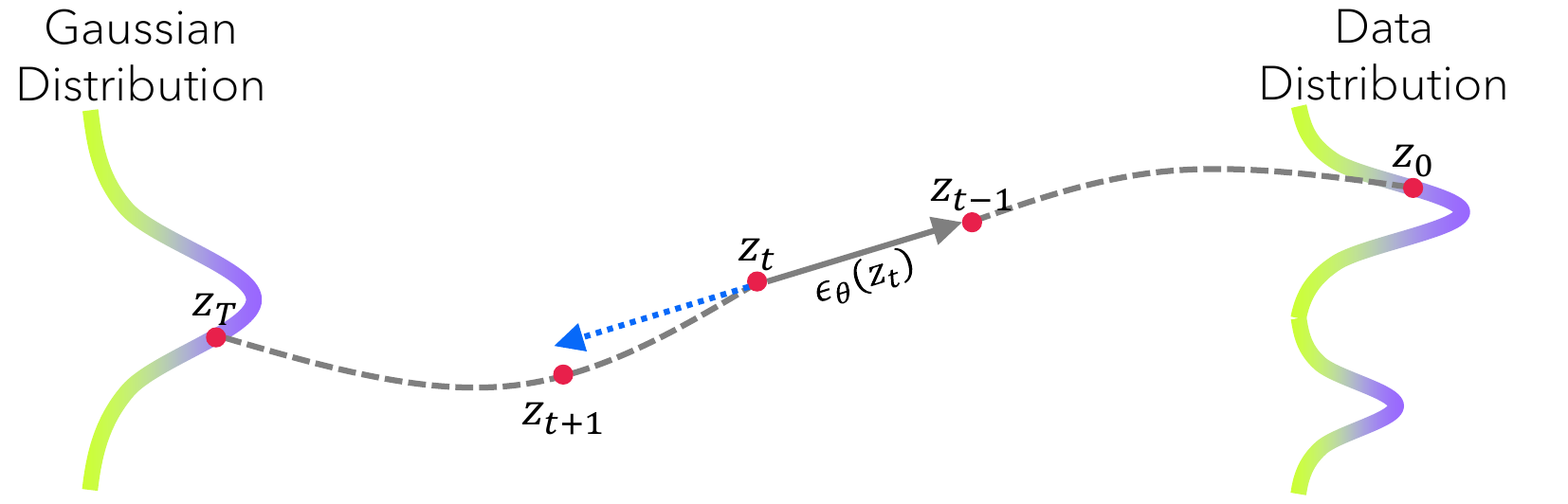}
\vspace{-0.425cm}
\caption{
The diffusion process samples a Gaussian noise and iteratively denoises it until reaching the data distribution. 
At each point along the denoising trajectory, the model predicts a direction, $\epsilon_\theta(z_t)$, to step to the next point along the trajectory. To invert a given image from the distribution, the direction from $z_t$ to $z_{t+1}$ is approximated with the inverse of the direction from $z_t$ to $z_{t-1}$ denoted by a dotted blue line.
    \ifeccv
    \else
        \vspace{-16pt}
    \fi
} 
\label{fig:trajectory}

%% file: figures/mini_pages/intro_fig_eccv.tex
\begin{figure}[t]
    \centering
    \scriptsize
    \begin{minipage}[b]{0.50\textwidth}
        \centering
        \input{figures/trajectory_figure_content}
    \end{minipage}
    \hfill
    \begin{minipage}[b]{0.46\textwidth}
        \centering
        \input{figures/ddim_with_renoise_content}
    \end{minipage}
    \vspace{-13pt}
\end{figure}

%% file: figures/ddim_with_renoise_content.tex
{\small
\centering

\begin{tabular}{c c c}
    \ifeccv
        {\begin{tabular}{c} Original \\ Image  \end{tabular}} &
        {\begin{tabular}{c} DDIM \\ Inversion  \end{tabular}} &
        {\begin{tabular}{c} {+1 ReNoise} \\ Step  \end{tabular}} \\
    \else
        {Original Image} &
        {DDIM Inversion} &
        {+ 1 ReNoise Step} \\
    \fi
    \includegraphics[width=0.33\columnwidth]{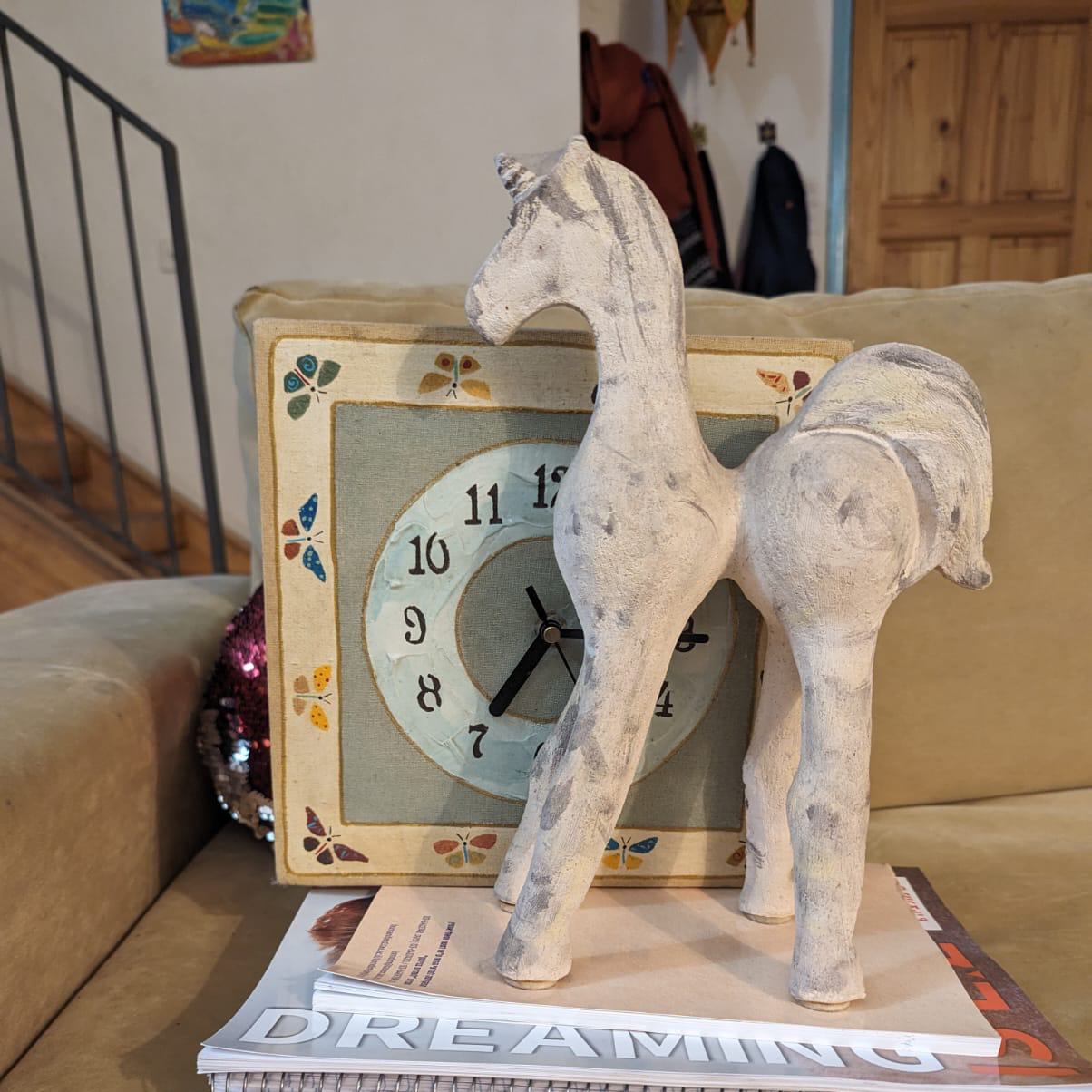} &
    \includegraphics[width=0.33\columnwidth]{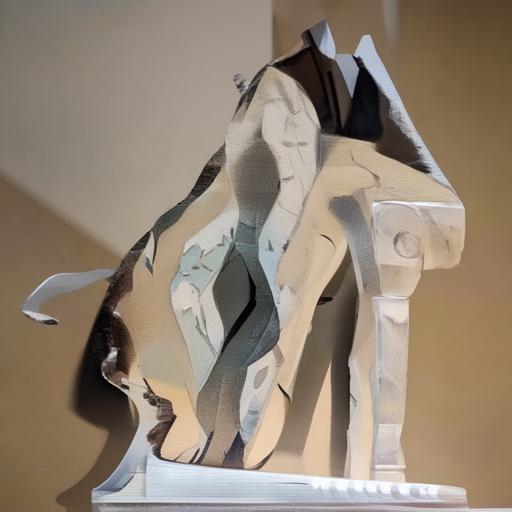} &
    \includegraphics[width=0.33\columnwidth]{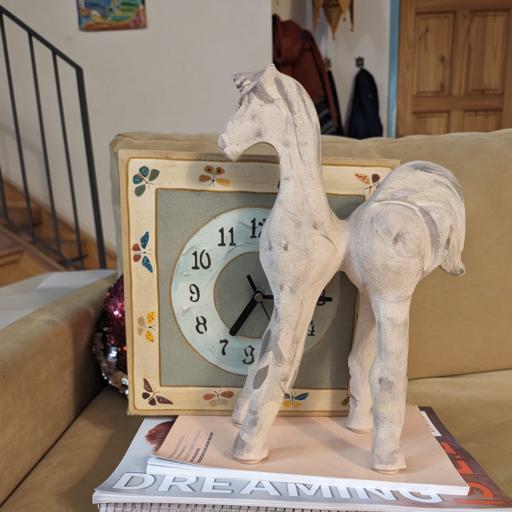} \\
    
    \includegraphics[width=0.33\columnwidth]{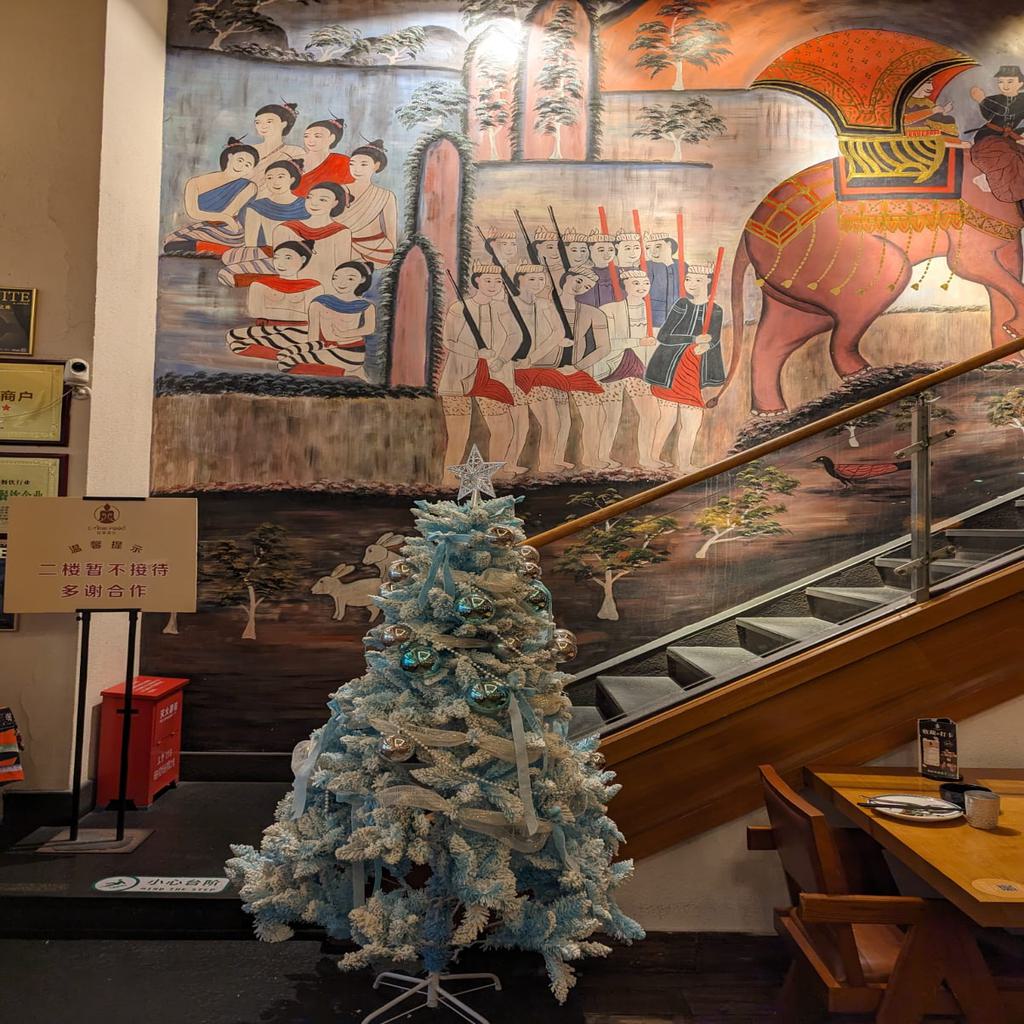} &
    \includegraphics[width=0.33\columnwidth]{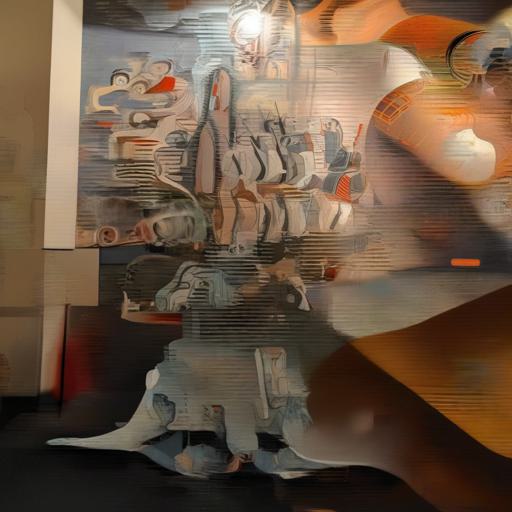} &
    \includegraphics[width=0.33\columnwidth]{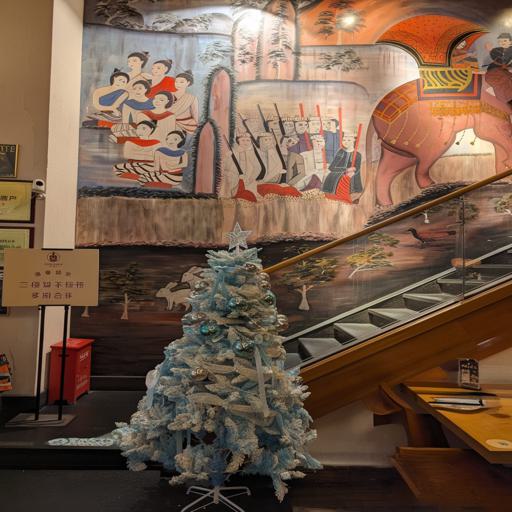} \\
    
\end{tabular}

}
\vspace{-0.325cm}
\caption{
    \ifeccv
        Comparing reconstruction results of plain DDIM Inv. with SDXL to DDIM Inv. with one renoising iteration.
    \else
        Comparing reconstruction results of plain DDIM inversion (middle column) on SDXL to DDIM inversion with one ReNoise iteration (rightmost column).
        \vspace{-15pt}
    \fi
}
\label{fig:ddim_with_renoise}

%% file: figures/ddim_with_renoise.tex
\begin{figure}
    \centering
    \setlength{\tabcolsep}{1pt}
    \addtolength{\belowcaptionskip}{-10pt}

\input{figures/ddim_with_renoise_content}
\end{figure}

%% file: 2-related.tex
\ifeccv
    \vspace{-0pt}
\else
    \vspace{-4pt}
    \input{figures/overview_figure}
\fi

\section{Related Work}
\ifeccv
    \vspace{-6pt}
\fi

\paragraph{\textbf{Image Editing via Diffusion Models}}
Recent advancements in diffusion models \cite{ho2020denoising, dhariwal2021diffusion} have resulted in unprecedented diversity and fidelity in visual content creation guided by free-form text prompts \cite{rombach2022highresolution, podell2023sdxl, ramesh2022hierarchical, saharia2022photorealistic}.
Text-to-image models do not directly support text-guided image editing. Therefore, harnessing the power of these models for image editing is a significant research area and many methods have utilized these models for different types of image editing~\cite{meng2022sdedit, kawar2023imagic, hertz2022prompttoprompt, pnpDiffusion2022, brooks2023instructpix2pix, patashnik2023localizing, parmar2023zeroshot, ge2023expressive, alaluf2023crossimage, hertz2023StyleAligned, epstein2023selfguidance, voynov2023anylens, avrahami2022blended, avrahami2022blended_latent, Couairon2022DiffEditDS}.
A common approach among these methods requires \emph{inversion}~\cite{song2022denoising, mokady2022nulltext, hubermanspiegelglas2023edit, wallace2022edict} to edit real images, i.e., obtaining a latent code $z_T$ such that denoising it with the pretrained diffusion model returns the original image.
Specifically, in this approach two backward processes are done simultaneously using $z_T$. One of the processes reconstructs the image using the original prompt, while the second one injects features from the first process (e.g., attention maps) to preserve some properties of the original image while manipulating other aspects of it.

\ifeccv
\vspace{-10pt}
\else
\vspace{-10pt}
\fi

\paragraph{\textbf{Inversion in Diffusion Models}}

Initial efforts in image inversion for real image editing focused on GANs~\cite{zhu2016generative, zhu2020improved, zhu2020indomain, Abdal_2019, Abdal_2020, richardson2021encoding, Alaluf_2021, tov2021encoder, alaluf2021hyperstyle, dinh2021hyperinverter, Parmar_2022}.
The advancements in diffusion models, and in diffusion-based image editing in particular have recently prompted works studying the inversion of a diffusion-based denoising process. This inversion depends on the sampler algorithm used during inference, which can be deterministic~\cite{song2022denoising} or non-deterministic~\cite{ho2020denoising, karras2022elucidating}. Inversion methods can be accordingly categorized into two: methods that are suitable for deterministic sampling, and methods suitable for non-deterministic sampling.  

Methods that approach the deterministic inversion commonly rely on the DDIM sampling method~\cite{song2022denoising}, and build upon DDIM inversion~\cite{dhariwal2021diffusion, song2022denoising}. 
Mokady et al.~\cite{mokady2022nulltext} observed that the use of classifier-free guidance during inference magnifies the accumulated error of DDIM inversion and therefore leads to poor reconstruction. Following this observation, several works
~\cite{mokady2022nulltext, miyake2023negativeprompt, han2023improving} focused on solving this issue by replacing the null text token with a different embedding, which is found after an optimization process or by a closed solution. 
However, excluding~\cite{mokady2022nulltext} which requires a lengthy optimization, these methods are limited by the reconstruction accuracy of DDIM inversion, which can be poor, especially when a small number of denoising steps is done. In our work, we present a method that improves the reconstruction quality of DDIM inversion and therefore can be integrated with methods that build on it. 

Another line of work~\cite{hubermanspiegelglas2023edit, wu2022unifying} tackles the inversion of DDPM sampler~\cite{ho2020denoising}. In these works~\cite{wu2022unifying, hubermanspiegelglas2023edit}, instead of inverting the image into an initial noise $z_T$, a series of noises $\{z_T, \epsilon_T, ..., \epsilon_1\}$ is obtained. The definition of this noises series ensures that generating an image with it returns the original input image. However, these methods require a large number of inversion and denoising steps to allow image editing. Applying these methods with an insufficient number of steps leads to too much information encoded in $\{ \epsilon_T, ..., \epsilon_1 \}$ which limits the ability to edit the generated image. As shall be shown, The editability issue of these methods is particularly evident in few-steps models~\cite{luo2023lcm, luo2023latent, sauer2023adversarial}.

Most relevant to our work, two recent inversion methods~\cite{Pan_2023_ICCV, meiri2023fixedpoint} also use the fixed-point iteration technique. Specifically, they improve the reconstruction accuracy of DDIM inversion~\cite{song2022denoising} with Stable Diffusion~\cite{rombach2022highresolution} without introducing a significant computational overhead. 
In our work, we focus on the problem of real image inversion for recently introduced few-step diffusion models, where the difficulties encountered by previous methods are amplified. Furthermore, we show that our inversion method successfully works with various models and different samplers.

\ifeccv
    \vspace{-7pt}
\else
    \vspace{-10pt}
\fi

\paragraph{\textbf{Few Steps Models}}
Recently, new methods~\cite{sauer2023adversarial, luo2023latent, luo2023lcm, song2023consistency, salimans2021progressive} that fine-tune text-to-image diffusion models enabled a significant reduction of the number of steps needed for high-quality image generation. While standard diffusion models typically require 50 denoising steps to generate high-quality images, recent accelerated models achieve high-quality synthesis with 1-4 steps only. These new methods pave the way for interactive editing workflows. 
However, as we show in this paper, using current methods for the inversion of an image with a small number of steps degrades the reconstruction quality in terms of accuracy~\cite{song2022denoising, dhariwal2021diffusion} or editability~\cite{hubermanspiegelglas2023edit, wu2022unifying}.

%% file: figures/overview_figure.tex
\begin{figure*} [t]
    \centering
    \includegraphics[width=1.0\linewidth]{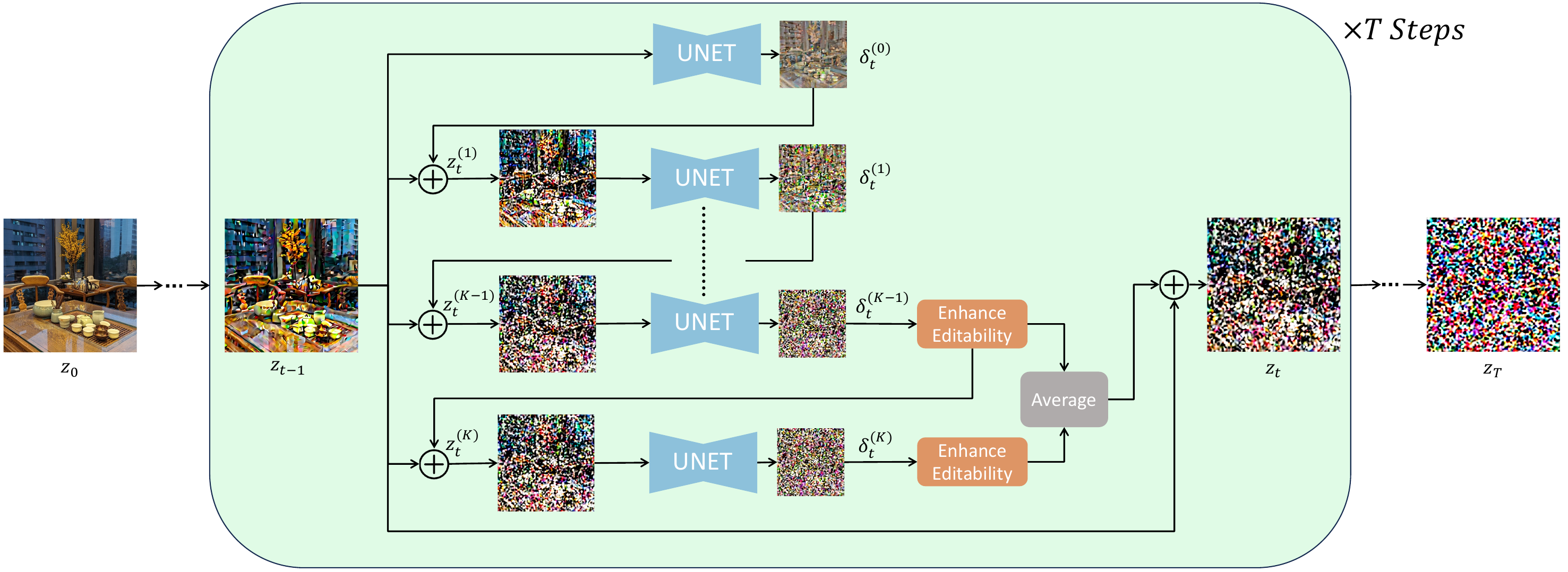}
    \ifeccv
        \vspace{-10pt}
    \else
    \fi
    \caption{
    Method overview. Given an image $z_0$, we iteratively compute $z_1, ..., z_T$, where each $z_t$ is calculated from $z_{t-1}$.
    At each time step, we apply the UNet ($\epsilon_\theta$) $\mathcal{K}+1$ times, each using a better approximation of $z_t$ as the input. The initial approximation is $z_{t-1}$. The next one, $z_t^{(1)}$, is the result of the reversed sampler step (i.e., DDIM). The reversed step begins at $z_{t-1}$ and follows the direction of $\epsilon_\theta(z_{t-1}, t)$.
    At the $k$ renoising iteration, $z_t^{(k)}$ is the input to the UNet, and we obtain a better $z_t$ approximation.
    For the lasts iterations, we optimize $\epsilon_\theta(z_{t}^{(k)}, t)$ to increase editability.
    As the final denoising direction, we use the average of the UNet predictions of the last few iterations.
    \ifeccv
    \else
        \vspace{-12pt}
    \fi
    }
    \label{fig:overview}
    \ifeccv
        \vspace{-15pt}
    \else
        \vspace{-7pt}
    \fi
\end{figure*}

%% file: 4-method.tex
\ifeccv
    \vspace{-12pt}
\else
\fi

\section{Method}

\ifeccv
    \vspace{-5pt}
\fi

\subsection{ReNoise Inversion}
\label{sec:renoise}

\ifeccv
\else
    \input{figures/graphic_renoising_figure}
\fi

\paragraph{\textbf{Reversing the Sampler}}
Samplers play a critical role in the diffusion-based image synthesis process. They define the noising and denoising diffusion processes and influence the processes' trajectories and quality of the generated images. While different samplers share the same pre-trained UNet model (denoted by $\epsilon_\theta$) as their backbone, their sampling approaches diverge, leading to nuanced differences in output.
The goal of the denoising sampler is to predict the latent code at the previous noise level, $z_{t-1}$, based on the current noisy data $z_t$, the pretrained UNet model, and a sampled noise, $\epsilon_t$.
Various denoising sampling algorithms adhere to the form:
\begin{equation} \label{eq:sampler}
    z_{t-1} = \phi_t z_t + \psi_t \epsilon_\theta(z_t, t, c) + \rho_t \epsilon_t,
\end{equation}
where $c$ represents a text embedding condition, and $\phi_t$, $\psi_t$, and $\rho_t$ denote sampler parameters.
At each step, these parameters control the extent to which the previous noise is removed ($\phi_t$), the significance assigned to the predicted noise from the UNet ($\psi_t$), and the given weight to the
additional noise introduced ($\rho_t$).

A given image $z_0$ can be inverted by reformulating Equation~\ref{eq:sampler} and applying it iteratively:
\begin{equation} \label{eq:reverse_eq}
z_{t} = \frac{z_{t-1} - \psi_t \epsilon_\theta(z_t, t, c) - \rho_t \epsilon_t}{\phi_t},
\end{equation}
where for non-deterministic samplers, a series of random noises $\{\epsilon_t\}_{t=1}^T$ is sampled and used during both inversion and image generation processes. 
However, directly computing $z_t$ from Equation~\ref{eq:reverse_eq} is infeasible since it relies on $\epsilon_\theta(z_t, t, c)$, which, in turn, depends on $z_t$, creating a circular dependency.
To solve this implicit function,
Dhariwal et al.~\cite{dhariwal2021diffusion} propose using the approximation $\epsilon_\theta(z_t, t, c) \approx \epsilon_\theta(z_{t-1}, t, c)$:
\begin{equation} \label{eq:reverse_eq_aprox}
z^{(1)}_{t} = \frac{z_{t-1} - \psi_t \epsilon_\theta(z_{t-1}, t, c) - \rho_t \epsilon_t}{\phi_t}.
\end{equation}
This method has several limitations. First, the assumption underlying the approximation used in \cite{dhariwal2021diffusion} is that the number of inversion steps is large enough, implying a trajectory close to linear. This assumption restricts the applicability of this inversion method in interactive image editing with recent few-step diffusion models~\cite{song2023consistency, luo2023latent, luo2023lcm, sauer2023adversarial}, as the inversion process would take significantly longer than inference. Second, this method struggles to produce accurate reconstructions in certain cases, such as highly detailed images or images with large smooth regions, see Figure~\ref{fig:ddim_with_renoise}. Moreover, we observe that this inversion method is sensitive to the prompt $c$ and may yield poor results for certain prompts.

\ifeccv

\input{figures/overview_figure}\input{figures/graphic_renoising_figure}
\fi

\paragraph{\textbf{ReNoise}}
In a successful inversion trajectory, the direction from $z_{t-1}$ to $z_t$ aligns with the direction from $z_t$ to $z_{t-1}$ in the denoising trajectory.
To achieve this, we aim to improve the approximation of $\epsilon_\theta(z_t, t, c)$ in Eq.~\ref{eq:reverse_eq} compared to the one used in \cite{dhariwal2021diffusion}.
Building on the fixed-point iteration technique~\cite{burden2015numerical}, our approach 
better estimates the instance of $z_t$ that is inputted to the UNet, rather than relying on $z_{t-1}$.

Intuitively, we utilize the observation that $z_t^{(1)}$ (from Eq.~\ref{eq:reverse_eq_aprox}) offers a more precise estimate of $z_t$ compared to $z_{t-1}$. 
Therefore, employing $z_t^{(1)}$ as the input to the UNet is likely to yield a more accurate direction, thus contributing to reducing the overall error in the inversion step. We illustrate this observation in Figure~\ref{fig:graphic_renoising_figure}. 
Iterating this process generates a series of estimations for $z_t$, denoted by $\{z_t^{(k)}\}_{k=1}^{\mathcal{K}+1}$. 
While the fixed-point iteration technique~\cite{burden2015numerical} does not guarantee convergence of this series in the general case, in Section~\ref{sec:theory}, we empirically show that convergence holds in our setting.
However, as the convergence is not monotonic, we refine our prediction of $z_t$ by averaging several $\{z_t^{(k)}\}$, thus considering more than a single estimation of $z_t$. See Figure~\ref{fig:convergence_scheme} for an intuitive illustration.

In more detail, our method iteratively computes estimations of $z_t$ during each inversion step $t$ by renoising the noisy latent $z_{t-1}$ multiple times, each with a different noise prediction (see Figure~\ref{fig:overview}).
Beginning with $z_t^{(1)}$, in the $k$-th renoising iteration, the input to the UNet is the result of the previous iteration, $z_t^{(k)}$. Then, $z_t^{(k+1)}$ is calculated using the inverted sampler while maintaining $z_{t-1}$ as the starting point of the step.
After $\mathcal{K}$ renoising iterations, we obtain a set of estimations $\{z_t^{(k)}\}_{k=1}^{\mathcal{K}+1}$. The next point on the inversion trajectory, $z_t$, is then defined as their weighted average, where $w_k$ is the weight assigned to $z_t^{(k)}$.
For a detailed description of our method, refer to Algorithm~\ref{alg:algo}.

\ifeccv
\vspace{-10pt}
\else
\input{figures/non_monotonic_figure}
\fi

\subsection{Reconstruction-Editability Tradeoff}

\paragraph{\textbf{Enhance Editability}}

The goal of inversion is to facilitate the editing of real images using a pretrained image generation model.
While the the renoising approach attains highly accurate reconstruction results, we observe that the resulting $z_T$ lacks editability. This phenomenon can be attributed to the reconstruction-editability tradeoff in image generative models~\cite{tov2021encoder}. To address this limitation, we incorporate a technique to enhance the editability of our method.

It has been shown~\cite{parmar2023zeroshot} that the noise maps predicted during the inversion process often diverge from the statistical properties of uncorrelated Gaussian white noise, thereby affecting editability.
To tackle this challenge, we follow pix2pix-zero~\cite{parmar2023zeroshot} and regularize the predicted noise at each step, $\epsilon_\theta(z_t, t, c)$, using the following loss terms.

First, we encourage $\epsilon_\theta(z_t, t, c)$ to follow the same distribution as $\epsilon_\theta(z_t', t, c)$, where $z_t'$ represents the input image $z_0$ with added random noise corresponding to the noise level at timestep $t$. We do so by dividing $\epsilon_\theta(z_t, t, c)$ and $\epsilon_\theta(z_t', t, c)$ into small patches (e.g., 4$\times$4), and computing the KL-divergence between corresponding patches. We denote this loss term by $\mathcal{L}_{\text{patch-KL}}$.
Second, we utilize $\mathcal{L}_{\text{pair}}$ proposed in pix2pix-zero~\cite{parmar2023zeroshot}, which penalizes correlations between pairs of pixels. We leverage these losses to enhance the editability of our method, and denote the combination of them as $\mathcal{L}_{\text{edit}}$. 
For any renoising iteration $k$ where $w_k > 0$, we regularize the UNet's prediction $\epsilon_\theta(z_t^{(k)}, t, c)$ using $\mathcal{L}_{\text{edit}}$ before computing $z_t^{(k+1)}$. See line 9 in Algorithm~\ref{alg:algo}.

\ifeccv
    \input{figures/algo}
\else
    \input{figures/algo}
\fi

\ifeccv
\else
\fi

\vspace{-5pt}
\paragraph{\textbf{Noise Correction in Non-deterministic Samplers}}
Non-deterministic samplers, in which $\rho_t > 0$, introduce noise ($\epsilon_t$) at each denoising step.
Previous methods~\cite{wu2022unifying, hubermanspiegelglas2023edit} suggested using $\epsilon_t$ to bridge the gap between the inversion and denoising trajectories in DDPM inversion.
Specifically, given a pair of points $z_{t-1}, z_t$ on the inversion trajectory, we denote by $\hat{z}_{t-1}$ the point obtained by denoising $z_t$. Ideally, $z_{t-1}$ and $\hat{z}_{t-1}$ should be identical. We define:
\begin{equation} \label{eq:update_e_t}
\epsilon_t = \frac{1}{\rho_t}(z_{t-1} - \phi_t z_{t} - \psi_t \epsilon_\theta(z_t, t, c)).
\end{equation}
Integrating this definition into Eq.~\ref{eq:sampler} yields $\hat{z}_{t-1} = z_{t-1}$.
However, we found that replacing $\epsilon_t$ with the above definition affects editability. Instead, we suggest a more tender approach, optimizing $\epsilon_t$ based on Eq.~\ref{eq:update_e_t} as our guiding objective:
\begin{equation} \label{eq:update_e_t_tender}
\epsilon_t = \epsilon_t - \nabla_{\epsilon_t} \frac{1}{\rho_t} (z_{t-1} - \phi_t z_{t} - \psi_t \epsilon_\theta(z_t, t, c)).
\end{equation}
This optimization improves the reconstruction fidelity while preserving the distribution of the noisy-latents.

%% file: figures/graphic_renoising_figure.tex
\ifeccv
    \begin{figure}
      \begin{minipage}[c]{0.55\textwidth}
        \caption{
           Geometric intuition for ReNoise. At each inversion step, we estimate $z_t$ (marked with a red star) based on $z_{t-1}$. The straightforward approach is to use the negated direction of the denoising step from $z_{t-1}$, assuming the trajectory is approximately linear. However, this assumption is inaccurate, especially in few-step models, where the size of the steps is large. We use the linearity assumption as an initial estimation and keep improving this estimation.
           We recalculate the denoising step from the previous estimation (which is closer to the target $z_t$) and then proceed with its negated direction from $z_{t-1}$ (see the orange vectors).
        } \label{fig:graphic_renoising_figure}
      \end{minipage}
      \hfill
      \begin{minipage}[c]{0.45\textwidth}
        \includegraphics[width=\textwidth]{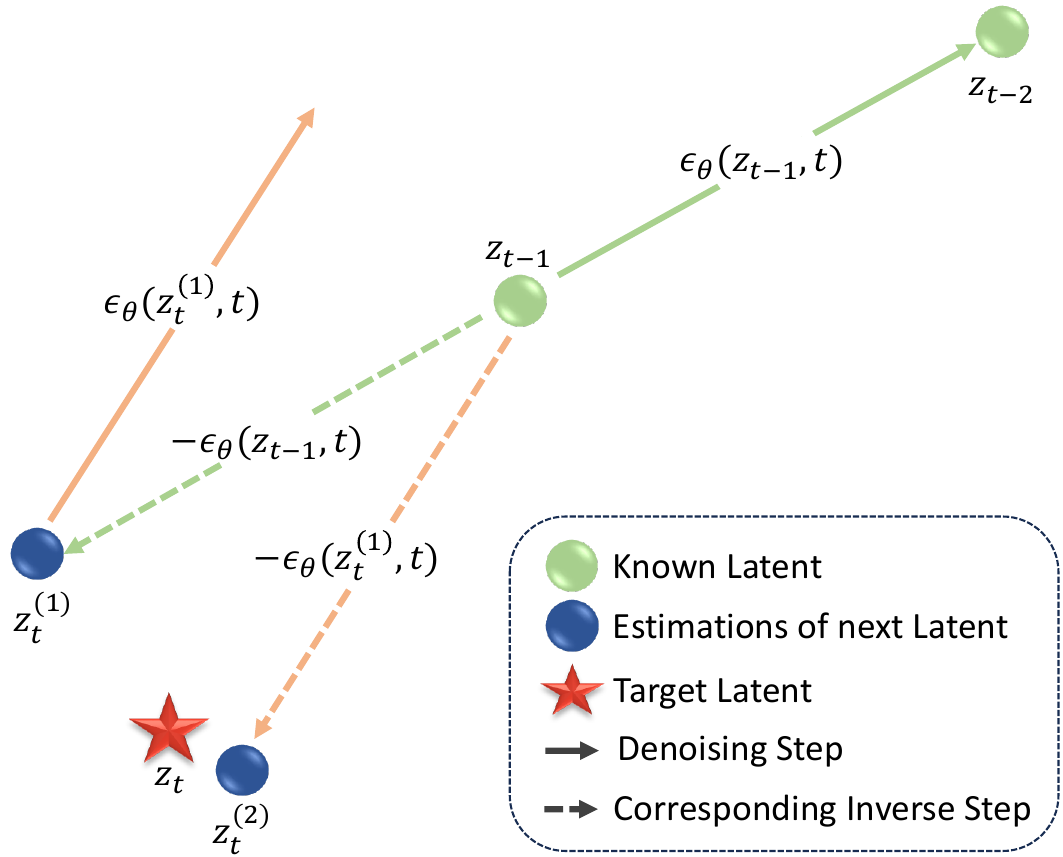}
      \end{minipage}
      \ifeccv
        \vspace{-30pt}
      \fi
    \end{figure}    
\else
    \begin{figure}
        \centering
        \includegraphics[width=0.9\linewidth]{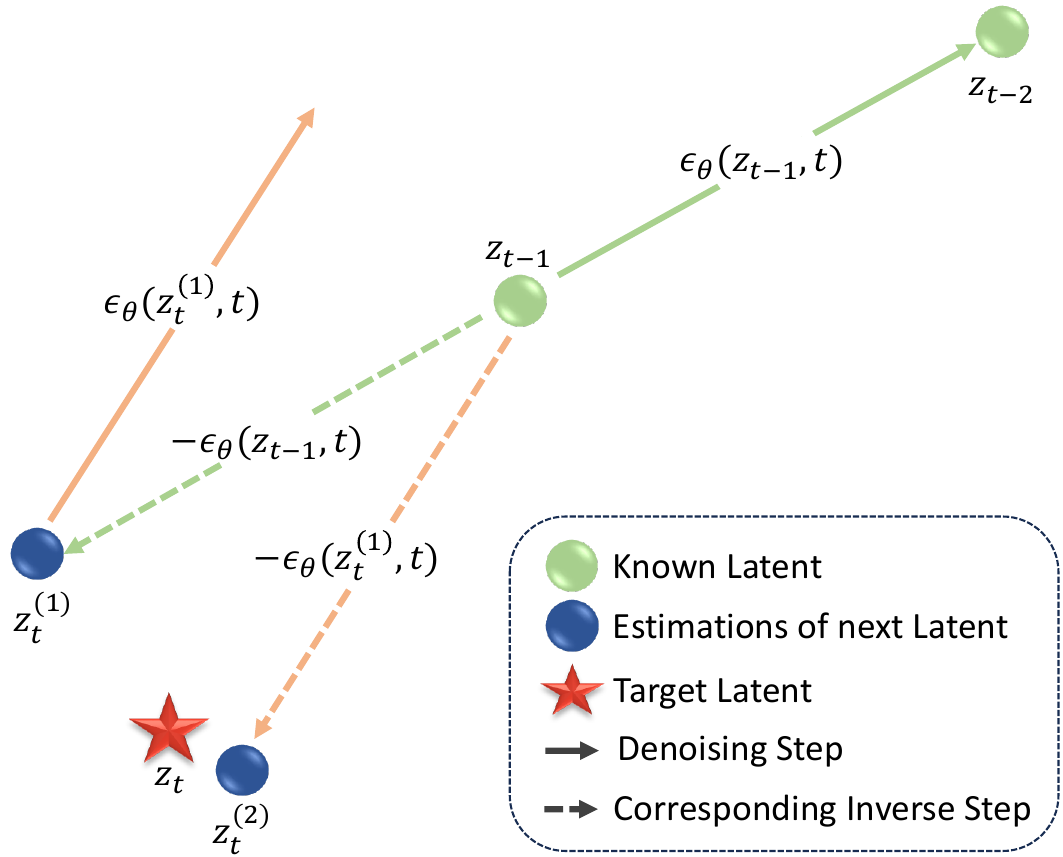}
        \vspace{-5pt}
        \caption{
        Geometric intuition for ReNoise. At each inversion step, we are trying to estimate $z_t$ (marked with a red star) based on $z_{t-1}$. The straightforward approach is to use the reverse direction of the denoising step from $z_{t-1}$, assuming the trajectory is approximately linear. However, this assumption is inaccurate, especially in few-step models, where the size of the steps is not small. We use the linearity assumption only as an initial estimation and keep improving the estimation. We recalculate the denoising step from the previous estimation (which is closer to $z_t$) and then proceed with its opposite direction from $z_{t-1}$ (see the orange vectors).
        \ifeccv
        \else
            \vspace{-16pt}
        \fi
        }
        \label{fig:graphic_renoising_figure}
    \end{figure}
\fi

%% file: figures/non_monotonic_figure.tex
\begin{figure}
    \centering
    \input{figures/non_monotonic_figure_content}
\end{figure}

%% file: figures/non_monotonic_figure_content.tex
\includegraphics[width=0.8\columnwidth]{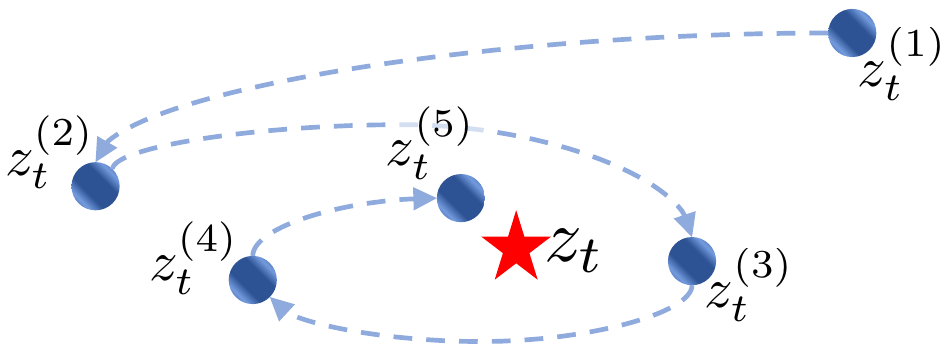}
\vspace{-7pt}
\caption{
Schematic illustration of the ReNoise convergence process to the true inversion of $z_{t-1}$.
While estimates may converge non-monotonically to the unknown target $z_t$, we found that averaging them improves true value estimation. Typically, the initial iteration exhibits an exponential decrease in the norm between consecutive elements.
        \ifeccv
        \else
            \vspace{-17pt}
        \fi
}
\label{fig:convergence_scheme}

%% file: figures/algo.tex
\ifeccv
    \vspace{-15pt}
    \begin{algorithm}
        \caption{ReNoise Inversion}
        \label{alg:algo}
        \textbf{Input:} An image $z_0$, number of renoising steps $\mathcal{K}$, number of inversion steps $T$, a series of renoising weights $\{w_k\}_{k=1}^{\mathcal{K}+1}$.\\
        \textbf{Output:} A noisy latent $z_T$ and set of noises $\{\epsilon_t\}_{t=1}^T$.\\
        \vspace{5pt}
        \begin{multicols}{2}
            \For{$t=0,1,\ldots,T$}{
                \text{sample} $\epsilon_t \sim \mathcal{N}(0,I)$ \\
                $z_{t}^{(0)} \gets z_{t-1}$ \\
                $z_t^{(\text{avg})} \gets 0$ \\
                \For{$k=0,\ldots,\mathcal{K}$} {
                    $\delta_t^{k} \gets \epsilon_\theta(z_{t}^{(k)}, t)$ \\
                    $\delta_t^{k} \gets \texttt{Enhance-edit}(\delta_t^{k}, w_{k+1})$ \\
                    $z_{t}^{(k+1)} \gets \texttt{Inverse-Step}(z_{t-1}, \delta_t^{k})$ \\
                 }
                \tcp{Average ReNoised predictions}
                $z_t^{(\text{avg})} \gets \sum_{k=1}^{\mathcal{K}+1}{w_k \cdot z_t^{(k)}}$ \\
                $\epsilon_t \gets \texttt{Noise-Corr} (z_{t}^{(\text{avg})}, t, \epsilon_t, z_{t-1}) $ \\
             }
            \Return{$(z_{T},\{\epsilon_t\}_{t=1}^T)$}
            \texttt{\\}
            \texttt{\\}
            \SetKwFunction{FMain}{Inverse-Step}
              \SetKwProg{Fn}{Function}{:}{}
              \Fn{\FMain{$z_{t-1}$, $\delta_t$, $t$}}{
                    \KwRet\ ${\frac{1}{\phi_t}z_{t-1} - \frac{\psi_t}{\phi_t} \delta_t - \frac{\rho_t}{\phi_t} \epsilon_t}$
              }
            \texttt{\\}
            \tcp{Enhance-editability}
            \SetKwFunction{FMain}{Enhance-edit}
              \SetKwProg{Fn}{Function}{:}{}
              \Fn{\FMain{$\delta_t^{k}$, $w_{k+1}$}}{
                    \If{$w_{k+1} > 0$} {
                        $\delta_t^k \gets \delta_t^k -\nabla_{\delta_t^k} \mathcal{L}_{\text{edit}}(\delta_t^k) $ \\
                    }
                    \KwRet\ $\delta_t^k$
              }
            \texttt{\\}
            \tcp{Noise Correction}
            \SetKwFunction{FMain}{Noise-Corr}
              \SetKwProg{Fn}{Function}{:}{}
              \Fn{\FMain{$z_t$, $t$, $\epsilon_t$, $z_{t-1}$}}{
                  $\delta_t \gets \epsilon_\theta(z_{t}, t)$ \\
                  $\epsilon_t \gets \epsilon_t -\nabla_{\epsilon_t} \frac{1}{\rho_t} (z_{t-1} - \phi_t z_{t} - \psi_t \delta_t) $ \\
                    \KwRet\ $\epsilon_t$
              }
        \end{multicols}
    \end{algorithm}
    \vspace{-15pt}
\else
    \begin{algorithm}[h] 
    \caption{ReNoise Inversion}
    \label{alg:algo}
    \textbf{Input:} An image $z_0$, number of renoising steps $\mathcal{K}$, number of inversion steps $T$, a series of renoising weights $\{w_k\}_{k=1}^{\mathcal{K}}$.\\
    \textbf{Output:} A noisy latent $z_T$ and set of noises $\{\epsilon_t\}_{t=1}^T$.\\
    
     \For{$t=1,2,\ldots,T$}{
        \text{sample} $\epsilon_t \sim \mathcal{N}(0,I)$ \\
        $z_{t}^{(0)} \gets z_{t-1}$ \\
        $z_t^{(\text{avg})} \gets 0$ \\
        \For{$k=1,\ldots,\mathcal{K}$} {
            $\delta_t^{k} \gets \epsilon_\theta(z_{t}^{(k-1)}, t)$ \\
            $\delta_t^{k} \gets \texttt{Enhance-editability}(\delta_t^{k}, w_k)$ \\
            $z_{t}^{(k)} \gets \texttt{Inverse-Step}(z_{t-1}, \delta_t^{k})$ \\
         }
        \tcp{Average ReNoised predictions}
        $z_t^{(\text{avg})} \gets \sum_{k=1}^{\mathcal{K}}{w_k \cdot z_t^{(k)}}$ \\
        $ \epsilon_t \gets \texttt{Noise-Correction} (z_{t}^{(\text{avg})}, t, \epsilon_t, z_{t-1}) $ \\
     }
    \Return{$(z_{T},\{\epsilon_t\}_{t=1}^T)$}
    
    \texttt{\\}
    \SetKwFunction{FMain}{Inverse-Step}
      \SetKwProg{Fn}{Function}{:}{}
      \Fn{\FMain{$z_{t-1}$, $\delta_t$, $t$}}{
            \KwRet\ ${\frac{1}{\phi_t}z_{t-1} - \frac{\psi_t}{\phi_t} \delta_t - \frac{\rho_t}{\phi_t} \epsilon_t}$
      }
    \texttt{\\}
    \SetKwFunction{FMain}{Enhance-editability}
      \SetKwProg{Fn}{Function}{:}{}
      \Fn{\FMain{$\delta_t^{(k)}$, $w_k$}}{
            \If{$w_k > 0$} {
                $\delta_t^k \gets \delta_t^k -\nabla_{\delta_t^k} \mathcal{L}_{\text{edit}}(\delta_t^k) $ \\
            }
            \KwRet\ $\delta_t^k$
      }
    \texttt{\\}
    \SetKwFunction{FMain}{Noise-Correction}
      \SetKwProg{Fn}{Function}{:}{}
      \Fn{\FMain{$z_t$, $t$, $\epsilon_t$, $z_{t-1}$}}{
          $\delta_t \gets \epsilon_\theta(z_{t}, t)$ \\
          $\epsilon_t \gets \epsilon_t -\nabla_{\epsilon_t} \frac{1}{\rho_t} (z_{t-1} - \phi_t z_{t} - \psi_t \delta_t) $ \\
            \KwRet\ $\epsilon_t$
      }
    \end{algorithm}
\fi

%% file: 5-theory.tex
\ifeccv
    \vspace{-5pt}
\fi

\section{Convergence Discussion} \label{sec:theory}

In this section, we first express the inversion process as a backward Euler process and our renoising iterations as fixed-point iterations. While these iterations do not converge in the general case, we present a toy example where they yield accurate inversions. Then, we analyze the convergence of the renoising iterations in our real-image inversion scenario and empirically verify our method's convergence.

\ifeccv
    \vspace{-7pt}
\fi

\paragraph{Inversion Process as Backward Euler}

The denoising process of diffusion models can be mathematically described as solving an ordinary differential equation (ODE). A common method for solving such equations is the Euler method, which takes small steps to approximate the solution.
For ODE in the form of $y'(t) = f(t, y(t))$, Euler solution is defined as:
\begin{equation*}
    y_{n+1} = y_n + h\cdot f(t_n, y_n),    
\end{equation*}
where $h$ is the step size.
The inversion process can be described as solving ODE using the backward Euler method (or implicit Euler method)~\cite{doi:https://doi.org/10.1002/0470868279.ch2}.
This method is similar to forward Euler, with the difference that $y_{n+1}$ appears on both sides of the equation:
\begin{equation*}
    y_{n+1} = y_n + h\cdot f(t_{n+1}, y_{n+1}).    
\end{equation*}

For equations lacking an algebraic solution, several techniques estimate $y_{n+1}$ iteratively.
As we described in Section~\ref{sec:renoise}, the inversion process lacks a closed-form solution, as shown in Equation~\ref{eq:reverse_eq}.
To address this, the ReNoise method leverages fixed-point iterations, which we refer to as \emph{reonising iterations,} to progressively refine the estimate of $y_{n+1}$:
\begin{equation*}
    y_{n+1}^{(0)} = y_n, \quad y_{n+1}^{(k+1)} = y_n + h\cdot f(t_{n+1}, y_{n+1}^{(k)}).
\end{equation*}
In our ReNoise method, we average these renoising iterations to mitigate convergence errors, leading to improvement in the reconstruction quality.

\paragraph{\textbf{Renoising Toy Example}}
We begin with the simple toy example, the diffusion of a shifted Gaussian.
Given the initial distribution $\mu_0 \sim \mathcal{N}(a, I)$, where $a$ is a non-zero shift value and $I$ is the identity matrix. The diffusion process defines the family of distributions $\mu_t \sim \mathcal{N}(ae^{-t}, I)$, and the probability flow ODE takes the form $\frac{dz}{dt} = -ae^{-t}$ (see \cite{khrulkov2022understanding} for details). The Euler solver step at a state $(z_t, t)$, and timestep $\Delta t$ moves it to $(z_{t+\Delta t}^{(1)}, t + \Delta t) = (z_t - ae^{-t} \cdot \Delta t, t+ \Delta t)$.
Notably, the backward Euler step at this point does not lead to $z_t$.
After applying the first renoising iteration, we get $(z^{(2)}_{t + \Delta t}, t + \Delta t) = (z_t - ae^{-(t + \Delta t)} \cdot \Delta t, t + \Delta t)$ and the backward Euler step at this point leads exactly to $(z_t, t)$.
Thus, in this simple example, we successfully estimates the exact pre-image after a single step.
While this convergence cannot be guaranteed in the general case, in the following, we discuss some sufficient conditions for the algorithm's convergence and empirically verify them for the image diffusion model.

\ifeccv
    \input{figures/mini_pages/convergence_fig_eccv}
\else
    \input{figures/consequent_diff_figure}
\fi

\vspace{-5pt}
\paragraph{\textbf{ReNoise Convergence}}
During the inversion process, we aim to find the next noise level inversion, denoted by $\hat{z}_{t}$, such that applying the denoising step to $\hat{z}_{t}$ recovers the previous state, $z_{t-1}$.
Given the noise estimation $\epsilon_{\theta}(z_{t}, t)$ and a given $z_{t-1}$, the ReNoise mapping defined in Section \ref{sec:renoise} can be written as $\mathcal{G}: z_{t} \to \mathrm{InverseStep}(z_{t-1}, \epsilon_{\theta}(z_{t}, t))$.
For example, in the case of using DDIM sampler the mapping is $\mathcal{G}(z_{t}) = \frac{1}{\phi_t}(z_{t-1} - \psi_t \epsilon_{\theta}(z_{t}, t))$.
The point $\hat{z}_{t}$, 
which is mapped to $z_{t-1}$ after the denoising step, is a stationary point of this mapping.
Given $z_{t}^{(1)}$, the first approximation of the next noise level $z_{t}$, our goal is to show that the sequence $z_{t}^{(k)} = \mathcal{G}^{k-1}(z_{t}^{(1)}), k \to \infty$ converges. As the mapping $\mathcal{G}$ is continuous, the limit point would be its stationary point. The definition of $\mathcal{G}$ gives:
$$
\| z_{t}^{(k + 1)} - z_{t}^{(k)} \| = \|\mathcal{G}(z_{t}^{(k)}) - \mathcal{G}(z_{t}^{(k - 1)})\|,
$$
where the norm is always assumed as the $l_2$-norm. For the ease of the notations, we define $\Delta^{(k)} = z_{t}^{(k)} - z_{t}^{(k - 1)}$. For convergence proof, it is sufficient to show that the sum of norms of these differences converges, which will imply that $z_{t}^{(k)}$ is the Cauchy sequence. Below we check that in practice $\|\Delta^{(k)}\|$ decreases exponentially as $k \to \infty$ and thus has finite sum.
In the assumption that $\mathcal{G}$ is $\mathcal{C}^{2}$-smooth, the Taylor series conducts:
\begin{multline*}
\|\Delta^{(k + 1)}\| = \|\mathcal{G}(z_{t}^{(k)}) - \mathcal{G}(z_{t}^{(k - 1)})\| = \\
\|\mathcal{G}(z_{t}^{(k - 1)}) + \frac{\partial \mathcal{G}}{\partial z}|_{z_{t}^{(k - 1)}} \cdot \Delta^{(k)} + O(\|\Delta^{(k)}\|^2) - \mathcal{G}(z_{t}^{(k - 1)})\| = \\
\|\frac{\partial \mathcal{G}}{\partial z}|_{z_{t}^{(k - 1)}} \cdot \Delta^{(k)} + O(\|\Delta^{(k)}\|^2)\| \leq 
\ifeccv
\else
\\
\fi
\|\frac{\partial \mathcal{G}}{\partial z}|_{z_{t}^{(k - 1)}}\| \cdot \|\Delta^{(k)}\| + O(\|\Delta^{(k)}\|^2) = \\
\frac{\psi_t}{\phi_t} \cdot \|\frac{\partial \epsilon_{\theta}}{\partial z}|_{z_{t}^{(k - 1)}}\| \cdot \|\Delta^{(k)}\| + O(\|\Delta^{(k)}\|^2)
\end{multline*}
Thus, in a sufficiently small neighborhood, the convergence dynamics is defined by the scaled Jacobian norm $\frac{\psi_t}{\phi_t}~\cdot~\|\frac{\partial \epsilon_{\theta}}{\partial z}|_{z_{t}^{(k - 1)}}\|$.
\ifeccv
In the supplementary materials, we show this scaled norm estimation for the SDXL diffusion model for various steps and ReNoise iterations indices $(k)$.
\else
In the Appendix~\ref{sec:sup_theory}, we show this scaled norm estimation for the SDXL diffusion model for various steps and ReNoise iterations indices $(k)$.
\fi
Remarkably, the ReNoise indices minimally impact the scale factor, consistently remaining below 1. This confirms in practice the convergence of the proposed algorithm.
Notably, the highest scaled norm values occur at smaller $t$ (excluding the first step) and during the initial renoising iteration. 
This validates the strategy of not applying ReNoise in early steps, where convergence tends to be slower compared to other noise levels.
Additionally, the scaled norm value for the initial $t$ approaches 0, which induces almost immediate convergence.

Figure~\ref{fig:consequent_diff} illustrates the exponential decrease in distances between consecutive elements $z_t^{(k)}$ and $z_t^{(k+1)}$, which confirms the algorithm's convergence towards the stationary point of the operator $\mathcal{G}$.

The proposed averaging strategy is aligned with the conclusions described above, and also converges to the desired stationary point.
\ifeccv
In The supplementary materials, we present a validation for this claim.
\else
In The Appendix~\ref{sec:sup_theory}, we present a validation for this claim.
\fi

%% file: figures/mini_pages/convergence_fig_eccv.tex
\begin{figure}[t]
    \centering
    \scriptsize
    \begin{minipage}[b]{0.57\textwidth}
        \centering
        \input{figures/non_monotonic_figure_content}
    \end{minipage}
    \hfill
    \begin{minipage}[b]{0.40\textwidth}
        \centering
        \input{figures/consequent_diff_content}
    \end{minipage}
    \vspace{-15pt}
\end{figure}

%% file: figures/consequent_diff_content.tex
\includegraphics[width=\columnwidth]{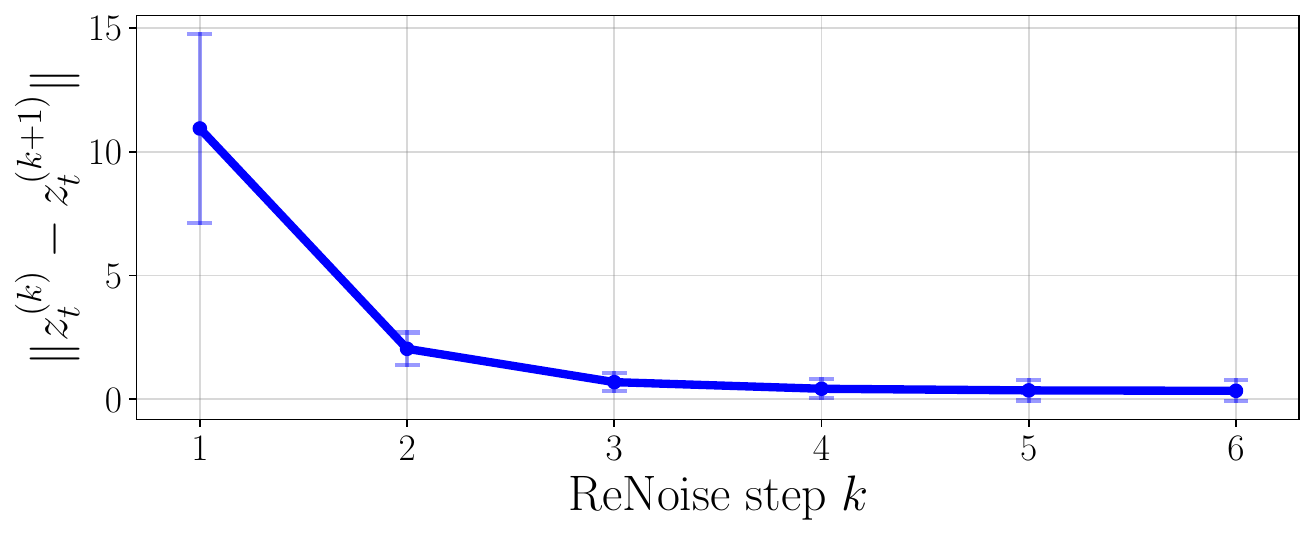}
\ifeccv
\else
    \vspace{-0.7cm}
\fi
\caption{
Average distance between consequent estimations $z_t^{(k)}$, and $z_t^{(k+1)}$. Vertical bars indicate the standard deviation. The averages are computed over 32 images and 10 different timesteps.
\ifeccv
\else
    \vspace{-26pt}
\fi
}
\label{fig:consequent_diff}

%% file: figures/consequent_diff_figure.tex
\begin{figure}
    \centering
    \input{figures/consequent_diff_content}
\end{figure}

%% file: 6-experiments.tex
\ifeccv
    \vspace{-10pt}
ֿ\else
    \input{figures/psnr_unet_op_figure}
\fi

\section{Experiments}

\ifeccv
    \vspace{-5pt}
\fi

In this section, we conduct extensive experiments to validate the effectiveness of our method. We evaluate both the reconstruction quality of our inversion and its editability. To demonstrate the versatility of our approach, we apply it to four models, SD~\cite{rombach2022highresolution}, SDXL~\cite{podell2023sdxl}, SDXL Turbo~\cite{sauer2023adversarial}, and LCM-LoRA~\cite{luo2023lcm}, with SDXL Turbo and LCM-LoRA being few-step models. Additionally, we use various sampling algorithms including both deterministic and non-deterministic ones.
\ifeccv
Implementation details for each model are provided in the supplementary materials.
\else
Implementation details for each model are provided in Appendix~\ref{sec:imp_details}.
\fi
Following previous works~\cite{mokady2022nulltext, Alaluf_2021}, we quantitatively evaluate our method with three metrics: $L_2$, LPIPS~\cite{zhang2018perceptual}, and PSNR.
Unless stated otherwise, for both inversion and generation we use the prompt obtained from BLIP2~\cite{li2023blip}.

\ifeccv
    \vspace{-12pt}
\else
\fi

\subsection{Reconstruction and Speed}

\ifeccv
    \vspace{-3pt}
\fi

We begin by evaluating the reconstruction-speed tradeoff.
The main computational cost of both the inversion and denoising processes is the forward pass through the UNet. In each renoising iteration, we perform one forward pass, which makes it computationally equal to a standard inversion step (as done in DDIM Inversion for example).
In the following experiments, we compare the results of a sampler reversing with our method, where we match the number of UNet passes between the methods.
For example, 8 steps of sampler reversing are compared against 4 steps with one renoising iteration at each step.

\ifeccv
    \input{figures/mini_pages/tables_eccv}
    \input{figures/mini_pages/exp_fig_eccv}
\else
    \input{figures/quan_comapre_unet_ops}
    \input{tables/unet_ops_budget_table}
\fi

\ifeccv
    \vspace{-8pt}
\else
    \vspace{-10pt}
\fi

\paragraph{\textbf{Qualitative Results}}
In Figure \ref{fig:quan_compare_unet_ops} we show qualitative results of image reconstruction on SDXL Turbo~\cite{sauer2023adversarial}. Here, we utilize DDIM as the sampler, and apply four denoising steps for all configurations. Each row exhibits results obtained using a different amount of UNet operations. In our method, we apply four inversion steps, and a varying number of renoising iterations. As can be seen, the addition of renoising iterations gradually improves the reconstruction results. 
Conversely, employing more inversion steps proves insufficient for capturing all details in the image, as evident by the background of the car, or even detrimental to the reconstruction, as observed in the Uluro example.

\ifeccv
    \vspace{-8pt}
\else
    \vspace{-12pt}
\fi

\paragraph{\textbf{Quantitative Results}}

For the quantitative evaluation, we use the MS-COCO 2017~\cite{cocodataset} validation dataset. Specifically, we retain images with a resolution greater than $420\times420$, resulting in a dataset containing 3,865 images.

We begin by evaluating both the sampler reversing approach and our ReNoise method, while varying the number of UNet operations during the inversion process and keeping the number of denoising steps fixed. This experiment is conducted using various models (SDXL, SDXL Turbo, LCM) and samplers.
For all models, we utilize the DDIM~\cite{song2022denoising} sampler. In addition, we employ the Ancestral-Euler scheduler for SDXL Turbo, and the default LCM sampler for LCM-LoRA.
We set the number of denoising steps to 50 for SDXL, and to 4 for SDXL Turbo and LCM-LoRA.
Quantitative results, using PSNR as the metric, are presented in Figure~\ref{fig:psnr_unet_op_figure}. 
We evaluate our method using different configurations.
The x-axis refers to the number of UNet operations in the inversion process.
\ifeccv
Other metrics results are provided in the supplementary materials.
\else
LPIPS metrics results are provided in Appendix~\ref{sec:sup_reconstruction_and_speed}.
\fi

\ifeccv
    \input{figures/psnr_unet_op_figure}
\fi

As depicted in the graphs, incorporating additional renoising iterations proves to be more beneficial for image reconstruction compared to adding more inversion steps. Note that the performance of the Ancestral-Euler and LCM samplers noticeably degrades when the number of inversion steps exceeds the number of denoising steps. Unlike DDIM, these samplers have $\Phi_t \approx 1$, resulting in an increase in the latent vector's norm beyond what can be effectively denoised in fewer steps.
In this experiment, we maintain the same number of UNet operations for both ReNoise and the sampler reversing approach. However, in ReNoise, the number of inversion steps remains fixed, and the additional operations are utilized for renoising iterations, refining each point on the inversion trajectory. Consequently, our method facilitates improved reconstruction when using these noise samplers.

We continue by evaluating both the sampler reversing approach and our method while maintaining a fixed total number of UNet operations for the inversion and denoising processes combined.
The results for SDXL with DDIM are presented in Table~\ref{tb:unet_ops_budget_table}.
The table displays various combinations of inversion, denoising, and renoising steps, totaling 100 UNet operations.
Despite employing longer strides along the inversion and denoising trajectories, our ReNoise method yields improved reconstruction accuracy, as evident in the table. Furthermore, a reduced number of denoising steps facilitates faster image editing, especially since it commonly involves reusing the same inversion for multiple edits.

\ifeccv
\else
    \input{figures/lcm_results_figure}
\fi

\ifeccv
    \vspace{-10pt}
\fi

\subsection{Reconstruction and Editability}
In Figure~\ref{fig:lcm_results}, we illustrate editing results generated by our method with LCM LoRA~\cite{luo2023lcm}. These results were obtained by inverting the image using a source prompt and denoising it with a target prompt.
Each row exhibits an image followed by three edits accomplished by modifying the original prompt. These edits entail either replacing the object word or adding descriptive adjectives to it.
As can be seen, the edited images retain the details present in the original image. For instance, when replacing the cat with a koala, the details in the background are adequately preserved.

\ifeccv
\else
    \input{figures/ablation}
\fi

\ifeccv
    \vspace{-10pt}
\fi

\subsection{Ablation Studies}

\ifeccv
    \vspace{-5pt}
\fi
\ifeccv
\else
    \paragraph{\textbf{Image Reconstruction}}
\fi

Figure \ref{fig:ablation} qualitatively demonstrates the effects of each component in our method, highlighting their contribution to the final outcome.
Here, we use SDXL Turbo model~\cite{sauer2023adversarial}, with the Ancestral-Euler sampler, which is non-deterministic.
As our baseline, we simply reverse the sampler process.
The reconstruction, while semantically capturing the main object, fails to reproduce the image's unique details. 
\ifeccv
    For example, in the bird image, the reconstruction contains a bird standing on a branch, but the branch is in a different pose and the bird is completely different.
\else
    For example, in the middle column, the image contains a bird standing on a branch, but the branch is in a different pose and the bird is completely different.
\fi
Using 9 ReNoise iterations significantly improves the reconstruction, recovering finer details like the bird's original pose and branch texture. However, some subtle details, such as the bird's colors or the color in Brad Pitt's image, remain incomplete.
Averaging the final iterations effectively incorporates information from multiple predictions, leading to a more robust reconstruction that captures finer details.
Regularize the UNet's noise prediction with $\mathcal{L}_{\text{edit}}$ can introduce minor artifacts to the reconstruction, evident in the smoother appearance of the hair of the two people on the left, or in the cake example.
Finally, we present our full method by adding the noise correction technique.

Table \ref{tb:quantitative_ablation} quantitatively showcases the effect each component has on reconstruction results. As can be seen, the best results were obtained by our full method or by averaging the last estimations of $z_t$.
Our final method also offers the distinct advantage of getting an editable latent representation.

\ifeccv
In the supplementary materials, we present an ablation study to justify our editability enhancement and noise correction components.
\else
In Appendix~\ref{sec:sup_img_edit_ablation}, we present an ablation study to justify our editability enhancement and noise correction components.
\fi

\ifeccv
    \input{figures/lcm_results_figure}
\else
    \input{tables/ablation_table}
    \input{figures/turbo_compare_ddpm}
\fi

\ifeccv
    \vspace{-10pt}
\fi

\subsection{Comparisons}

\ifeccv
    \vspace{-7pt}
\fi

\paragraph{\textbf{Inversion for Non-deterministic Samplers.}}

In Figure~\ref{fig:turbo_compare_ddpm} we show a qualitative comparison with ``an edit-friendly DDPM''~\cite{hubermanspiegelglas2023edit} where we utilize SDXL Turbo~\cite{sauer2023adversarial}. Specifically, we assess the performance of the edit-friendly DDPM method alongside our ReNoise method in terms of both reconstruction and editing.

We observe that in non-deterministic samplers like DDPM, the parameter $\rho_0$ in Equation~\ref{eq:sampler} equals zero. This means that in the final denoising step, the random noise addition is skipped to obtain a clean image. In long diffusion processes (e.g., 50-100 steps), the final denoising step often has minimal impact as the majority of image details have already been determined. Conversely, shorter diffusion processes rely on the final denoising step to determine fine details of the image.
Due to focusing solely on noise correction to preserve the original image in the inversion process, the edit-friendly DDPM struggles to reconstruct fine details of the image, such as the shower behind the cat
\ifeccv
.
\else
in the right example.
\fi
However, our ReNoise method finds an inversion trajectory that faithfully reconstructs the image and does not rely solely on noise corrections. This allows us to better reconstruct fine details such as the shower.

Additionally, encoding a significant amount of information within only a few external noise vectors, $\epsilon_t$, limits editability in certain scenarios, 
\ifeccv
See more examples in the supplementary materials
\else
such as the ginger cat example. It is evident that the edit-friendly DDPM method struggles to deviate significantly from the original image while also failing to faithfully preserve it. For instance, it encounters difficulty in transforming the cat into a ginger cat while omitting the preservation of the decoration in the top left corner.
\fi

\ifeccv
    \input{figures/mini_pages/exp_fig_eccv2}
\else
    \input{figures/reconstruction_comparisons_figure}
\fi

\ifeccv
    \vspace{-7pt}
\fi

\paragraph{\textbf{Null-prompt Inversion Methods}}

In Figure~\ref{fig:reconstruction_comparisons_figure}, we present a qualitative comparison between our method and null-text based inversion methods. For this comparison, we utilize Stable Diffusion~\cite{rombach2022highresolution} since these methods rely on a CFG~\cite{ho2022classifierfree} mechanism, which is not employed in SDXL Turbo~\cite{sauer2023adversarial}.
Specifically, we compare DDIM Inversion~\cite{song2022denoising} with one renoising iteration to Null-Text Inversion (NTI)\cite{mokady2022nulltext} and Negative-Prompt Inversion (NPI)\cite{miyake2023negativeprompt}. Both NTI and NPI enhance the inversion process by replacing the null-text token embedding when applying CFG.
Our method achieves results comparable to NTI, while NPI highlights the limitations of plain DDIM inversion. This is because NPI sets the original prompt as the negative prompt, essentially resulting in an inversion process identical to plain DDIM inversion.
Regarding running time, our ReNoise inversion process takes 13 seconds, significantly faster than NTI's 3 minutes. For comparison, plain DDIM inversion and NPI each take 9 seconds.

%% file: figures/psnr_unet_op_figure.tex
\begin{figure*}
    \centering
    \setlength{\tabcolsep}{1pt}
    \addtolength{\belowcaptionskip}{-10pt}
    {\small
    \centering

    \begin{tabular}{c c c}
        { SDXL } &
        { SDXL Turbo } &
        { LCM } \\
        
        \includegraphics[width=0.33\textwidth]{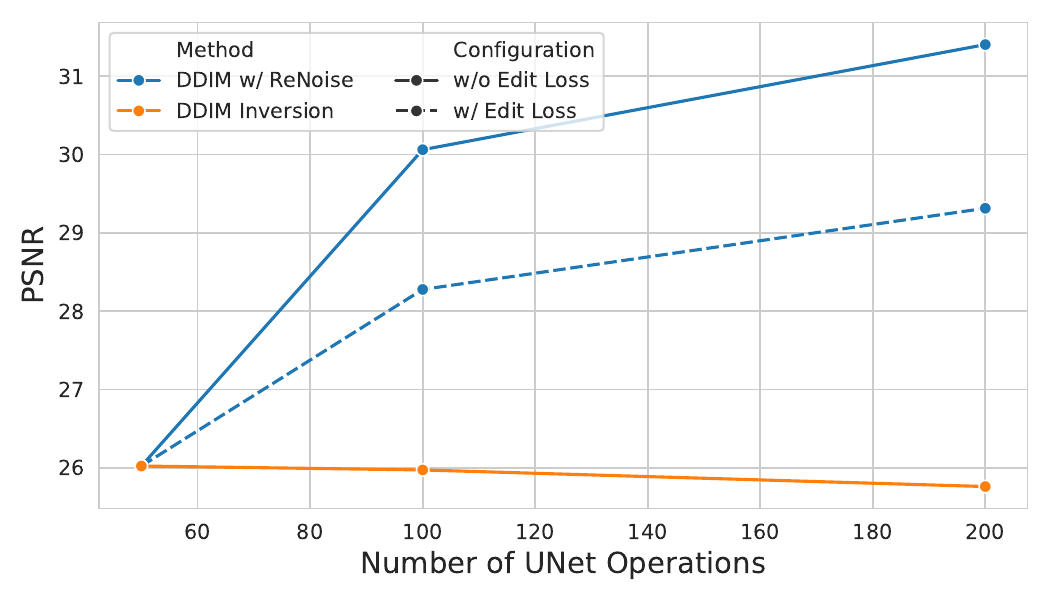} &
        \includegraphics[width=0.33\textwidth]{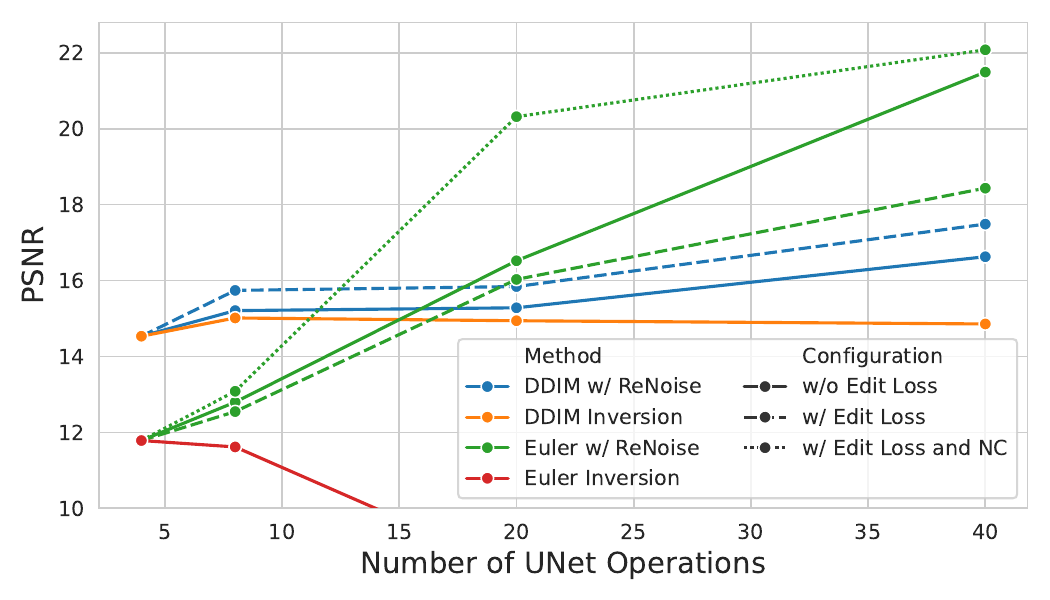} &
        \includegraphics[width=0.33\textwidth]{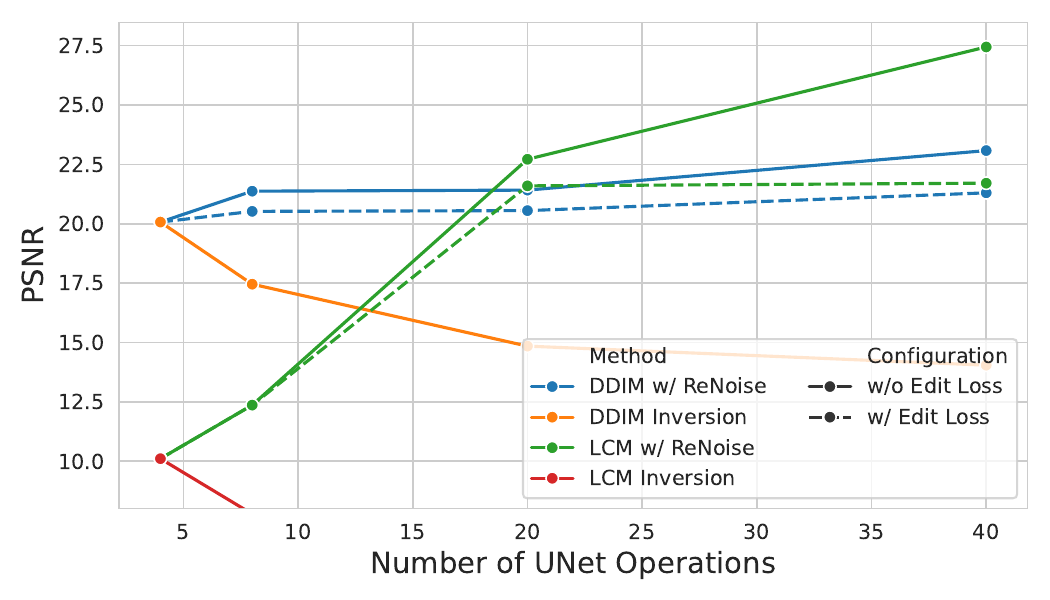} \\      
    \end{tabular}
    
    }
    \vspace{-0.225cm}    
    \caption{
    Image reconstruction results comparing sampler reversing inversion techniques across different samplers (e.g., vanilla DDIM inversion) with our ReNoise method using the same sampler. The number of denoising steps remains constant. However, the number of UNet passes varies, with the sampler reversing approach increasing the number of inversion steps, while our method increases the number of renoising iterations.
    We present various configuration options for our method, including options with or without edit enhancement loss and Noise Correction (NC).
    \ifeccv
    \else
        \vspace{-14pt}
    \fi
    }
    \label{fig:psnr_unet_op_figure}
    \ifeccv
    \vspace{-10pt}
    \else
    \vspace{-0.15cm}
    \fi
\end{figure*}

%% file: figures/mini_pages/tables_eccv.tex
\begin{table}[t]
    \centering
    \scriptsize
    \begin{minipage}[b]{0.48\textwidth}
        \centering
        \input{tables/unet_ops_budget_table_content}
    \end{minipage}
    \hfill
    \begin{minipage}[b]{0.48\textwidth}
        \centering
        \input{tables/ablation_table_content}
    \end{minipage}
    \vspace{-18pt}
\end{table}

%% file: tables/unet_ops_budget_table_content.tex
\caption{
Image reconstruction results with a fixed number of 100 UNet operations. Each row showcases the results obtained using different combinations of inversion steps, denoising steps, and renoising iterations, totaling 100 operations. As observed, allocating some of the operations to renoising iterations improves the reconstruction quality while maintaining the same execution time.
\ifeccv
\else
    \vspace{-26pt}
\fi
}
\ifeccv
    \vspace{-5pt}
\else
\fi
\begin{tabular}{c c c | c c c}
\toprule
\ifeccv
    \multicolumn{6}{c}{Image Reconstruction Results} \\
\else
    \multicolumn{6}{c}{Image Reconstruction With a Fixed Number of UNet Operations} \\
\fi

\midrule

\ifeccv
Inv         & Inf       & ReNoise   & L2 $\downarrow$     & PSNR $\uparrow$    & LPIPS $\downarrow$ \\
\else
Inversion   & Inference & ReNoise   & L2 $\downarrow$     & PSNR $\uparrow$    & LPIPS $\downarrow$ \\
\fi
Steps       & Steps     & Steps     &                     &                    & \\
\midrule
50          & 50        & 0         & 0.00364             & 26.023             & 0.06273 \\
75          & 25        & 0         & 0.00382             & 25.466             & 0.06605 \\
80          & 20        & 0         & 0.00408             & 25.045             & 0.07099 \\
90          & 10        & 0         & 0.01023             & 20.249             & 0.10305 \\
25          & 25        & 2         & \underline{0.00182} & \underline{29.569} & \underline{0.03637} \\
20          & 20        & 3         & \textbf{0.00167}    & \textbf{29.884}    & \textbf{0.03633} \\
10          & 10        & 8         & 0.00230             & 28.156             & 0.04678 \\
\bottomrule
\end{tabular}
\label{tb:unet_ops_budget_table}

%% file: tables/ablation_table_content.tex
\caption{
Quantitative ablation study on SDXL Turbo. We demonstrate the impact of each component of our inversion method on reconstruction results. The results improve with additional renoising iterations and significant enhancements occur through averaging final estimations. Additionally, we observe a reconstruction-editability trade-off, with edit losses causing degradation that is effectively mitigated by Noise Correction.
\ifeccv
\else
    \vspace{-10pt}
\fi
}
\vspace{-5pt}
\begin{tabular}{l c c c} 
    \toprule
    \multicolumn{4}{c}{Ablation - Image Reconstruction } \\
    \midrule
                        & L2 $\downarrow$     & PSNR $\uparrow$     &  LPIPS$ \downarrow$  \\
    \midrule
    Euler Inversion     & 0.0700              & 11.784              & 0.20337 \\
    + 1 ReNoise         & 0.0552             & 12.796              & 0.20254 \\
    + 4 ReNoise         & 0.0249              & 16.521              & 0.14821 \\
    + 9 ReNoise         &\underline{0.0126}   & 19.702              & 0.10850 \\
    + Averaging ReNoise & \textbf{0.0087}     & \underline{21.491}  & \underline{0.08832} \\
    + Edit Losses       & 0.0276              & 18.432              & 0.12616 \\
    + Noise Correction  & 0.0196              & \textbf{22.077}     & \textbf{0.08469} \\
    \bottomrule
\end{tabular}
\label{tb:quantitative_ablation}

%% file: figures/mini_pages/exp_fig_eccv.tex
\begin{figure}[t]
    \centering
    \scriptsize
    \begin{minipage}[b]{0.48\textwidth}
        \centering
        \input{figures/quan_comapre_unet_ops_content}
    \end{minipage}
    \hfill
    \begin{minipage}[b]{0.48\textwidth}
        \centering
        \input{figures/ablation_content}
    \end{minipage}
    \vspace{-20pt}
\end{figure}

%% file: figures/quan_comapre_unet_ops_content.tex
{\scriptsize
\centering

\begin{tabular}{c c c c c}
   
    \ifeccv
        \raisebox{7pt}{\rotatebox{90}{\begin{tabular}{c} Input \end{tabular}}} &
    \else
        \raisebox{7pt}{\rotatebox{90}{\begin{tabular}{c} Original \\ Image  \end{tabular}}} &
    \fi
    
    \includegraphics[width=0.21\columnwidth]{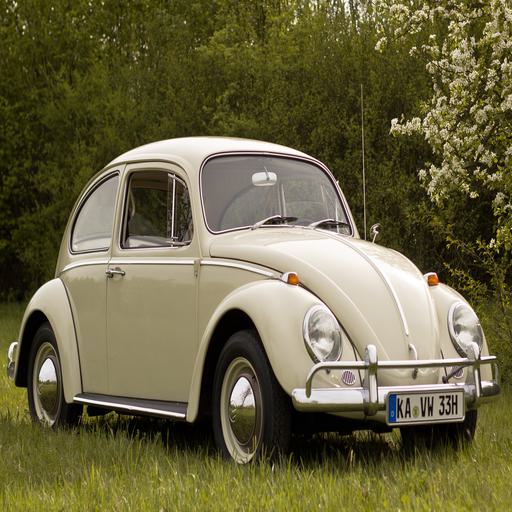} &
    \includegraphics[width=0.21\columnwidth]{images/dataset/car_2.jpg} &
    \includegraphics[width=0.21\columnwidth]{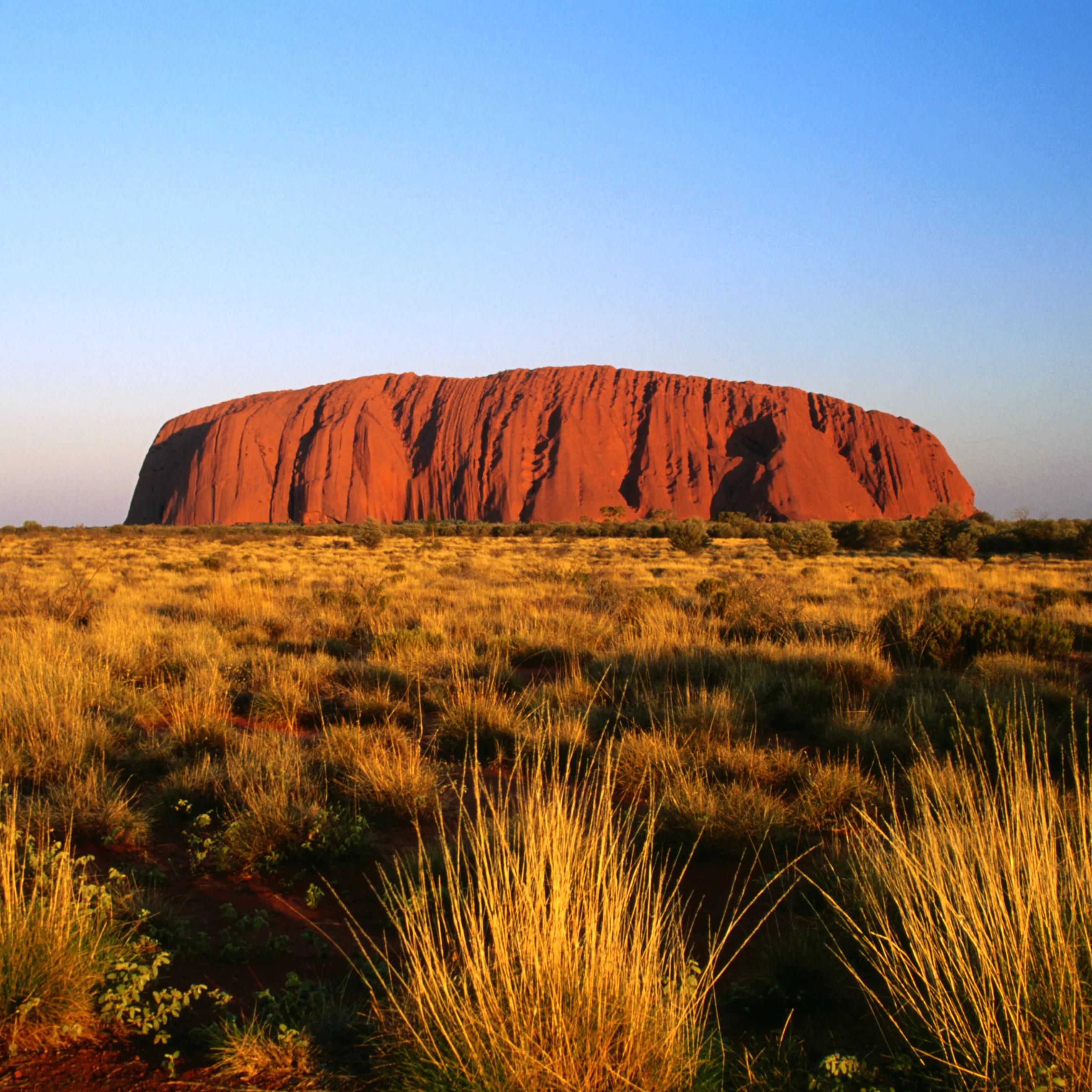} &
    \includegraphics[width=0.21\columnwidth]{images/dataset/landscape_1.jpg} \\

    \ifeccv
        \raisebox{8pt}{\rotatebox{90}{\begin{tabular}{c} 4 Ops  \end{tabular}}} &
    \else
        \raisebox{3pt}{\rotatebox{90}{\begin{tabular}{c} 4 UNet \\ Operations  \end{tabular}}} &
    \fi
    \includegraphics[width=0.21\columnwidth]{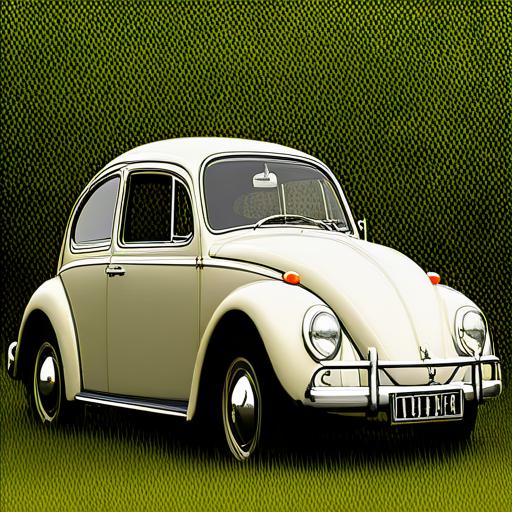} &
    \includegraphics[width=0.21\columnwidth]{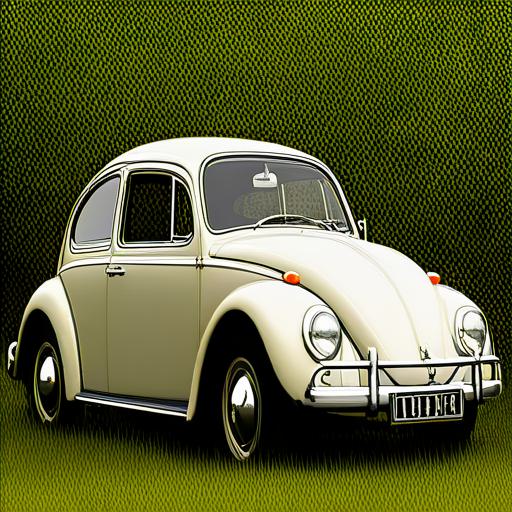} &
    \includegraphics[width=0.21\columnwidth]{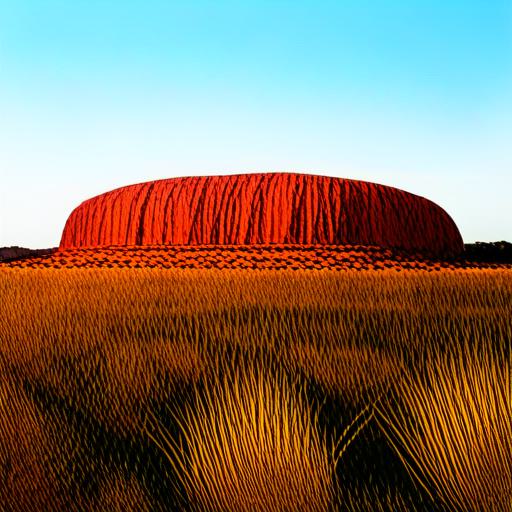} &
    \includegraphics[width=0.21\columnwidth]{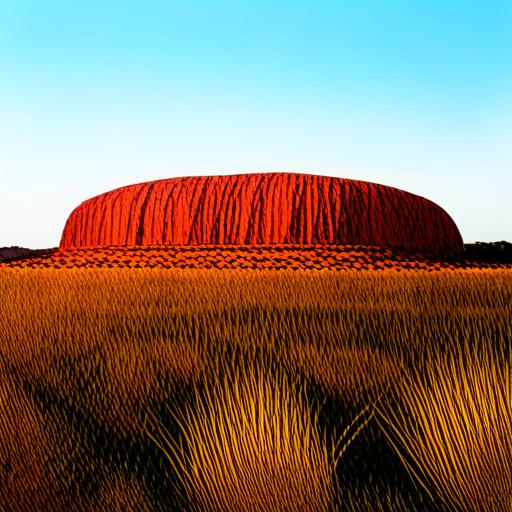}  \\

    \ifeccv
        \raisebox{8pt}{\rotatebox{90}{\begin{tabular}{c} 8 Ops  \end{tabular}}} &
    \else
        \raisebox{3pt}{\rotatebox{90}{\begin{tabular}{c} 8 UNet \\ Operations  \end{tabular}}} &
    \fi
    \includegraphics[width=0.21\columnwidth]{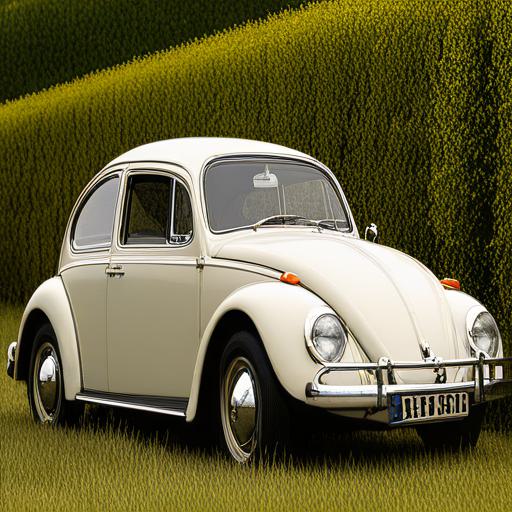} &
    \includegraphics[width=0.21\columnwidth]{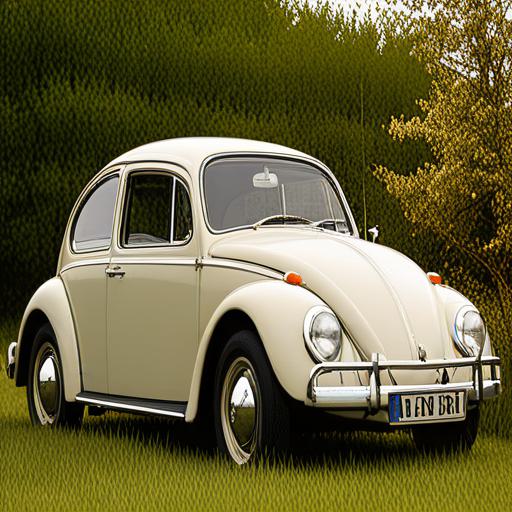} &
    \includegraphics[width=0.21\columnwidth]{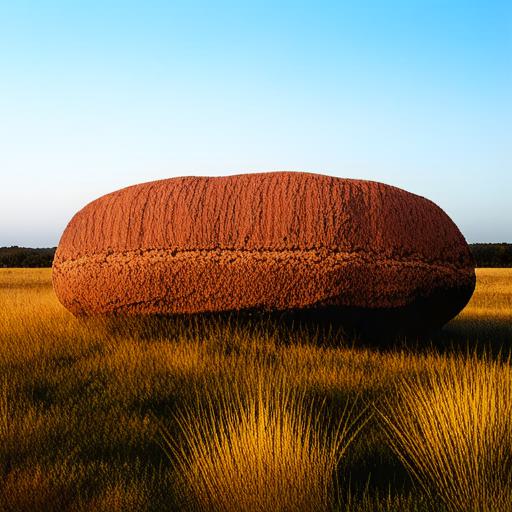} &
    \includegraphics[width=0.21\columnwidth]{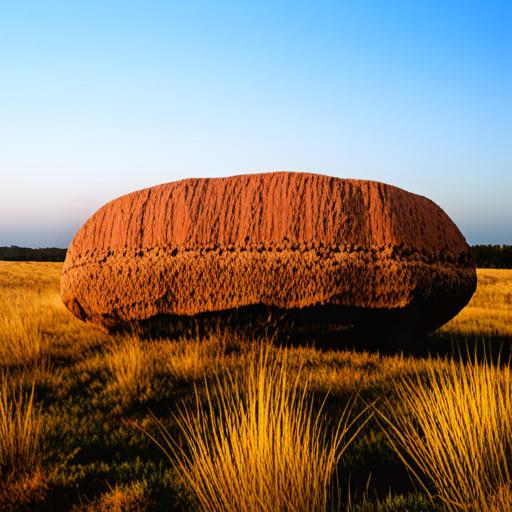}  \\

    \ifeccv
        \raisebox{5pt}{\rotatebox{90}{\begin{tabular}{c} 20 Ops  \end{tabular}}} &
    \else
        \raisebox{3pt}{\rotatebox{90}{\begin{tabular}{c} 20 UNet \\ Operations  \end{tabular}}} &
    \fi
    \includegraphics[width=0.21\columnwidth]{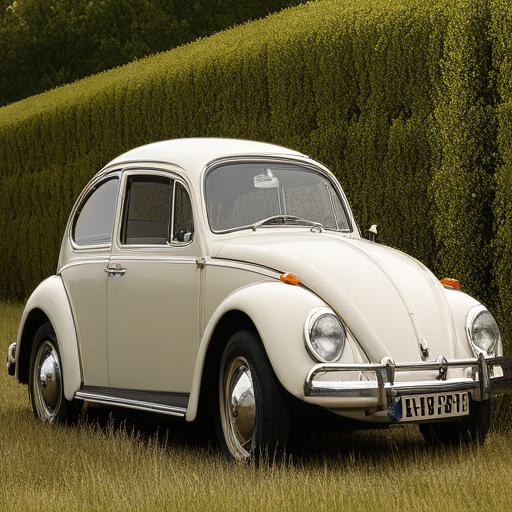} &
    \includegraphics[width=0.21\columnwidth]{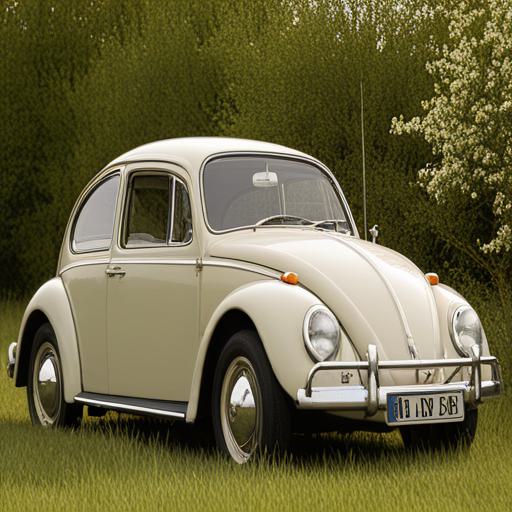} &
    \includegraphics[width=0.21\columnwidth]{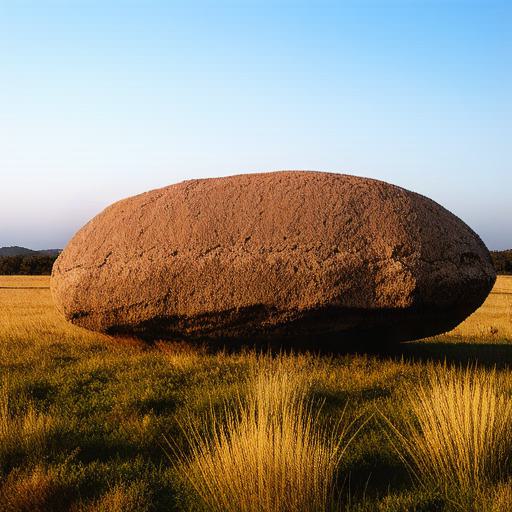} &
    \includegraphics[width=0.21\columnwidth]{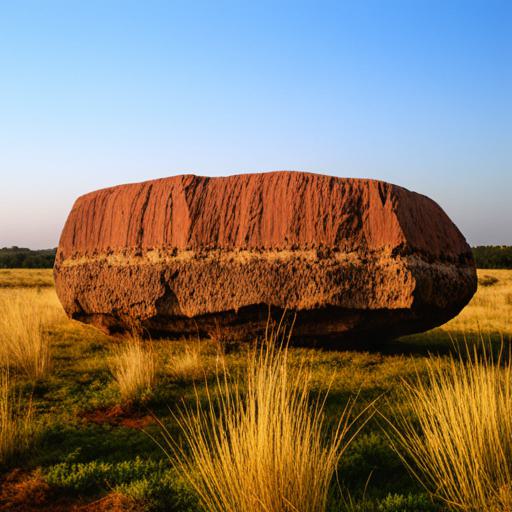}  \\

    \ifeccv
        \raisebox{5pt}{\rotatebox{90}{\begin{tabular}{c} 40 Ops  \end{tabular}}} &
    \else
        \raisebox{3pt}{\rotatebox{90}{\begin{tabular}{c} 40 UNet \\ Operations  \end{tabular}}} &
    \fi
    \includegraphics[width=0.21\columnwidth]{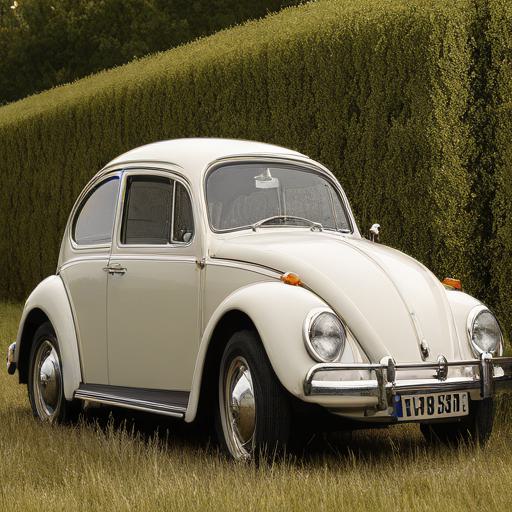} &
    \includegraphics[width=0.21\columnwidth]{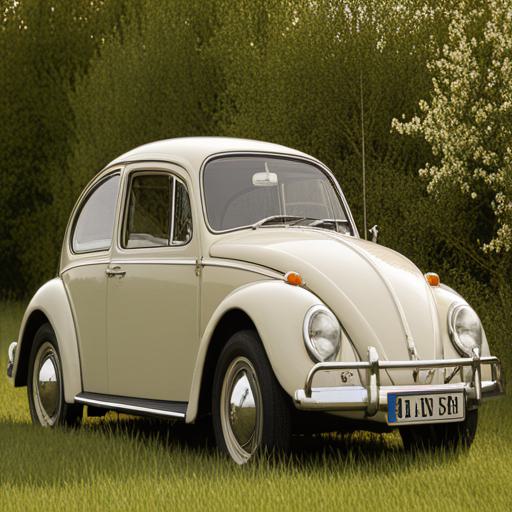} &
    \includegraphics[width=0.21\columnwidth]{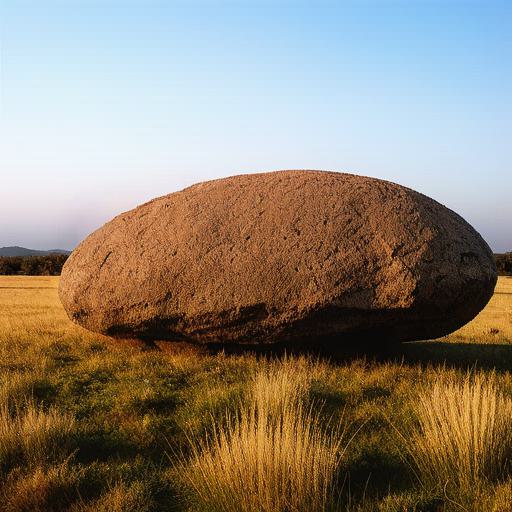} &
    \includegraphics[width=0.21\columnwidth]{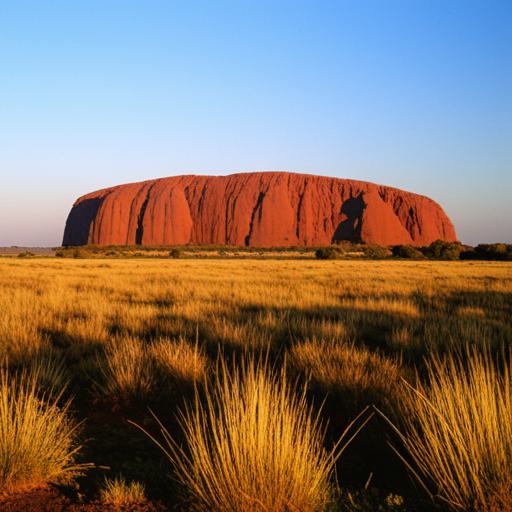}  \\
    
    &
    \ifeccv
        \begin{tabular}{c} DDIM \\ Inv.  \end{tabular} &
        \begin{tabular}{c} w/ Re-\\ Noise  \end{tabular} &
        \begin{tabular}{c} DDIM \\ Inv.  \end{tabular} &
        \begin{tabular}{c} w/ Re-\\ ReNoise  \end{tabular} \\
    \else
        \begin{tabular}{c} DDIM \\ Inversion  \end{tabular} &
        \begin{tabular}{c} w/ \\ ReNoise  \end{tabular} &
        \begin{tabular}{c} DDIM \\ Inversion  \end{tabular} &
        \begin{tabular}{c} w/ \\ ReNoise  \end{tabular} \\
    \fi
\end{tabular}

}
\ifeccv
    \vspace{-0.225cm}    
\else
    \vspace{-0.16cm}
\fi
\caption{
\ifeccv
Qualitative comparison between DDIM Inversion and our ReNoise method on SDXL Turbo (4 denoising steps). The first row displays the original images. The following rows display reconstructions from both methods, using the same number of UNet operations. DDIM employs smaller strides, while ReNoise utilizes more renoising steps.
\else
Qualitative comparison between DDIM Inversion to our ReNoise method using the DDIM sampler on SDXL Turbo. The first row presents the input images. 
In each subsequent row, we present the reconstruction results of both approaches, each utilizing the same number of UNet operations in the inversion process. To generate the images, we use 4 denoising steps in all cases. DDIM Inversion performs more inversion steps (i.e., smaller strides), while our method performs more renoising steps.
\vspace{-25pt}
\fi
}
\label{fig:quan_compare_unet_ops}

%% file: figures/ablation_content.tex
{\scriptsize
\centering

\ifeccv
    \begin{tabular}{c c c c c}
    
        \raisebox{7pt}{\rotatebox{90}{\begin{tabular}{c} Input  \end{tabular}}} &
        \includegraphics[width=0.22\columnwidth]{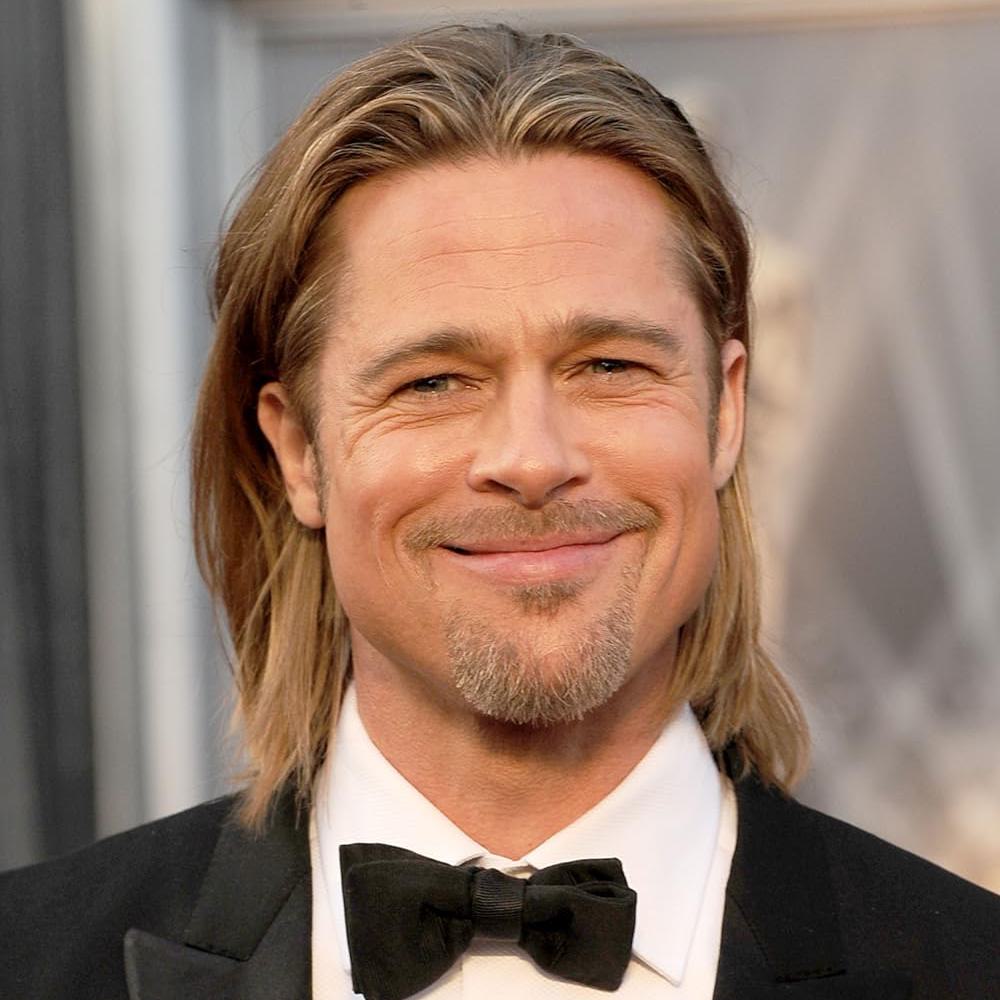} &
        \includegraphics[width=0.22\columnwidth]{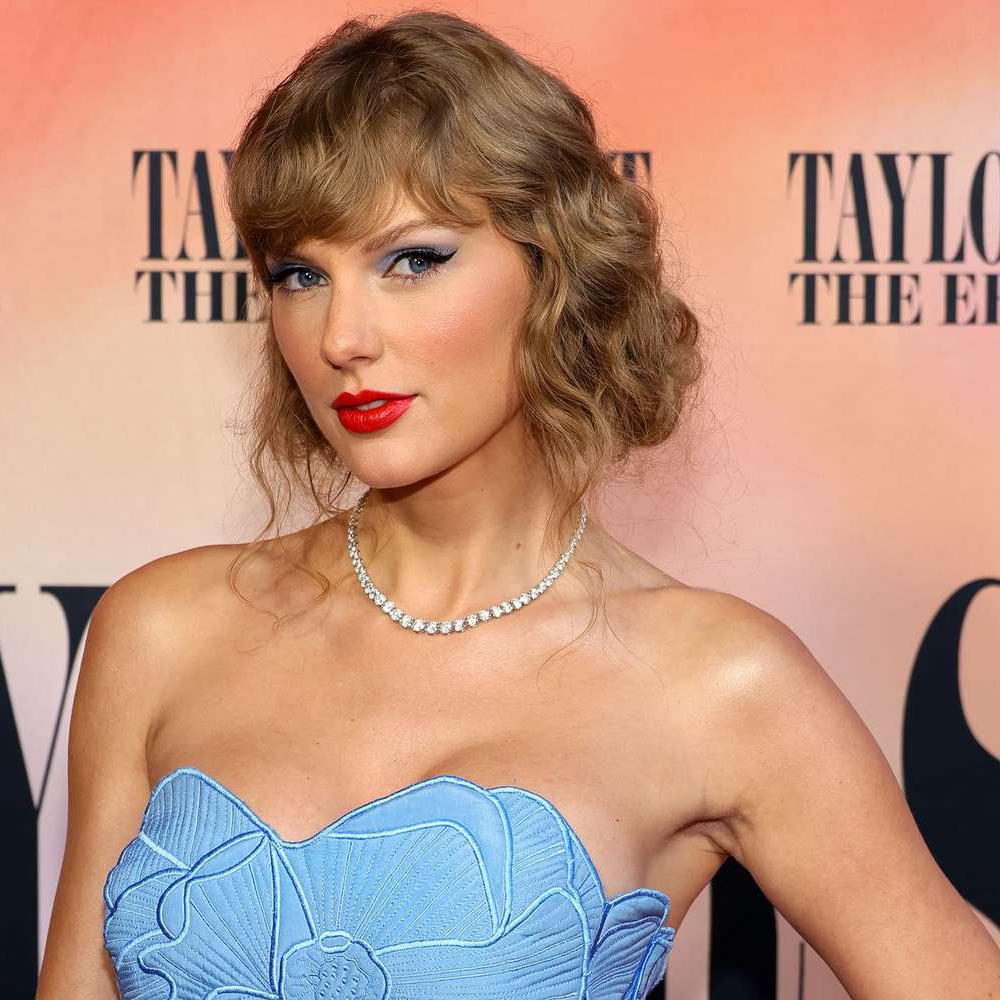} &
        \includegraphics[width=0.22\columnwidth]{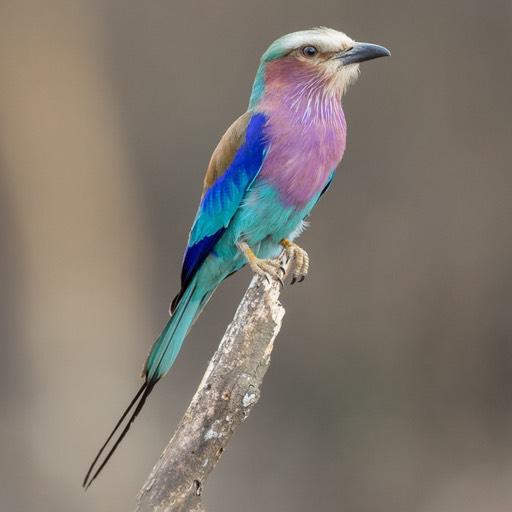} &
        \includegraphics[width=0.22\columnwidth]{images/ablation/more/coco51_origin.jpg} \\
        
        \raisebox{-1pt}{\rotatebox{90}{\begin{tabular}{c} Euler Inv.  \end{tabular}}} &
        \includegraphics[width=0.22\columnwidth]{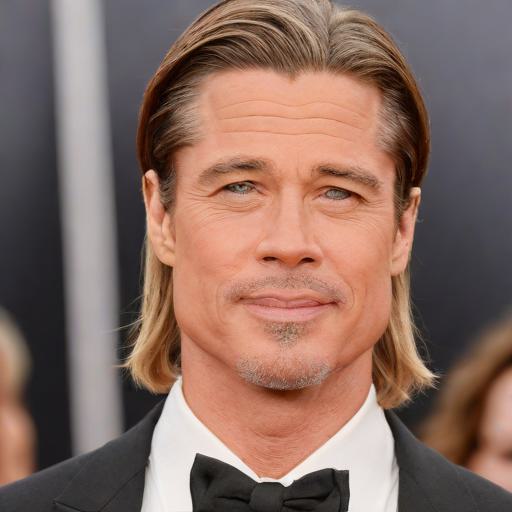} &
        \includegraphics[width=0.22\columnwidth]{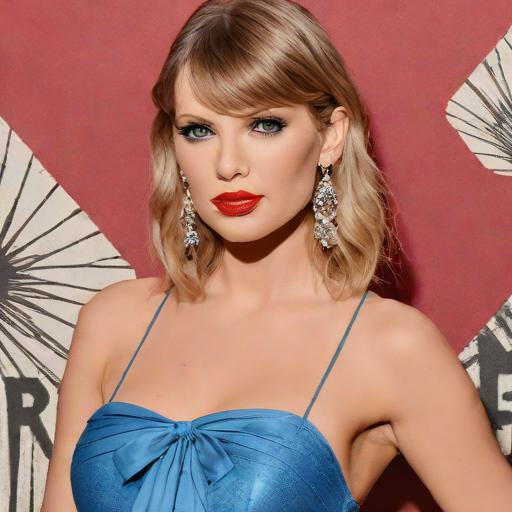} &
        \includegraphics[width=0.22\columnwidth]{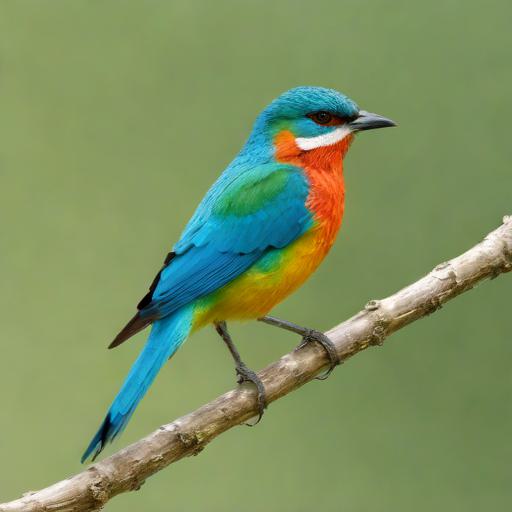} &
        \includegraphics[width=0.22\columnwidth]{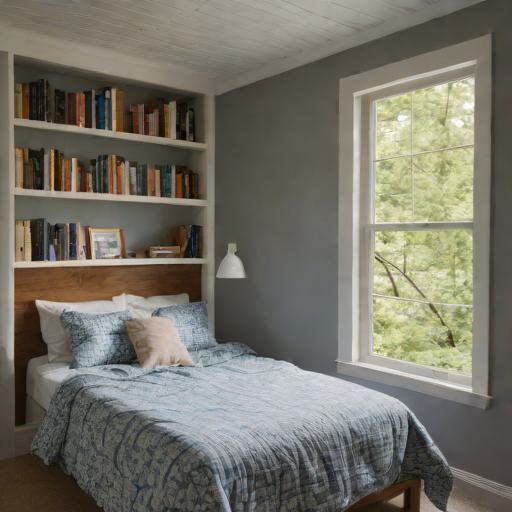} \\
    
        \raisebox{3pt}{\rotatebox{90}{\begin{tabular}{c} ReNoise  \end{tabular}}} &
        \includegraphics[width=0.22\columnwidth]{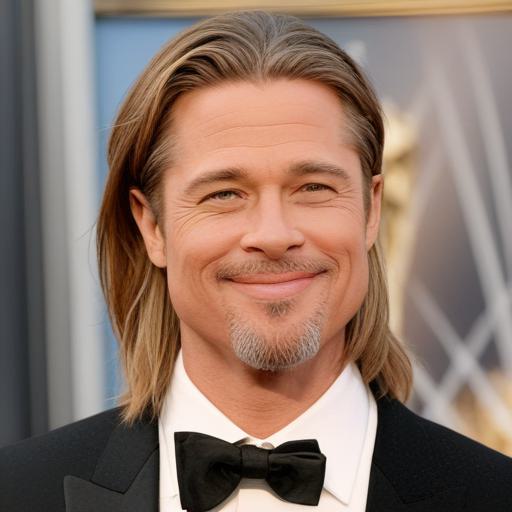} &
        \includegraphics[width=0.22\columnwidth]{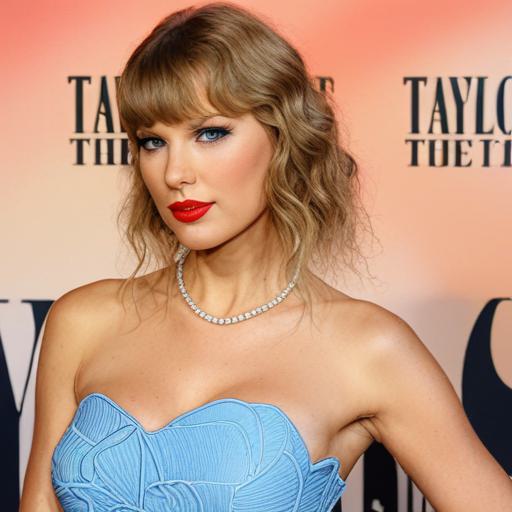} &
        \includegraphics[width=0.22\columnwidth]{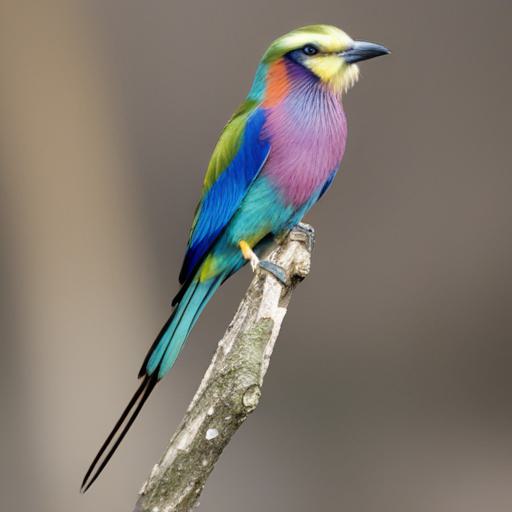} &
        \includegraphics[width=0.22\columnwidth]{images/ablation/more/coco51_9renise.jpg} \\
        
        \raisebox{4pt}{\rotatebox{90}{\begin{tabular}{c} w/ Avg.  \end{tabular}}} &
        \includegraphics[width=0.22\columnwidth]{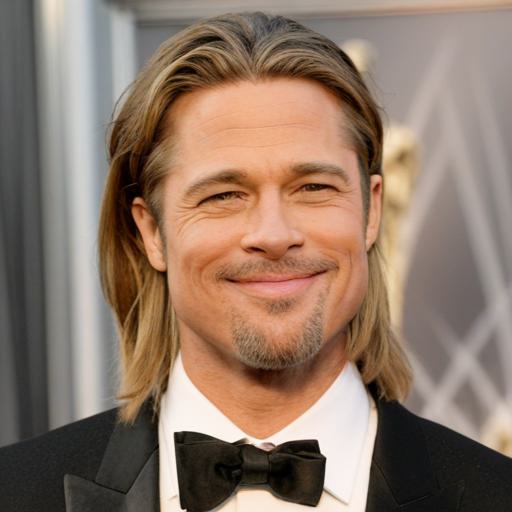} &
        \includegraphics[width=0.22\columnwidth]{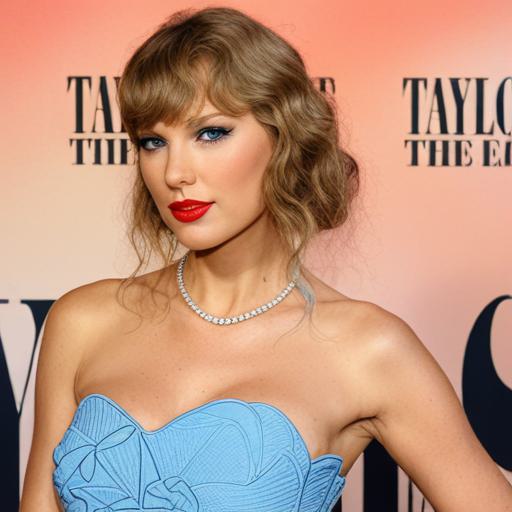} &
        \includegraphics[width=0.22\columnwidth]{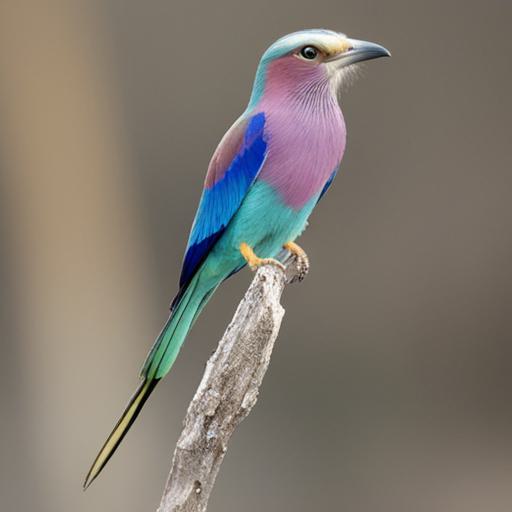} &
        \includegraphics[width=0.22\columnwidth]{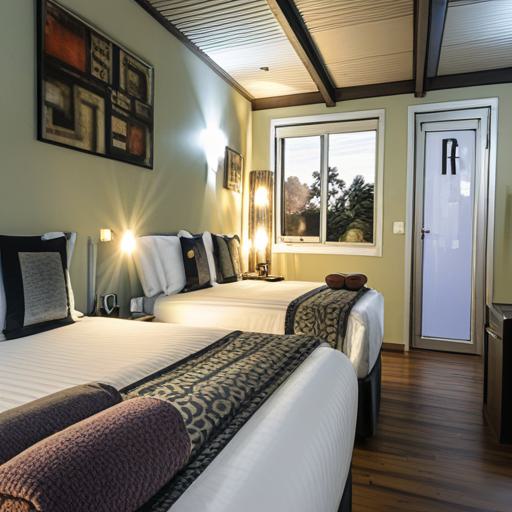} \\
        
        \raisebox{3pt}{\rotatebox{90}{\begin{tabular}{c} w/ $\mathcal{L}_{\text{loss}}$  \end{tabular}}} &
        \includegraphics[width=0.22\columnwidth]{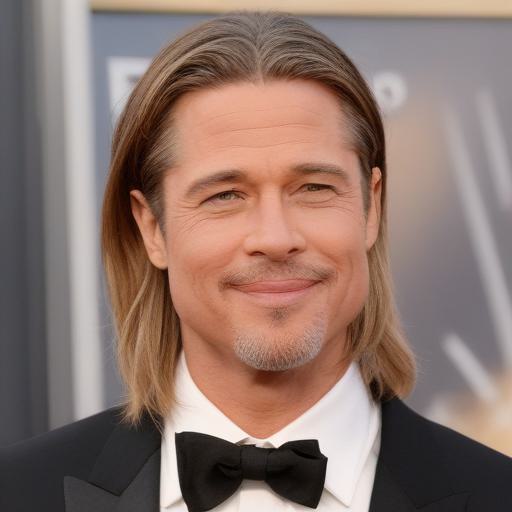} &
        \includegraphics[width=0.22\columnwidth]{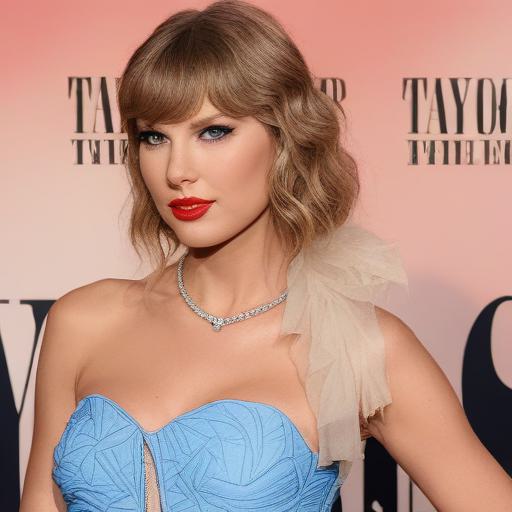} &
        \includegraphics[width=0.22\columnwidth]{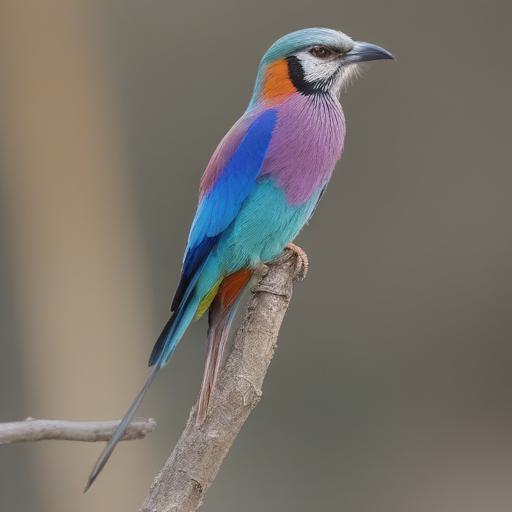} &
        \includegraphics[width=0.22\columnwidth]{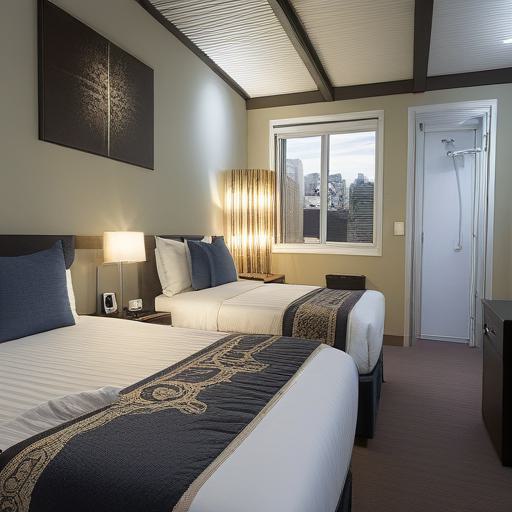} \\
        
        \raisebox{6pt}{\rotatebox{90}{\begin{tabular}{c} w/ NC  \end{tabular}}} &
        \includegraphics[width=0.22\columnwidth]{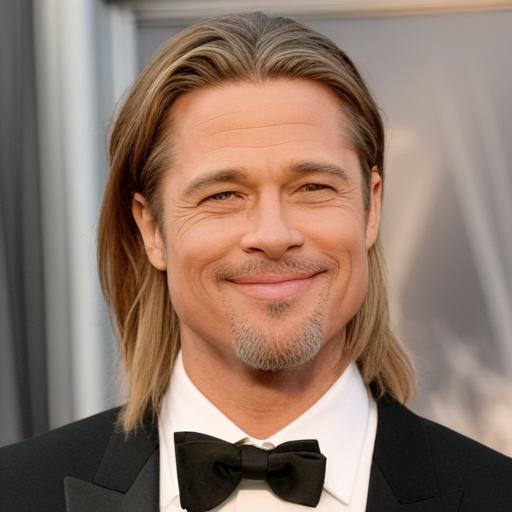} &
        \includegraphics[width=0.22\columnwidth]{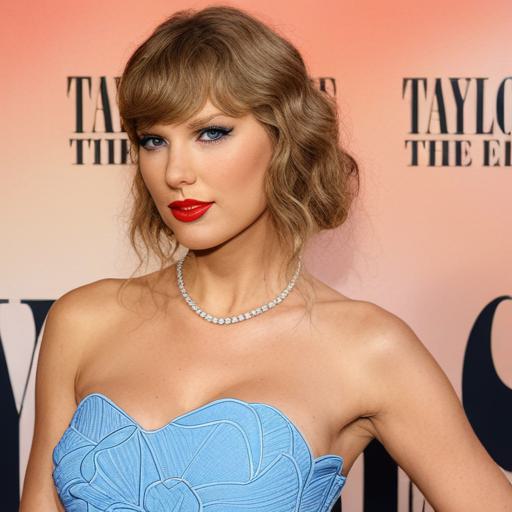} &
        \includegraphics[width=0.22\columnwidth]{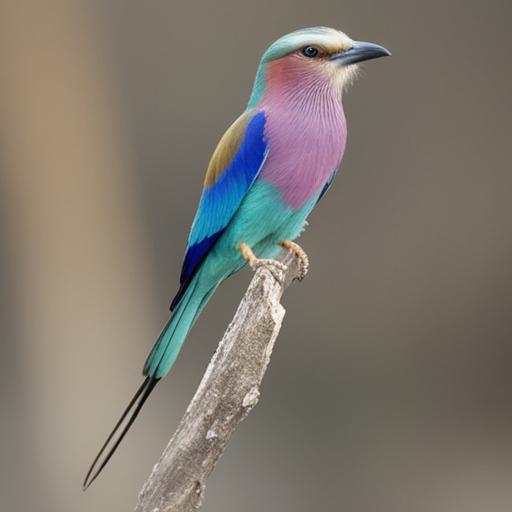} &
        \includegraphics[width=0.22\columnwidth]{images/ablation/more/coco51_full.jpg} \\
        
    \end{tabular}
\else
    \begin{tabular}{c c c c c c}
    
        \raisebox{7pt}{\rotatebox{90}{\begin{tabular}{c} Original \\ Image  \end{tabular}}} &
        \includegraphics[width=0.18\columnwidth]{images/dataset/person_3.jpg} &
        \includegraphics[width=0.18\columnwidth]{images/dataset/person_2.jpg} &
        \includegraphics[width=0.18\columnwidth]{images/dataset/bird_3.jpg} &
        \includegraphics[width=0.18\columnwidth]{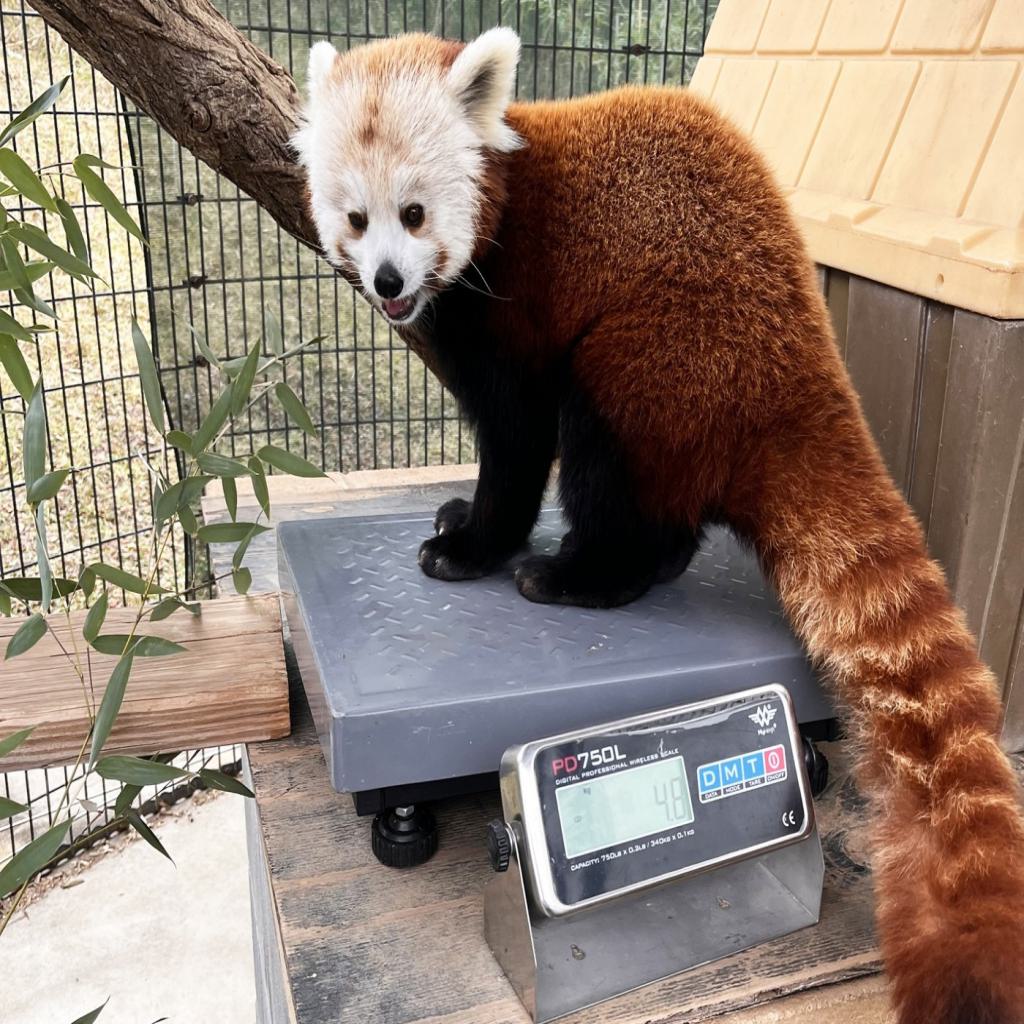} &
        \includegraphics[width=0.18\columnwidth]{images/ablation/more/coco51_origin.jpg} \\
        
        \raisebox{3pt}{\rotatebox{90}{\begin{tabular}{c} Euler \\ Inversion  \end{tabular}}} &
        \includegraphics[width=0.18\columnwidth]{images/ablation/ablation-01-only-reverse-euler/output_28_0.jpg} &
        \includegraphics[width=0.18\columnwidth]{images/ablation/ablation-01-only-reverse-euler/output_27_0.jpg} &
        \includegraphics[width=0.18\columnwidth]{images/ablation/ablation-01-only-reverse-euler/output_6_0.jpg} &
        \includegraphics[width=0.18\columnwidth]{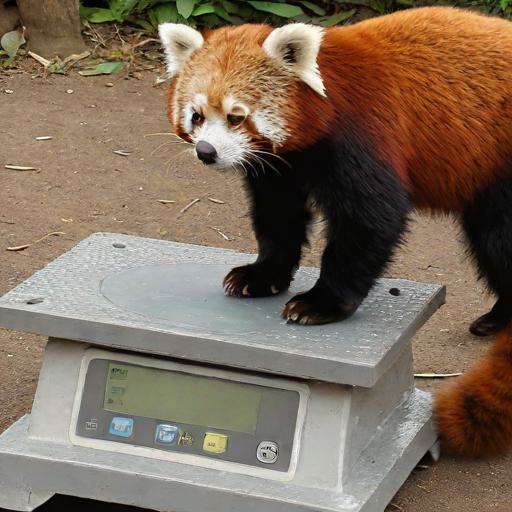} &
        \includegraphics[width=0.18\columnwidth]{images/ablation/more/coco2_inverse.jpg} \\
    
        \raisebox{2pt}{\rotatebox{90}{\begin{tabular}{c} With \\ ReNoise  \end{tabular}}} &
        \includegraphics[width=0.18\columnwidth]{images/ablation/ablation-05-9approx-wo-first/output_28_0.jpg} &
        \includegraphics[width=0.18\columnwidth]{images/ablation/ablation-05-9approx-wo-first/output_27_0.jpg} &
        \includegraphics[width=0.18\columnwidth]{images/ablation/ablation-05-9approx-wo-first/output_6_0.jpg} &
        \includegraphics[width=0.18\columnwidth]{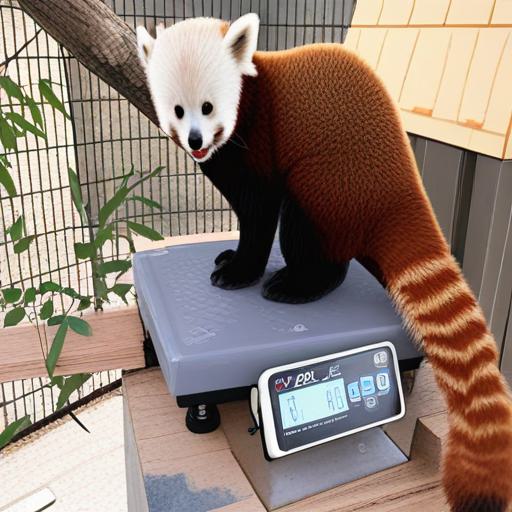} &
        \includegraphics[width=0.18\columnwidth]{images/ablation/more/coco51_9renise.jpg} \\
        
        \raisebox{2pt}{\rotatebox{90}{\begin{tabular}{c} With \\ Averaging  \end{tabular}}} &
        \includegraphics[width=0.18\columnwidth]{images/ablation/ablation-06-9approx-avg/output_28_0.jpg} &
        \includegraphics[width=0.18\columnwidth]{images/ablation/ablation-06-9approx-avg/output_27_0.jpg} &
        \includegraphics[width=0.18\columnwidth]{images/ablation/ablation-06-9approx-avg/output_6_0.jpg} &
        \includegraphics[width=0.18\columnwidth]{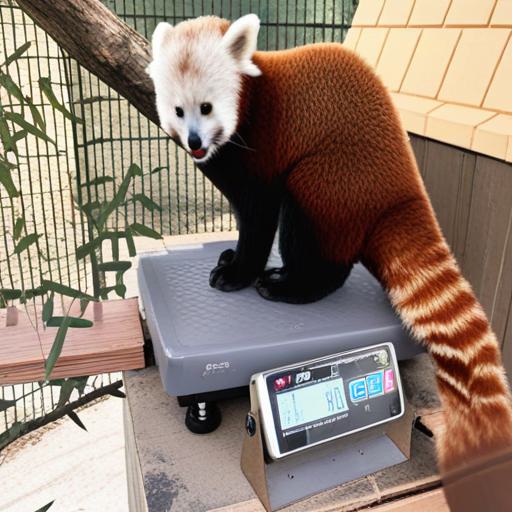} &
        \includegraphics[width=0.18\columnwidth]{images/ablation/more/coco51_avg.jpg} \\
        
        \raisebox{2pt}{\rotatebox{90}{\begin{tabular}{c} With Edit\\ Losses  \end{tabular}}} &
        \includegraphics[width=0.18\columnwidth]{images/ablation/ablation-07-9approx-edit-enhanced/output_28_0.jpg} &
        \includegraphics[width=0.18\columnwidth]{images/ablation/ablation-07-9approx-edit-enhanced/output_27_0.jpg} &
        \includegraphics[width=0.18\columnwidth]{images/ablation/ablation-07-9approx-edit-enhanced/output_6_0.jpg} &
        \includegraphics[width=0.18\columnwidth]{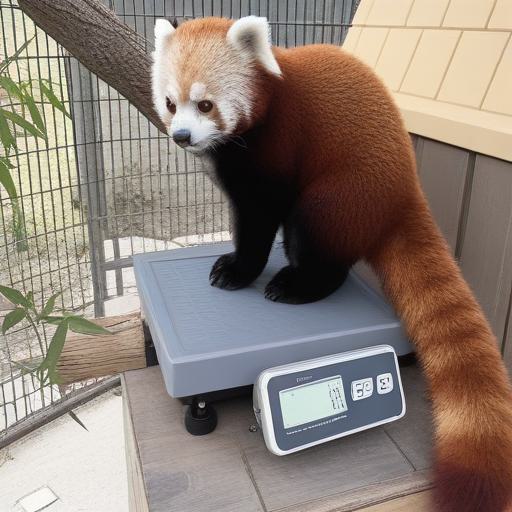} &
        \includegraphics[width=0.18\columnwidth]{images/ablation/more/coco51_w_loss.jpg} \\
        
        \raisebox{2pt}{\rotatebox{90}{\begin{tabular}{c} With Noise\\ Correction  \end{tabular}}} &
        \includegraphics[width=0.18\columnwidth]{images/ablation/ablation-08-9approx-optimize-epsilon/output_28_0.jpg} &
        \includegraphics[width=0.18\columnwidth]{images/ablation/ablation-08-9approx-optimize-epsilon/output_27_0.jpg} &
        \includegraphics[width=0.18\columnwidth]{images/ablation/ablation-08-9approx-optimize-epsilon/output_6_0.jpg} &
        \includegraphics[width=0.18\columnwidth]{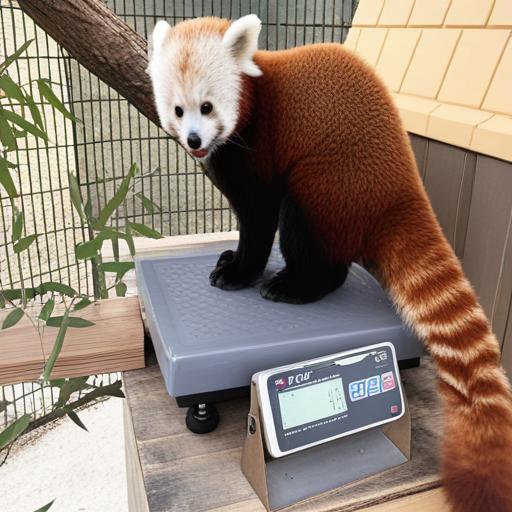} &
        \includegraphics[width=0.18\columnwidth]{images/ablation/more/coco51_full.jpg} \\
        
    \end{tabular}
\fi

}
\vspace{-0.225cm}    
\caption{
Ablation study on SDXL Turbo. The first row presents the input image. In each subsequent row, we show the reconstruction results using an additional component of our inversion method. The images in the bottom row represent the results obtained by our full method.
\ifeccv
    NC stands for Noise Correction.
\else
    \vspace{-15pt}
\fi
}
\label{fig:ablation}

%% file: figures/quan_comapre_unet_ops.tex
\begin{figure}
    \centering
    \setlength{\tabcolsep}{1pt}
    \addtolength{\belowcaptionskip}{-10pt}

\input{figures/quan_comapre_unet_ops_content}
    \vspace{-0.15cm}
\end{figure}

%% file: tables/unet_ops_budget_table.tex
\begin{table}
\small
\centering
\setlength{\tabcolsep}{2.5pt}
\input{tables/unet_ops_budget_table_content}
\vspace{-0.1cm}
\end{table}

%% file: figures/lcm_results_figure.tex
\begin{figure} [t]
    \centering
    \setlength{\tabcolsep}{1pt}
    \addtolength{\belowcaptionskip}{-10pt}
    {\scriptsize
    \centering

    \ifeccv
    \begin{tabular}{c ccc c ccc}
        {\begin{tabular}{c} Original \\ Image  \end{tabular}} & 
        \multicolumn{3}{c}{$\longleftarrow$ Editing Results $\longrightarrow$} &
        {\begin{tabular}{c} Original \\ Image  \end{tabular}} & 
        \multicolumn{3}{c}{$\longleftarrow$ Editing Results $\longrightarrow$} \\
        
        \includegraphics[width=0.12\columnwidth]{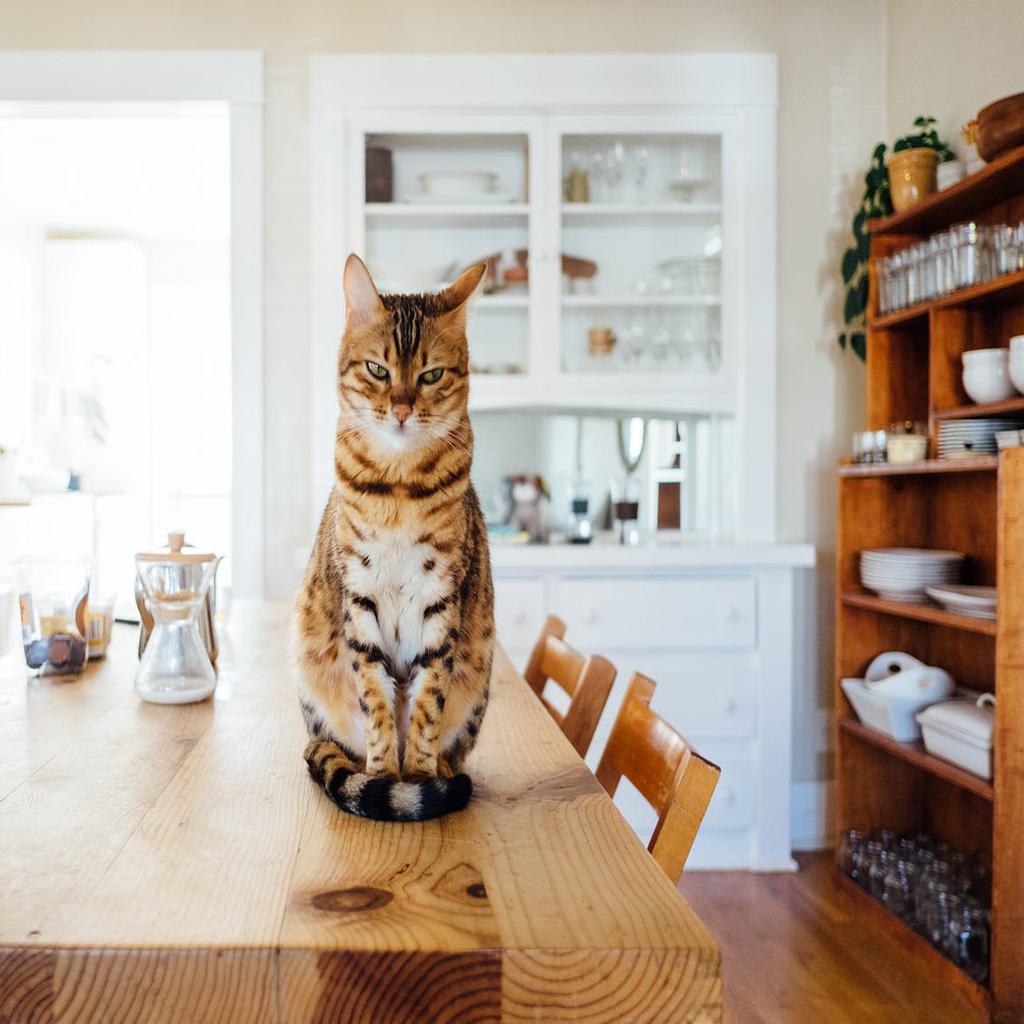} &
        \includegraphics[width=0.12\columnwidth]{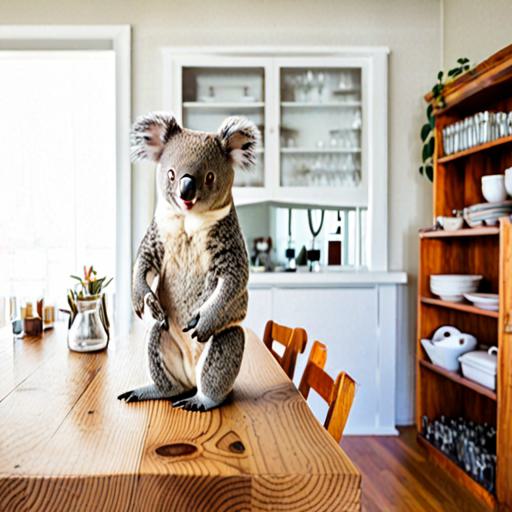} &
        \includegraphics[width=0.12\columnwidth]{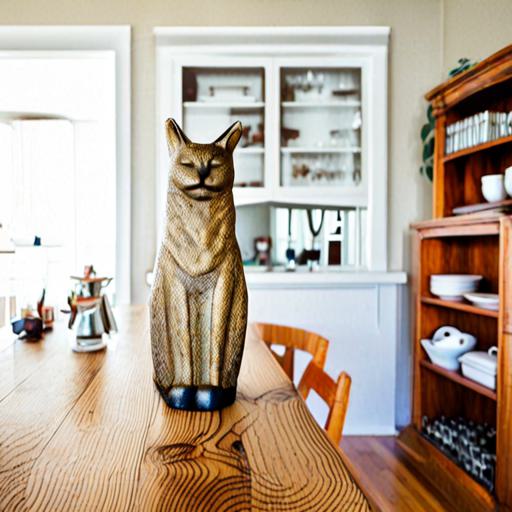} &
        \includegraphics[width=0.12\columnwidth]{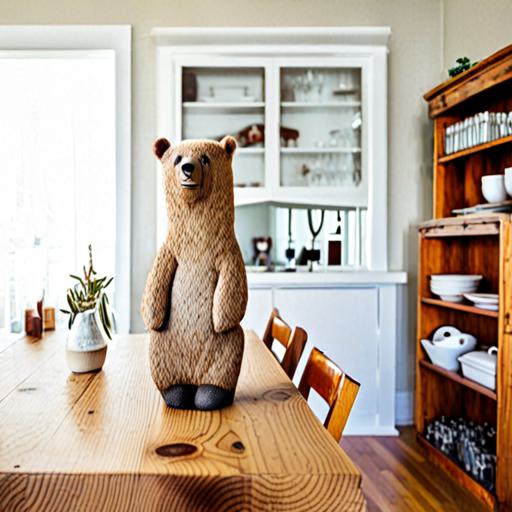} &
        
        \includegraphics[width=0.12\columnwidth]{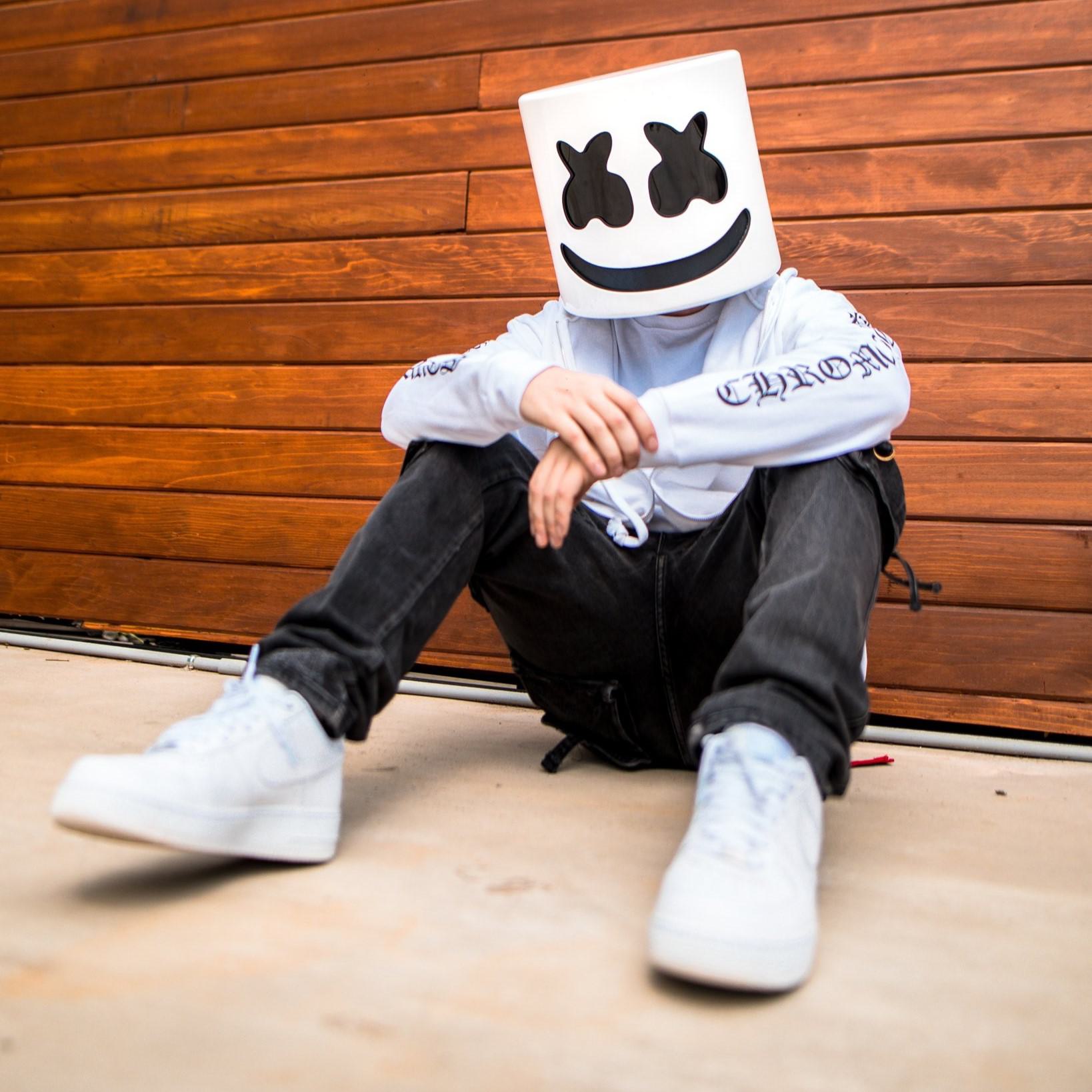} &
        \includegraphics[width=0.12\columnwidth]{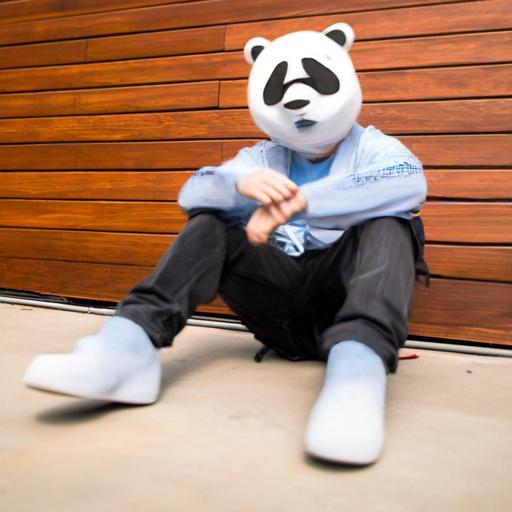} &
        \includegraphics[width=0.12\columnwidth]{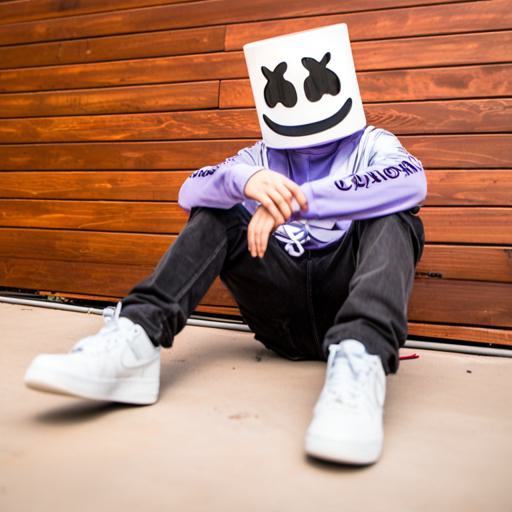} &
        \includegraphics[width=0.12\columnwidth]{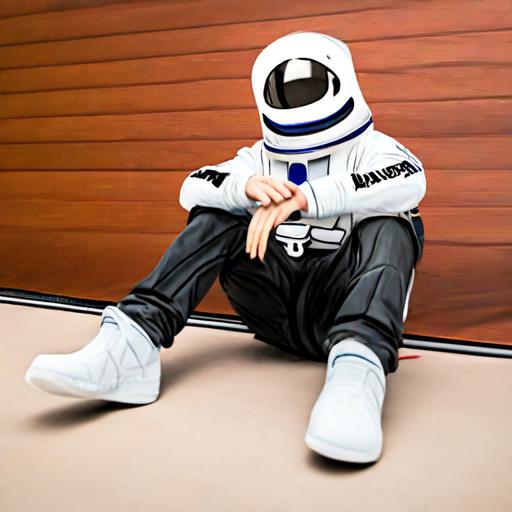} \\

        {``cat''} &
        {``koala''} &
        {``cat statue''} &
        {``bear''} &

        {``person''} &
        {\begin{tabular}{c} ``panda \\ mask'' \end{tabular}} &
        {\begin{tabular}{c} ``purple \\ shirt'' \end{tabular}} &
        {``astronaut''} \\
    \end{tabular}
    \else
    \begin{tabular}{c ccc}
        {Original Image } & 
        \multicolumn{3}{c}{$\longleftarrow$ Editing Results $\longrightarrow$} \\
        
        \includegraphics[width=0.24\columnwidth]{images/dataset-2/11-cat.jpg} &
        \includegraphics[width=0.24\columnwidth]{images/lcm_results/cat/koala.jpg} &
        \includegraphics[width=0.24\columnwidth]{images/lcm_results/cat/cat_statue.jpg} &
        \includegraphics[width=0.24\columnwidth]{images/lcm_results/cat/bear.jpg} \\

        {``cat''} &
        {``koala''} &
        {``cat statue''} &
        {``bear''} \\
        
        \includegraphics[width=0.24\columnwidth]{images/dataset-2/36-marshmello.jpg} &
        \includegraphics[width=0.24\columnwidth]{images/lcm_results/marshmello/panda.jpg} &
        \includegraphics[width=0.24\columnwidth]{images/lcm_results/marshmello/purple_shirt.jpg} &
        \includegraphics[width=0.24\columnwidth]{images/lcm_results/marshmello/astro3.jpg} \\

        {``person''} &
        {``panda mask''} &
        {``purple shirt''} &
        {``astronaut''} \\

    \end{tabular}
    \fi

    }
    \vspace{-0.245cm}    
    \caption{  
    LCM Editing Results.
    \ifeccv
    We showcase two examples of real images, each followed by three edits.
    \else
    Each row showcases one image. The leftmost image is the original, followed by three edited versions. 
    \fi
    The text below each edited image indicates the specific word or phrase replaced or added to the original prompt for that specific edit.
    \ifeccv
    \else
        \vspace{-13pt}
    \fi
    }
    \label{fig:lcm_results}
    \ifeccv
        \vspace{-10pt}
    \fi
\end{figure}

%% file: figures/ablation.tex
\begin{figure}
    \centering
    \setlength{\tabcolsep}{1pt}
    \addtolength{\belowcaptionskip}{-10pt}

\input{figures/ablation_content}
    \vspace{-0.15cm}
\end{figure}

%% file: tables/ablation_table.tex
\begin{table}
\small
\centering
\setlength{\tabcolsep}{2.5pt}
\input{tables/ablation_table_content}
\vspace{-0.1cm}
\end{table}

%% file: figures/turbo_compare_ddpm.tex
\begin{figure} [t]
    \centering
    \setlength{\tabcolsep}{1pt}
    \addtolength{\belowcaptionskip}{-10pt}
    \input{figures/turbo_compare_ddpm_content}

    \vspace{-0.15cm}
\end{figure}

%% file: figures/turbo_compare_ddpm_content.tex
{\scriptsize

    \ifeccv
        \begin{tabular}{c cccc}
        \hspace{-8px} \raisebox{12pt}{\rotatebox{90}{Ours}} &
        \includegraphics[width=0.24\textwidth]{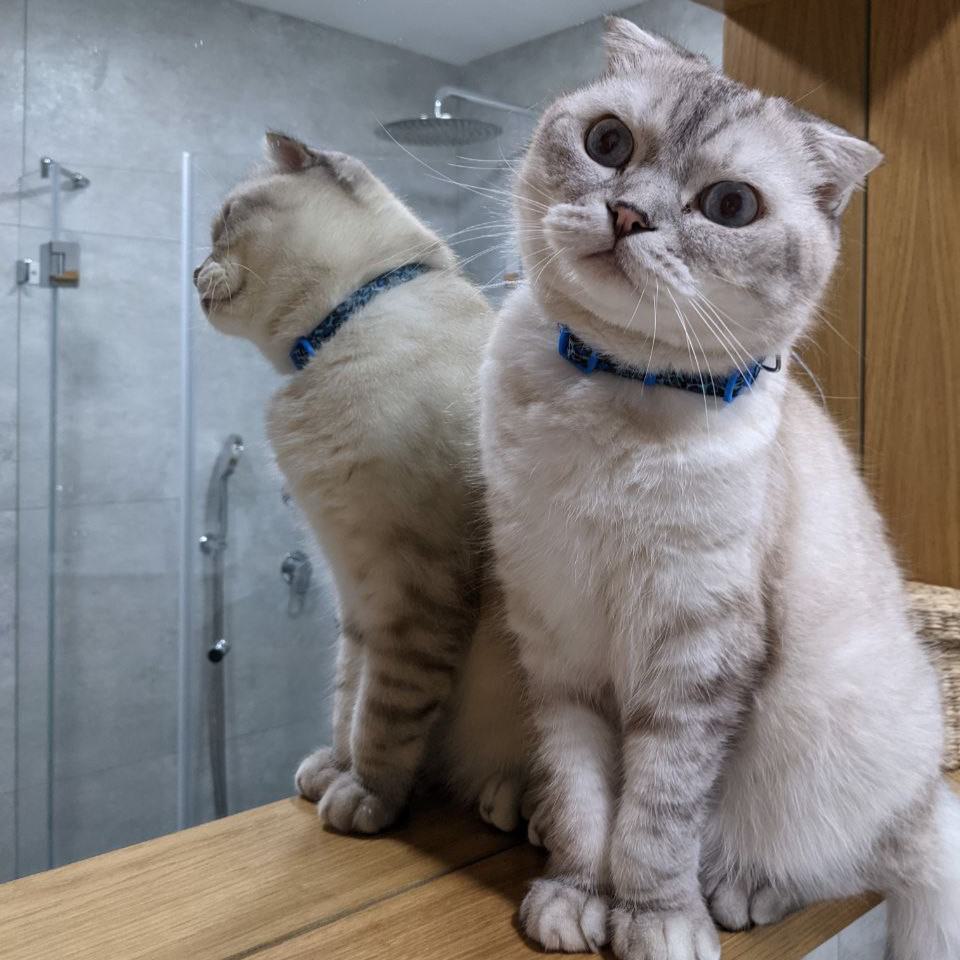} &
        \includegraphics[width=0.24\textwidth]{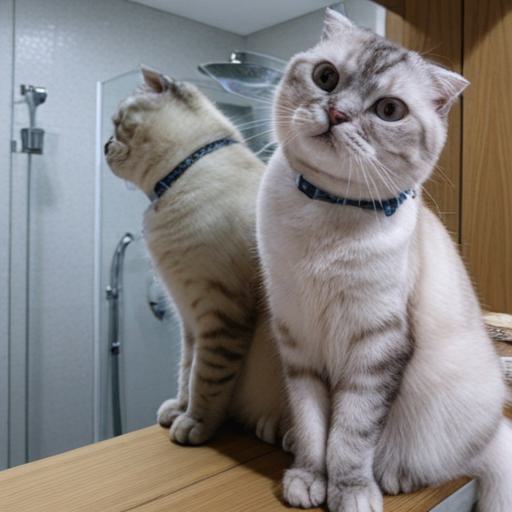} &
        \includegraphics[width=0.24\textwidth]{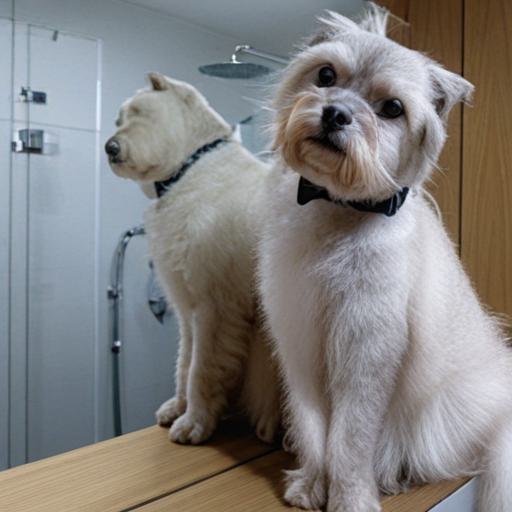} &
        \includegraphics[width=0.24\textwidth]{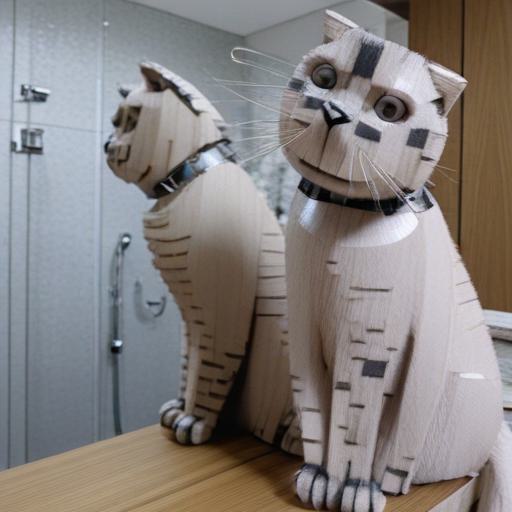} \\
        \hspace{-8px} \raisebox{6pt}{\rotatebox{90}{Friendly}} &
        \includegraphics[width=0.24\textwidth]{images/dataset-2/02-gnochi_mirror_sq.jpg} &
        \includegraphics[width=0.24\textwidth]{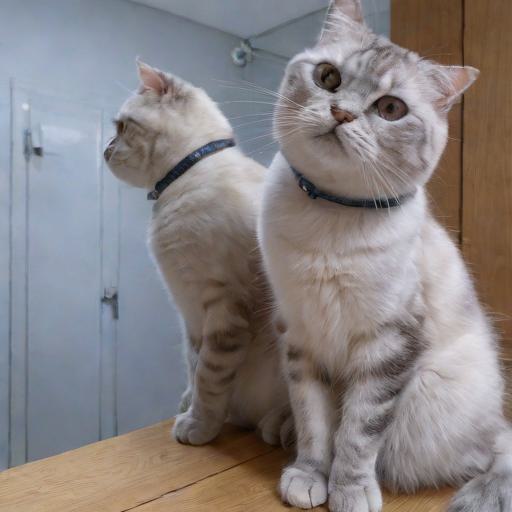} &
        \includegraphics[width=0.24\textwidth]{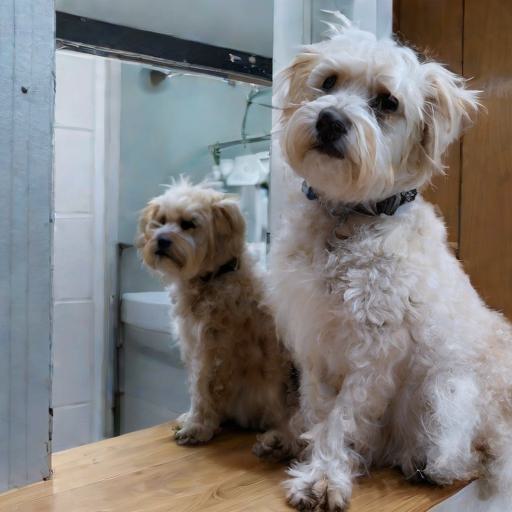} &
        \includegraphics[width=0.24\textwidth]{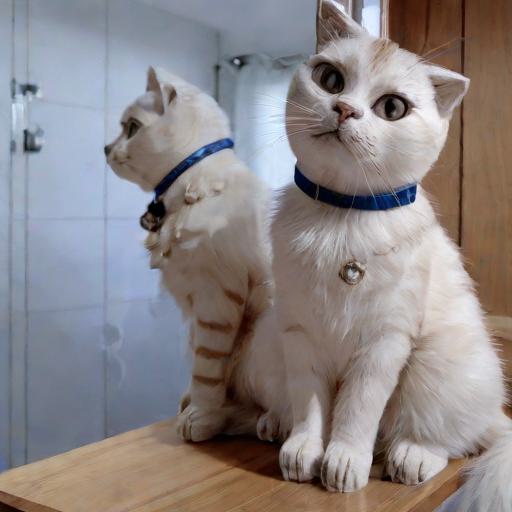} \\
        & Input & {\begin{tabular}{c} Recon- \\ struction  \end{tabular}} & {\begin{tabular}{c} ``cat'' $\rightarrow$ \\ ``dog''   \end{tabular}} & {\begin{tabular}{c} ``wooden \\  cat''  \end{tabular}} \\
    \end{tabular}
    \else
        \begin{tabular}{c c c c c}
            \raisebox{18pt}{\rotatebox{90}{Ours}} &
            \includegraphics[width=0.23\columnwidth]{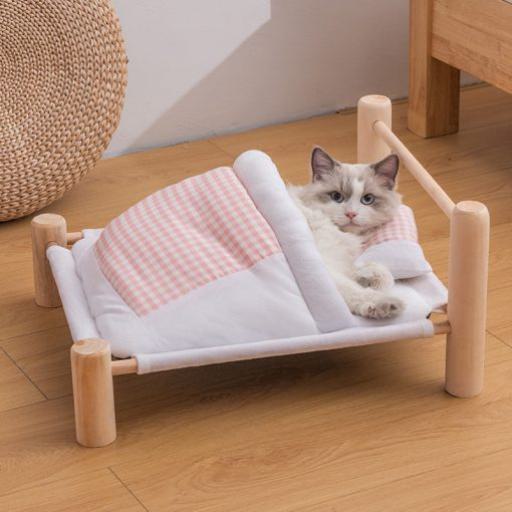} &
            \includegraphics[width=0.23\columnwidth]{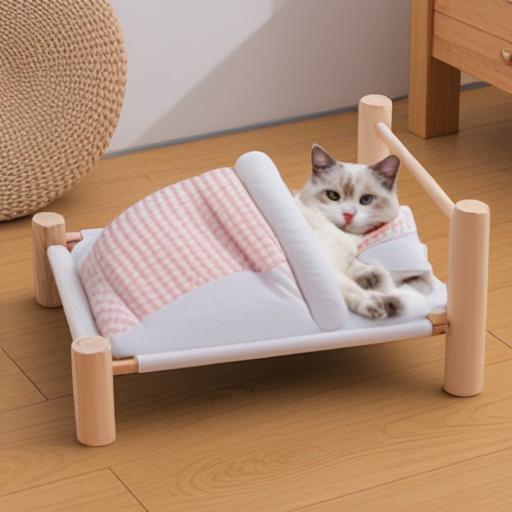} &
            \includegraphics[width=0.23\columnwidth]{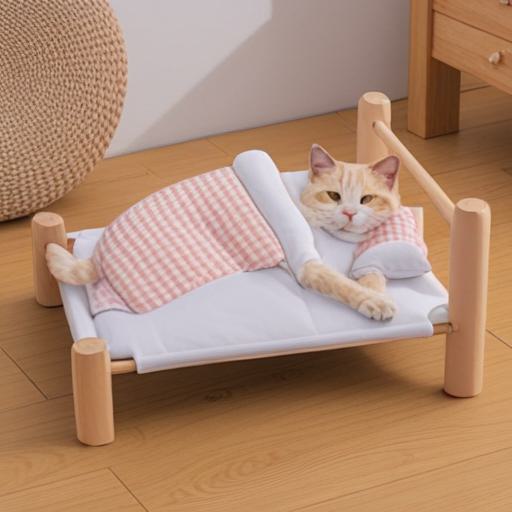} &
            \includegraphics[width=0.23\columnwidth]{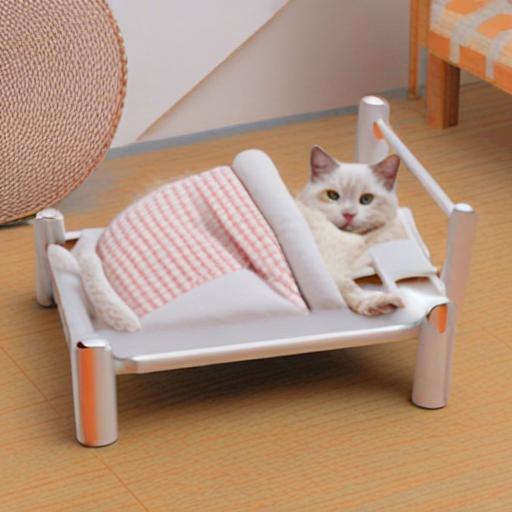} \\
            \raisebox{8pt}{\rotatebox{90}{Edit Friendly}} &
            \includegraphics[width=0.23\columnwidth]{images/dataset-2/21-cat-in-bed.jpg} &
            \includegraphics[width=0.23\columnwidth]{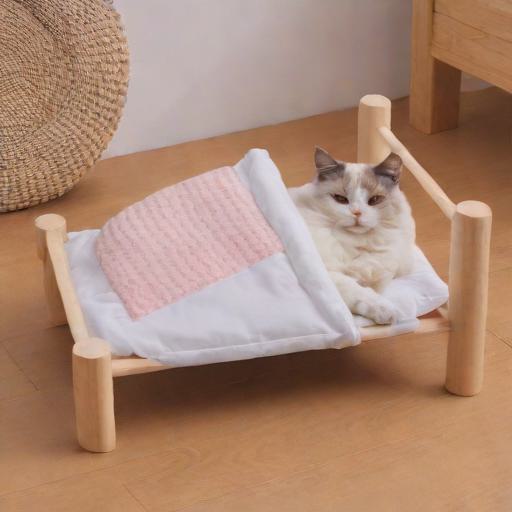} &
            \includegraphics[width=0.23\columnwidth]{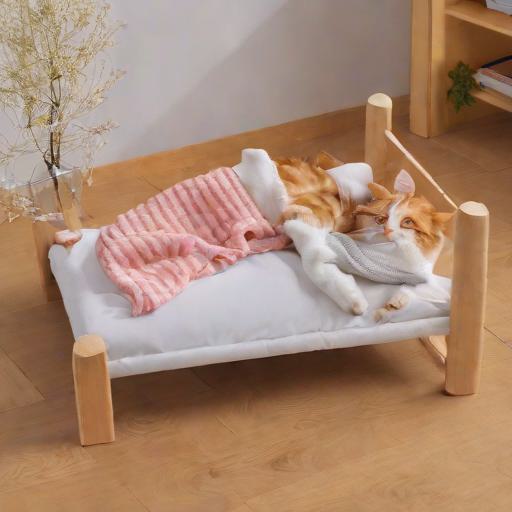} &
            \includegraphics[width=0.23\columnwidth]{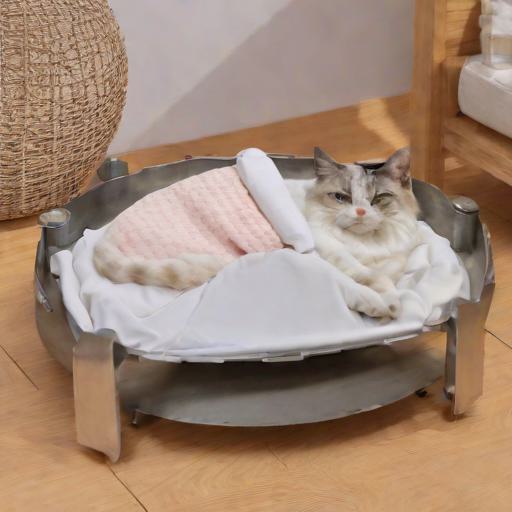} \\
            & Original & Reconstruction & ``ginger cat'' & ``wood'' $\rightarrow$ ``metal'' \\
            \raisebox{18pt}{\rotatebox{90}{Ours}} &
            \includegraphics[width=0.23\columnwidth]{images/dataset-2/02-gnochi_mirror_sq.jpg} &
            \includegraphics[width=0.23\columnwidth]{images/turbo_compare_ddpm/our/gnochi_recon.jpg} &
            \includegraphics[width=0.23\columnwidth]{images/turbo_compare_ddpm/our/gnochi_dog.jpg} &
            \includegraphics[width=0.23\columnwidth]{images/turbo_compare_ddpm/our/gnochi_wooden.jpg} \\
            \raisebox{8pt}{\rotatebox{90}{Edit Friendly}} &
            \includegraphics[width=0.23\columnwidth]{images/dataset-2/02-gnochi_mirror_sq.jpg} &
            \includegraphics[width=0.23\columnwidth]{images/turbo_compare_ddpm/ddpm/gnochi_recon.jpg} &
            \includegraphics[width=0.23\columnwidth]{images/turbo_compare_ddpm/ddpm/gnochi_dog.jpg} &
            \includegraphics[width=0.23\columnwidth]{images/turbo_compare_ddpm/ddpm/gnochi_wooden.jpg} \\
            & Original & Reconstruction & ``cat'' $\rightarrow$ ``dog'' & ``wooden cat''
        \end{tabular}
    \fi
    
    }
    \ifeccv
        \vspace{-0.325cm}
    \else
        \vspace{-0.175cm}
    \fi
    \caption{
    \ifeccv
    Comparison with edit-friendly with SDXL Turbo. We inverted the image with the prompt ``a cat is sitting in front of a mirror'' and applied edits.
    \else
    Comparison with edit-friendly DDPM Inversion with SDXL Turbo. We invert two images with the prompts: ``a cat laying in a bed made out of wood'' (left) and ``a cat is sitting in front of a mirror'' (right) and apply two edits to each image.
        \vspace{-10pt}
    \fi 
    }
    \label{fig:turbo_compare_ddpm}

%% file: figures/mini_pages/exp_fig_eccv2.tex
\begin{figure}[t]
    \centering
    \scriptsize
    \begin{minipage}[b]{0.48\textwidth}
        \centering
        \input{figures/turbo_compare_ddpm_content}
    \end{minipage}
    \hfill
    \begin{minipage}[b]{0.48\textwidth}
        \centering
        \input{figures/reconstruction_comparisons_figure_content}
    \end{minipage}
    \vspace{-20pt}
\end{figure}

%% file: figures/reconstruction_comparisons_figure_content.tex
{\scriptsize
    \centering

    \ifeccv
        \begin{tabular}{cccc}            
            \includegraphics[width=0.24\textwidth]{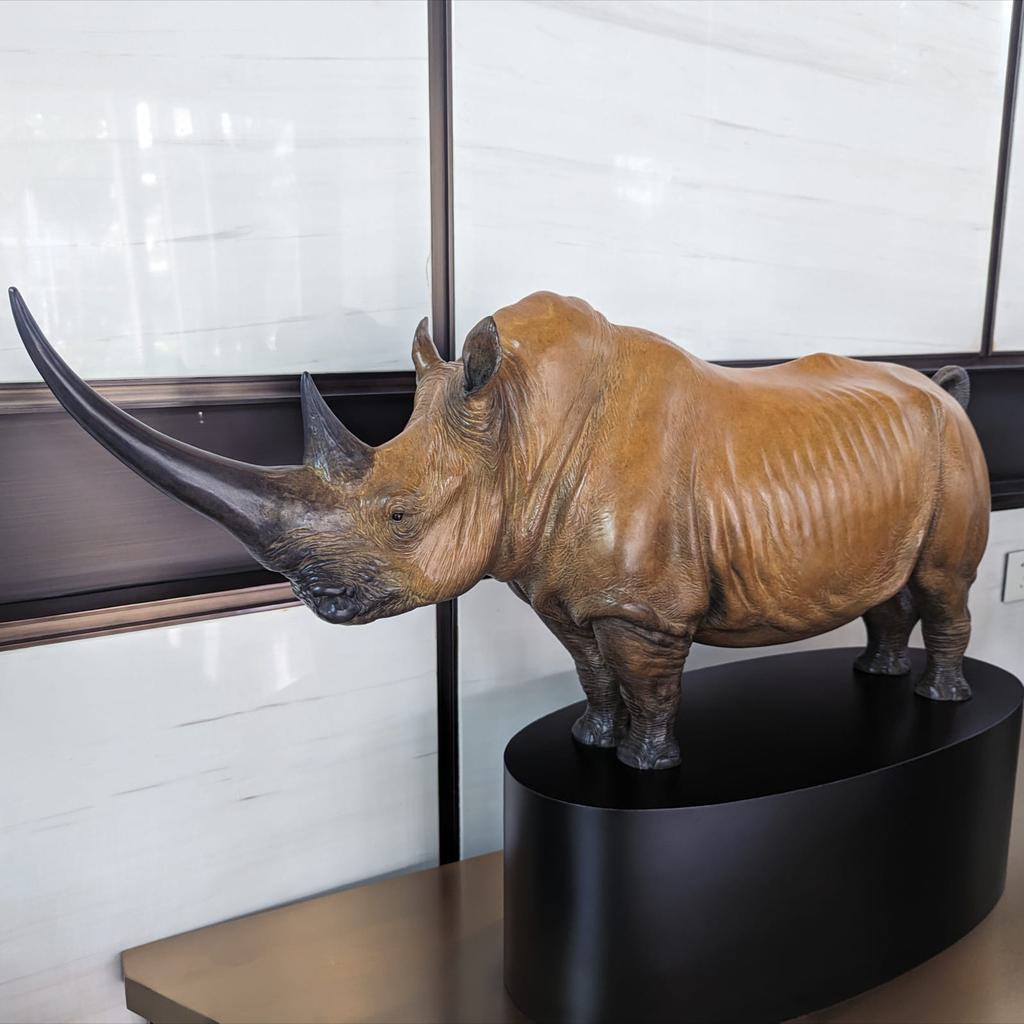} &
            \includegraphics[width=0.24\textwidth]{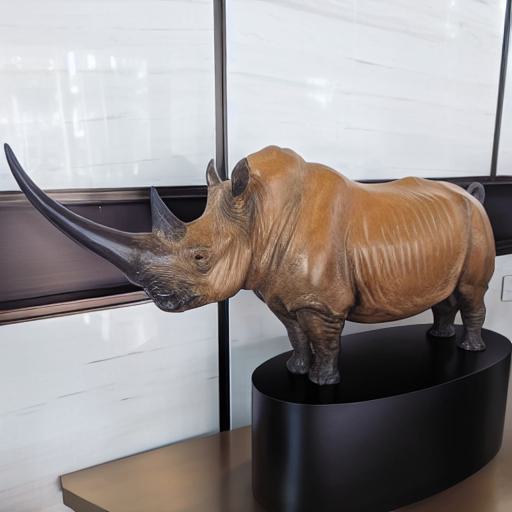} &
            \includegraphics[width=0.24\textwidth]{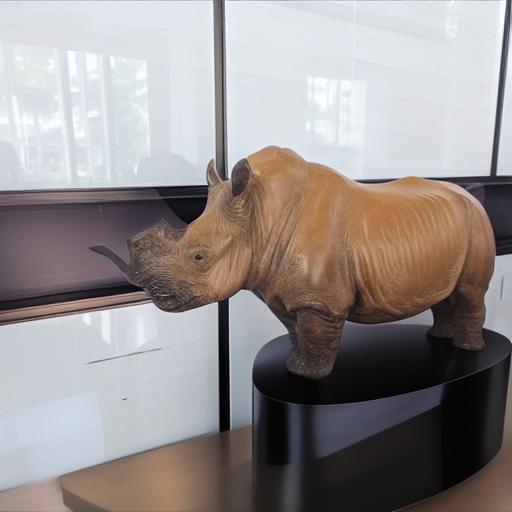} &
            \includegraphics[width=0.24\textwidth]{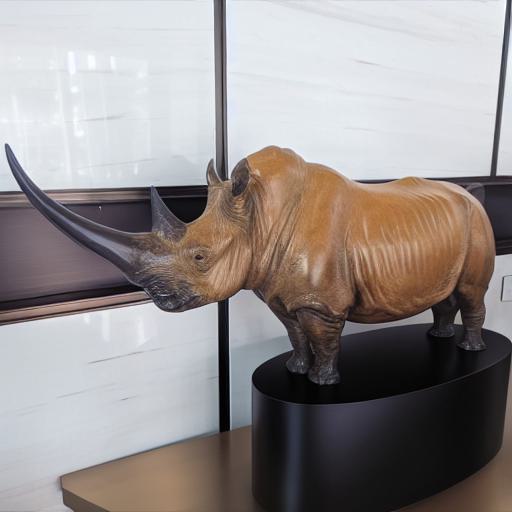} \\
            
            \includegraphics[width=0.24\textwidth]{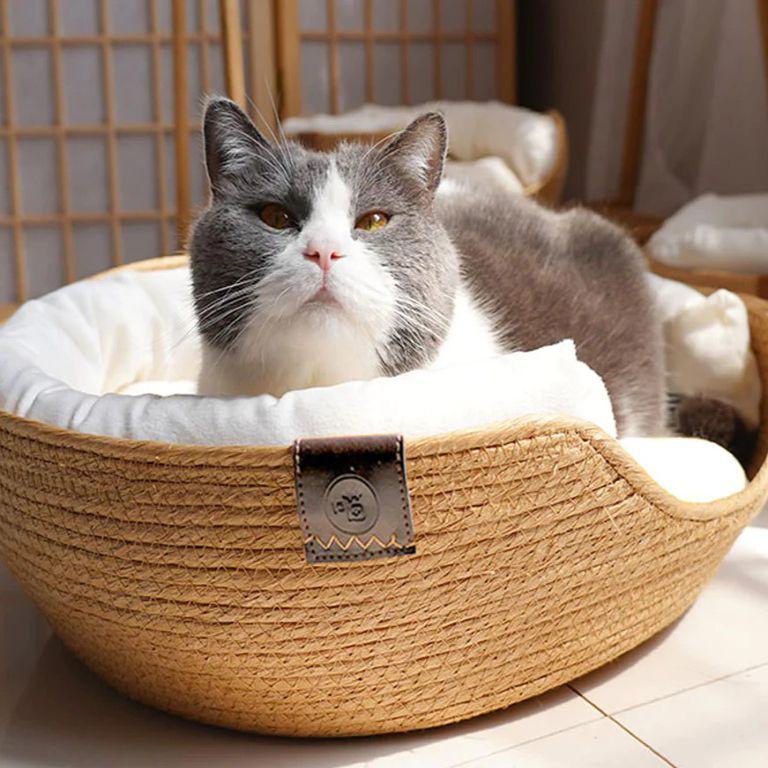} &
            \includegraphics[width=0.24\textwidth]{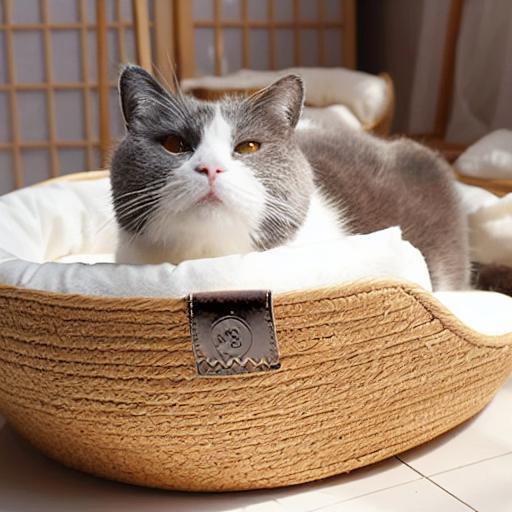} &
            \includegraphics[width=0.24\textwidth]{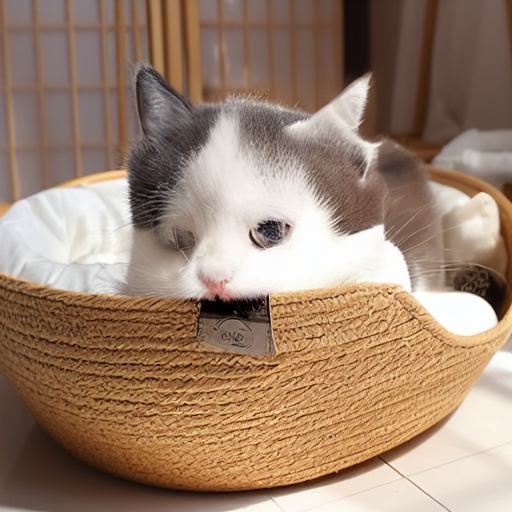} &
            \includegraphics[width=0.24\textwidth]{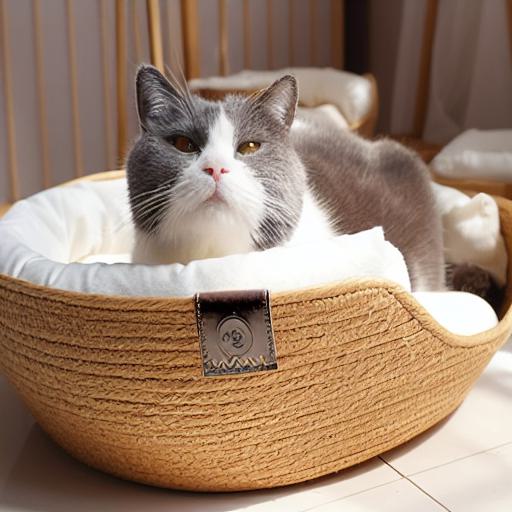} \\

            \begin{tabular}{c} Original \\ Image  \end{tabular} &
            \begin{tabular}{c} NTI \end{tabular} &
            \begin{tabular}{c} NPI  \end{tabular} &
            \begin{tabular}{c} DDIM w/ \\ ReNoise  \end{tabular} \\
        \end{tabular}
    \else
        \begin{tabular}{c c c c}

            \includegraphics[width=0.24\columnwidth]{images/dataset-2/12-rhino.jpg} &
            \includegraphics[width=0.24\columnwidth]{images/reconstruction_comparisions/NTI/11.jpg} &
            \includegraphics[width=0.24\columnwidth]{images/reconstruction_comparisions/NPI/output_11.jpg} &
            \includegraphics[width=0.24\columnwidth]{images/reconstruction_comparisions/OUR/output_11.jpg} \\

            \includegraphics[width=0.24\columnwidth]{images/dataset-2/33-kitten.jpg} &
            \includegraphics[width=0.24\columnwidth]{images/reconstruction_comparisions/NTI/32.jpg} &
            \includegraphics[width=0.24\columnwidth]{images/reconstruction_comparisions/NPI/output_32.jpg} &
            \includegraphics[width=0.24\columnwidth]{images/reconstruction_comparisions/OUR/output_32.jpg} \\
            
            \begin{tabular}{c} Original \\ Image  \end{tabular} &
            \begin{tabular}{c} Null Text \\ Inversion  \end{tabular} &
            \begin{tabular}{c} Negative \\ Prompt Inv  \end{tabular} &
            \begin{tabular}{c} DDIM inv \\ w/ ReNoise  \end{tabular} \\
        \end{tabular}
    \fi
    
}
\ifeccv
    \vspace{-0.325cm}
\else
    \vspace{-0.15cm}
\fi
\ifeccv
    \caption{
    Image reconstruction comparisons with SD. While NTI and our method achieve comparable results, ours demonstrates significant speed improvement.
    }
\else
    \caption{
    Image reconstruction comparisons with Stable Diffusion. We present the results of Null-Text Inversion (NTI), Negative-Prompt Inversion (NPI), and our method.
    While NTI and our method achieve comparable results, ours demonstrates significant speed improvement.
    \vspace{-15pt}
    }
\fi
\label{fig:reconstruction_comparisons_figure}

%% file: figures/reconstruction_comparisons_figure.tex
\begin{figure}[t]
    \centering
    \setlength{\tabcolsep}{1pt}
    \addtolength{\belowcaptionskip}{-10pt}

\input{figures/reconstruction_comparisons_figure_content}
    \vspace{-0.15cm}
\end{figure}

%% file: 7-conclusion.tex
\ifeccv
    \vspace{-10pt}
\fi

\section{Conclusion} \label{sec:conclusion}

\ifeccv
    \vspace{-5pt}
\fi

In this work, we have introduced ReNoise, a universal approach that enhances various inversion algorithms of diffusion models. ReNoise gently guides the inversion curve of a real image towards the source noise from which a denoising process reconstructs the image.
ReNoise can be considered as a meta-algorithm that warps the trajectory of any iterative diffusion inversion process. 
Our experiments demonstrate that averaging the last few renoising iterations significantly enhances reconstruction quality.
For a fixed amount of computation, ReNoise shows remarkably higher reconstruction quality and editability. The method is theoretically supported and our experiments reconfirm its effectiveness on a variety of diffusion models and sampling algorithms. Moreover, the method is numerically stable, and always converges to some inversion trajectory that eases hyperparameters adjustment.

Beyond the net introduction of an effective inversion, the paper presents a twofold important contribution: an effective inversion for few-steps diffusion models, which facilitates effective editing on these models.

A limitation of ReNoise is the model-specific hyperparameter tuning required for Edit Enhancement and Noise Correction. While these hyperparameters remain stable for a given model, they may vary across models, and tuning them is necessary to achieve high reconstruction quality while maintaining editability.

While ReNoise demonstrates the potential for editing few-step diffusion models, more extensive testing with advanced editing methods is needed.  It is worth noting that no such editing has been demonstrated for the few-step diffusion models.
We believe and hope that our ReNoise method will pave the way for fast and effective editing methods based on the few-steps models. 
We also believe that ReNoise can be adapted to the challenging problem of inverting video-diffusion models.

%% file: 8-sup.tex
\input{8.1-Sup-theory}

\ifeccv
\else
    \input{figures/sup_turbo_results}
    \input{figures/lpips_unet_op_figure}
    \vspace{-5pt}
\fi

\section{Implementation Details}
\label{sec:imp_details}

We used BLIP-2~\cite{li2023blip} to generate captions for the input images, which were then used as prompts for the diffusion models.
\ifeccv
In the following table
\else
In Table~\ref{tb:impl_details}
\fi
, we provide all hyperparameters of ReNoise inversion per model, optimized for the best reconstruction-editability trade-off:

\ifeccv
    \input{tables/imp_details}
\fi

Where $\{w_i\}$ are the renoising estimations averaging weights, and $\lambda_{\text{pair}}$ and $\lambda_{\text{patch-KL}}$ are the weights we assign to each component of the edit enhancement loss:
\begin{equation*}
\mathcal{L}_{\text{edit}} = \lambda_{\text{pair}} \cdot \mathcal{L}_{\text{pair}} + \lambda_{\text{patch-KL}} \cdot \mathcal{L}_{\text{patch-KL}} 
\end{equation*}

\input{8.2-Sup-experiments}

%% file: 8.1-Sup-theory.tex
\section{Convergence Discussion}
\label{sec:sup_theory}

\ifeccv
    \subsection{Inversion process as backward Euler}
    
    The denoising process of diffusion models can be mathematically described as solving an ordinary differential equation (ODE). A common method for solving such equations is the Euler method, which takes small steps to approximate the solution.
    For ODE in the form of $y^{'}(t) = f(t, y(t))$, Euler solution will be:
    \begin{equation*}
        y_{n+1} = y_n + h\cdot f(t_n, y_n),    
    \end{equation*}
    where $h$ is the step size.
    The inversion process can be described as solving ODE using the backward Euler method (or implicit Euler method)~\cite{doi:https://doi.org/10.1002/0470868279.ch2}.
    This method is similar to forward Euler, with the difference that $y_{n+1}$ appears on both sides of the equation:
    \begin{equation*}
        y_{n+1} = y_n + h\cdot f(t_{n+1}, y_{n+1}).    
    \end{equation*}
    
    For equations lacking an algebraic solution, several techniques estimate $y_{n+1}$ iteratively.
    As we described in Section 3.1 (main paper), the inversion process lacks a closed-form solution, as shown in Equation 2. To address this, the ReNoise method leverages the Picard iteration method to progressively refine the estimate of $y_{n+1}$:
    \begin{equation*}
        \frac{\partial y_{n+1}^{(k+1)}}{\partial t} =  f(t_{n+1}, y_{n+1}^{(k)}).
    \end{equation*}
    
    While the naive approach might not perfectly converge to the true solution, averaging the estimates of the last iterations mitigates these errors, leading to improvement in the reconstruction process.
    
    \begin{figure}
        \centering
        \includegraphics[width=0.9\textwidth]{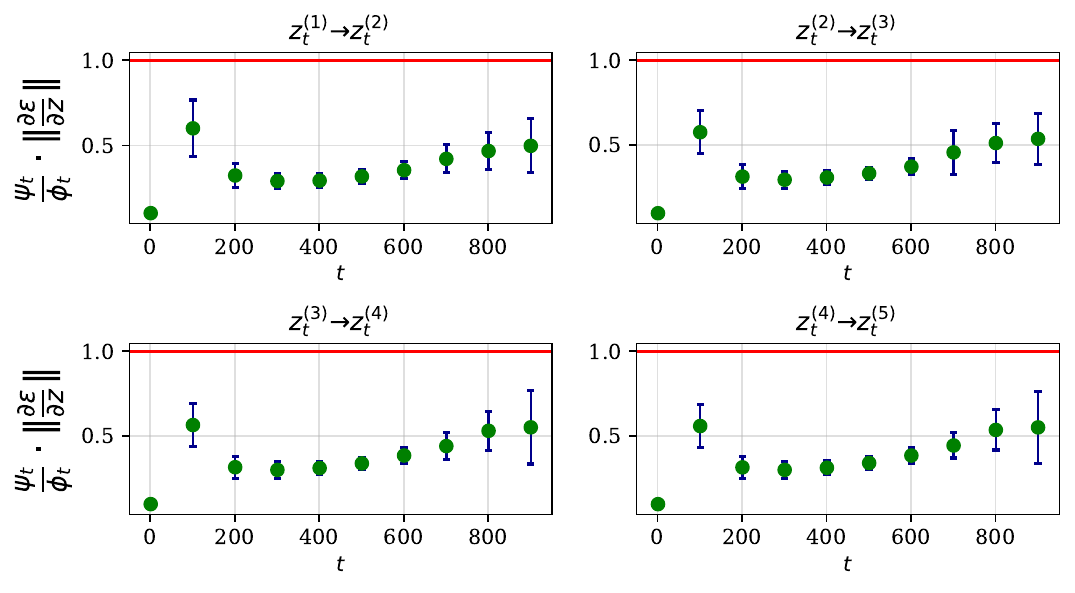}
        \caption{Scaled Jacobian norm of the mapping $z^{(k)}_{t} \to z^{(k)}_{t}$ calculated for various noise levels $t$, iterations $k$. We report the average and the standard deviation calculated over 32 images. Values below 1 indicate the exponential convergence of the ReNoise algorithm.}
        \label{fig:jacobian_norms}
    \end{figure}

\fi

\subsection{ReNoise As Contraction mapping}

\ifeccv
In this section, we discuss the proposed method from a convergence perspective. For completeness we will present the entire discussion, including parts from the main paper.
\else

    In this section, we continue the convergence discussion of our method.

\ifeccv
    \paragraph{\textbf{Toy Example.}}
    We begin with the simple toy example, the diffusion of a shifted Gaussian.
    Given the initial distribution $\mu_0 \sim \mathcal{N}(a, I)$, where $a$ is a non-zero shift value and $I$ is the identity matrix. The diffusion process defines the family of distributions $\mu_t \sim \mathcal{N}(ae^{-t}, I)$, and the probability flow ODE takes the form $\frac{dz}{dt} = -ae^{-t}$ (see \cite{khrulkov2022understanding} for details). The Euler solver step at a state $(z_t, t)$, and time step $\Delta t$ moves it to $(z_{t+\Delta t}^{(1)}, t + \Delta t) = (z_t - ae^{-t} \cdot \Delta t, t+ \Delta t)$.
    Notably, the backward Euler step at this point does not lead us to $z_t$.
    After applying the first renoising iteration, we get $(z^{(2)}_{t + \Delta t}, t + \Delta t) = (z_t - ae^{-(t + \Delta t)} \cdot \Delta t, t + \Delta t)$ and the backward Euler step at this point leads exactly to $(z_t, t)$.
    Thus, in this simple example, the ReNoise algorithm successfully estimates the exact pre-image after a single step.
    While we cannot guarantee that in the general case, we will discuss some sufficient conditions for the algorithm's convergence and empirically verify them for the image diffusion model.
    
    \paragraph{\textbf{ReNoise Convergence.}}
    During the inversion process, we aim to find the next noise level inversion, denoted by $\hat{z}_{t}$, such that applying the denoising step to $\hat{z}_{t}$ recovers the previous state, $z_{t-1}$.
    Given the noise estimation $\epsilon_{\theta}(z_{t}, t)$ and fixed $z_{t-1}$, the ReNoise mapping defined in Section 3 in the main paper can be written as $\mathcal{G}: z_{t} \to \mathrm{InverseStep}(z_{t-1}, \epsilon_{\theta}(z_{t}, t))$.
    For example, in the case of using DDIM sampler the mapping is $\mathcal{G}(z_{t}) = \frac{1}{\phi_t}(z_{t-1} - \psi_t \epsilon_{\theta}(z_{t}, t))$.
    The point $\hat{z}_{t}$, which maps after the denoising step to $z_{t-1}$, is a stationary point of this mapping. 
    Given $z_{t}^{(1)}$, the first approximation of the next noise level $z_{t}$, our goal is to show that the sequence $z_{t}^{(k)} = \mathcal{G}^{k-1}(z_{t}^{(1)}), k \to \infty$ converges. As the mapping $\mathcal{G}$ is continuous, the limit point would be its stationary point. The definition of $\mathcal{G}$ gives us
    \begin{equation*}
    \| z_{t}^{(k + 1)} - z_{t}^{(k)} \| = \|\mathcal{G}(z_{t}^{(k)}) - \mathcal{G}(z_{t}^{(k - 1)})\|,
    \end{equation*}
    where the norm is always assumed as the $l_2$-norm. For the ease of the notations, let us define $\Delta^{(k)} = z_{t}^{(k)} - z_{t}^{(k - 1)}$. For convergence proof it is sufficient to show that the sum on norms of these differences converges which will imply that $z_{t}^{(k)}$ is the Cauchy sequence. Below we check that in practice $\|\Delta^{(k)}\|$ decreases exponentially as $k \to \infty$ and thus has finite sum.
    In the assumption that $\mathcal{G}$ is $\mathcal{C}^{2}$-smooth, the Taylor series conducts:
    \begin{multline*}
    \|\Delta^{(k + 1)}\| = \|\mathcal{G}(z_{t}^{(k)}) - \mathcal{G}(z_{t}^{(k - 1)})\| = \\
    \|\mathcal{G}(z_{t}^{(k - 1)}) + \frac{\partial \mathcal{G}}{\partial z}|_{z_{t}^{(k - 1)}} \cdot \Delta^{(k)} + O(\|\Delta^{(k)}\|^2) - \mathcal{G}(z_{t}^{(k - 1)})\| = \\
    \|\frac{\partial \mathcal{G}}{\partial z}|_{z_{t}^{(k - 1)}} \cdot \Delta^{(k)} + O(\|\Delta^{(k)}\|^2)\| \leq
    \ifeccv
    \else
    \\
    \fi
    \|\frac{\partial \mathcal{G}}{\partial z}|_{z_{t}^{(k - 1)}}\| \cdot \|\Delta^{(k)}\| + O(\|\Delta^{(k)}\|^2) = \\
    \frac{\psi_t}{\phi_t} \cdot \|\frac{\partial \epsilon_{\theta}}{\partial z}|_{z_{t}^{(k - 1)}}\| \cdot \|\Delta^{(k)}\| + O(\|\Delta^{(k)}\|^2)
    \end{multline*}
    Thus in a sufficiently small neighbour the convergence dynamics is defined by the scaled Jacobian norm $\frac{\psi_t}{\phi_t}~\cdot~\|\frac{\partial \epsilon_{\theta}}{\partial z}|_{z_{t}^{(k - 1)}}\|$.
\fi
\ifeccv
\else
    \paragraph{\textbf{Scaled Jacobian Norm of ReNoise Mapping}}
\fi
Figure \ref{fig:jacobian_norms} shows this scaled norm estimation for the SDXL diffusion model for various steps and ReNoise iterations indices $(k)$.
Remarkably, the ReNoise indices minimally impact the scale factor, consistently remaining below 1. This confirms in practice the convergence of the proposed algorithm.
Notably, the highest scaled norm values occur at smaller $t$ (excluding the first step) and during the initial renoising iteration. 
This validates the strategy of not applying ReNoise in early steps, where convergence tends to be slower compared to other noise levels.
Additionally, the scaled norm value for the initial $t$ approaches 0, which induces almost immediate convergence, making ReNoise an almost identical operation.

\ifeccv
Figure 7 in the main paper 
illustrates the exponential decrease in distances between consecutive elements $z_t^{(k)}$ and $z_t^{(k+1)}$, which confirms the algorithm's convergence towards the stationary point of the operator $\mathcal{G}$.
\else
\fi

\ifeccv
\else
    \input{tables/imp_details}
\fi

\paragraph{\textbf{Validation for the Averaging Strategy}}
Notably, the proposed averaging strategy is aligned with the conclusions described in the main paper and also converges to the desired stationary point. 
To verify this claim we will show that if a sequence $z_{t+1}^{(k)}$ converges to some point $z_{t+1}$, then the averages $\bar{z}_{t+1}^{(k)} = \frac{1}{k}\sum_{i=1}^{k}z_{t+1}^{(i)}$ converges to the same point.
That happens to be the stationary point of the operator $\mathcal{G}$.
We demonstrate it with the basic and standard calculus.
Assume that $z_{t+1}^{(k)} = \varepsilon_{t+1}^{(k)} + z_{t+1}$ with $\|\varepsilon_{t+1}^{(k)}\| \to 0$ as $k \to \infty$. For a fixed $\varepsilon$ we need to show that there exists $K$ so that $\|\bar{z}_{t+1}^{(k)} - z_{t+1}\| < \epsilon$ for any $k > K$.
One has
\begin{equation*}
    \bar{z}_{t+1}^{(k)} - z_{t+1} = \sum_{i=1}^{k} \frac{\varepsilon_{t+1}^{(i)}}{k}
\end{equation*}
There exists $m$ such that $\|\varepsilon_{t+1}^{(k)}\| < 0.5 \cdot \varepsilon$ once $k > m$. Then we have
\begin{multline*}
\|\sum_{i=1}^{k} \frac{\varepsilon_{t+1}^{(i)}}{k}\| \leq \frac{\|\sum_{i=1}^{m} \varepsilon_{t+1}^{(i)}\|}{k} + \frac{\|\sum_{i=m + 1}^{k} \varepsilon_{t+1}^{(i)}\|}{k} \leq \\
\frac{\|\sum_{i=1}^{m} \varepsilon_{t+1}^{(i)}\|}{k} + \frac{\varepsilon}{2} \frac{\cdot (k - m)}{k} \leq \frac{\|\sum_{i=1}^{m} \varepsilon_{t+1}^{(i)}\|}{k} + \frac{\varepsilon}{2}
\end{multline*}
given that $m$ is fixed, we can always take $k$ sufficiently large such that
\begin{equation*}
\frac{\|\sum_{i=1}^{m} \varepsilon_{t+1}^{(i)}\|}{k} < \frac{\varepsilon}{2}
\end{equation*}
this ends the proof. The very same computation conducts a similar result if the elements' weights $w_k$ are non-equal.

\ifeccv
\else
    \begin{figure}
        \centering
        \includegraphics[width=0.9\textwidth]{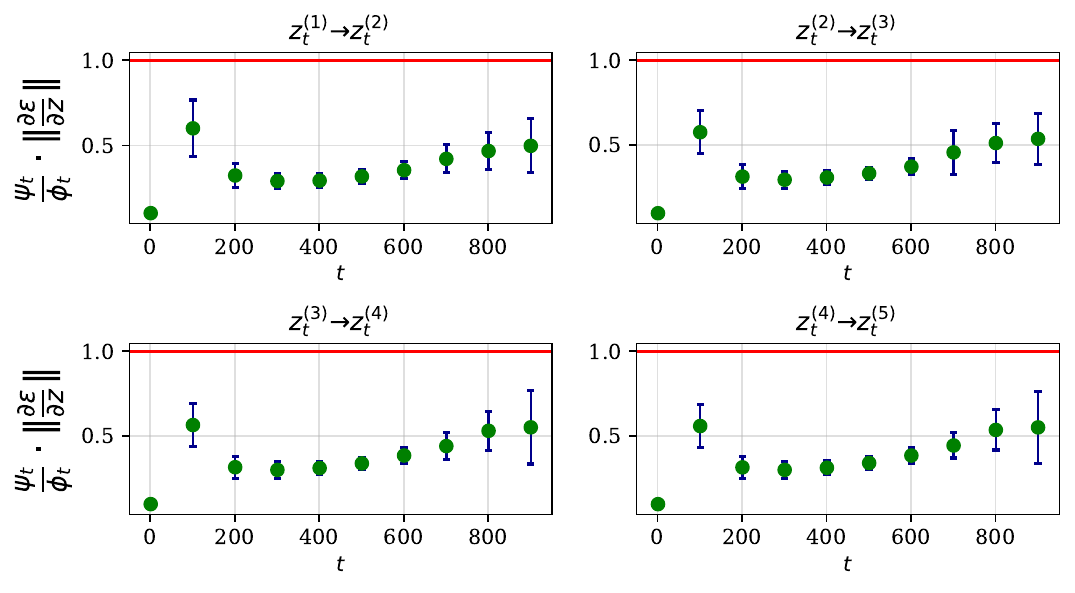}
        \caption{Scaled Jacobian norm of the mapping $z^{(k)}_{t} \to z^{(k)}_{t}$ calculated for various noise levels $t$, iterations $k$. We report the average and the standard deviation calculated over 32 images. Values below 1 indicate the exponential convergence of the ReNoise algorithm.}
        \label{fig:jacobian_norms}
    \end{figure}
\fi

%% file: tables/imp_details.tex
\begin{table*}
\ifeccv
\else
\small
\fi
\centering
\setlength{\tabcolsep}{2.5pt}
\ifeccv
\else
    \caption{
    Implementation details of ReNoise with Stable Diffusion~\cite{rombach2022highresolution}, SDXL~\cite{podell2023sdxl}, SDXL Turbo~\cite{sauer2023adversarial} and LCM LoRA~\cite{luo2023lcm}.
    }
\fi
\begin{tabular}{l | c | c |c | c |} 
    \toprule
    \multicolumn{5}{c}{\textbf{Implementation details} } \\
    \midrule
    Model Name                          & \textbf{Stable Diffusion}     & \textbf{SDXL}     & \textbf{SDXL Turbo}   & \textbf{LCM LoRA}  \\
    Noise Sampler                       & DDIM                          & DDIM              & Ancestral-Euler       & DDIM  \\
    Number of denoising steps           & 50                            & 50                & 4                     & 4  \\
    Number of renoising iterations       & 1                             & 1                 & 9                     & 7  \\
    Weights for $t<250$                & $w_1,w_2=0.5$                 & $w_1,w_2=0.5$     & $w_1,...,w_4=0.25$    & $w_1,...,w_4=0.25$  \\
    Weights for $t>250$                & $w_2=1.0$                     & $w_2=1.0$         & $w_8,...,w_{10}=0.33$   & $w_6,...,w_8=0.33$  \\
    $\lambda_\text{pair}$               & 10                            & 10                & 10                    & 20   \\
    $\lambda_\text{patch-KL}$           & 0.05                          & 0.055             & 0.055                 & 0.075   \\
    \bottomrule
\end{tabular}
\label{tb:impl_details}
\end{table*}

%% file: figures/sup_turbo_results.tex
\begin{figure} [t]
    \centering
    \setlength{\tabcolsep}{1pt}
    \addtolength{\belowcaptionskip}{-10pt}
    {\scriptsize
    \centering

    \begin{tabular}{c ccc}
        {\begin{tabular}{c} Original \\ Image  \end{tabular}} & 
        \multicolumn{3}{c}{$\longleftarrow$ Editing Results $\longrightarrow$} \\
        
        \includegraphics[width=0.205\columnwidth]{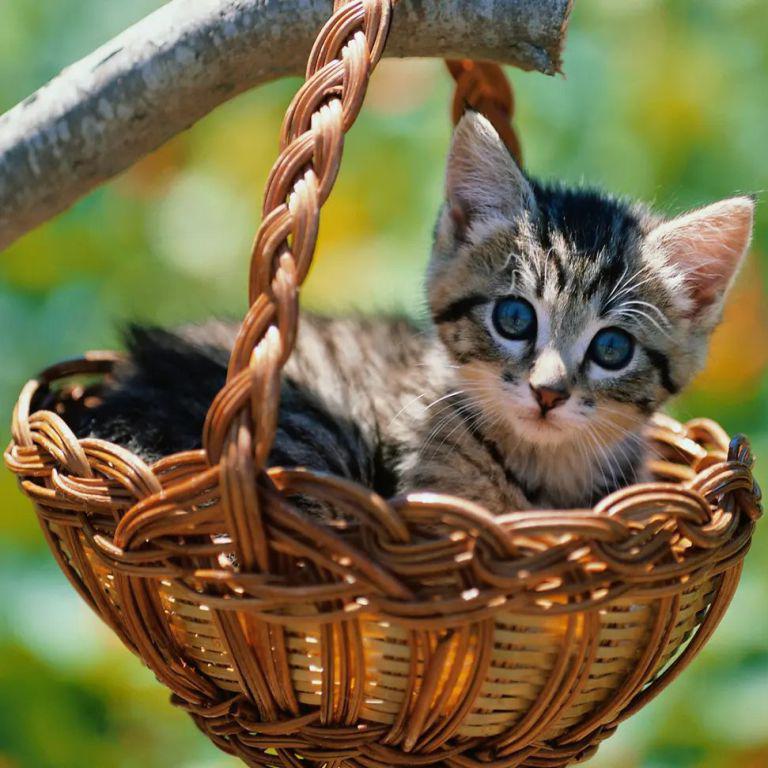} &
        \includegraphics[width=0.205\columnwidth]{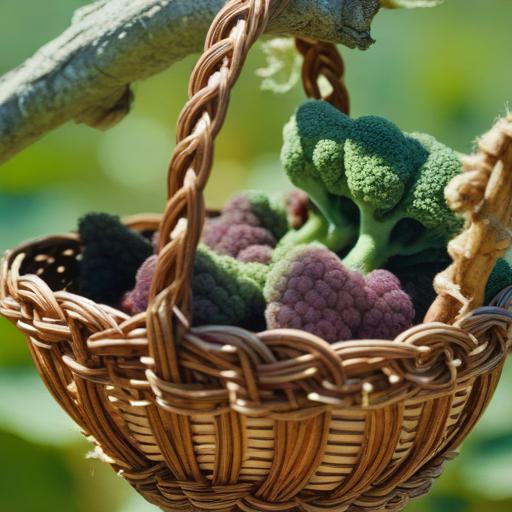} &
        \includegraphics[width=0.205\columnwidth]{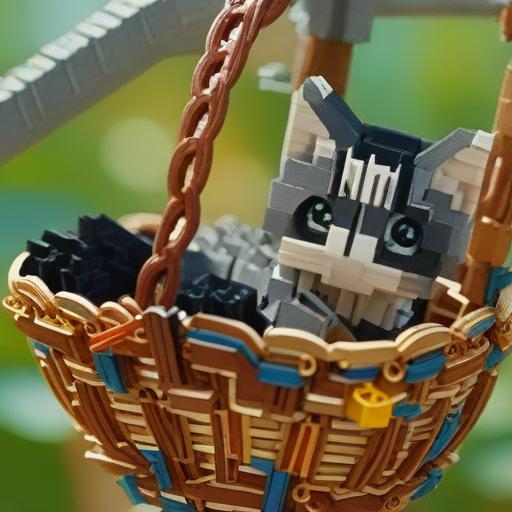} &
        \includegraphics[width=0.205\columnwidth]{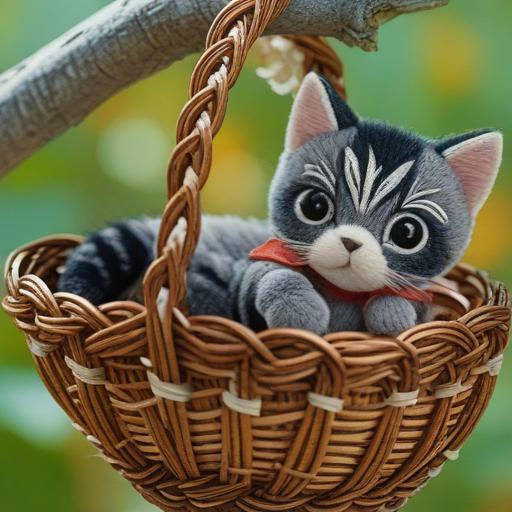} \\
        
        {``kitten''} &
        {``broccoli''} &
        {``made of lego''} &
        {``plush toy''} \\
        
        \includegraphics[width=0.205\columnwidth]{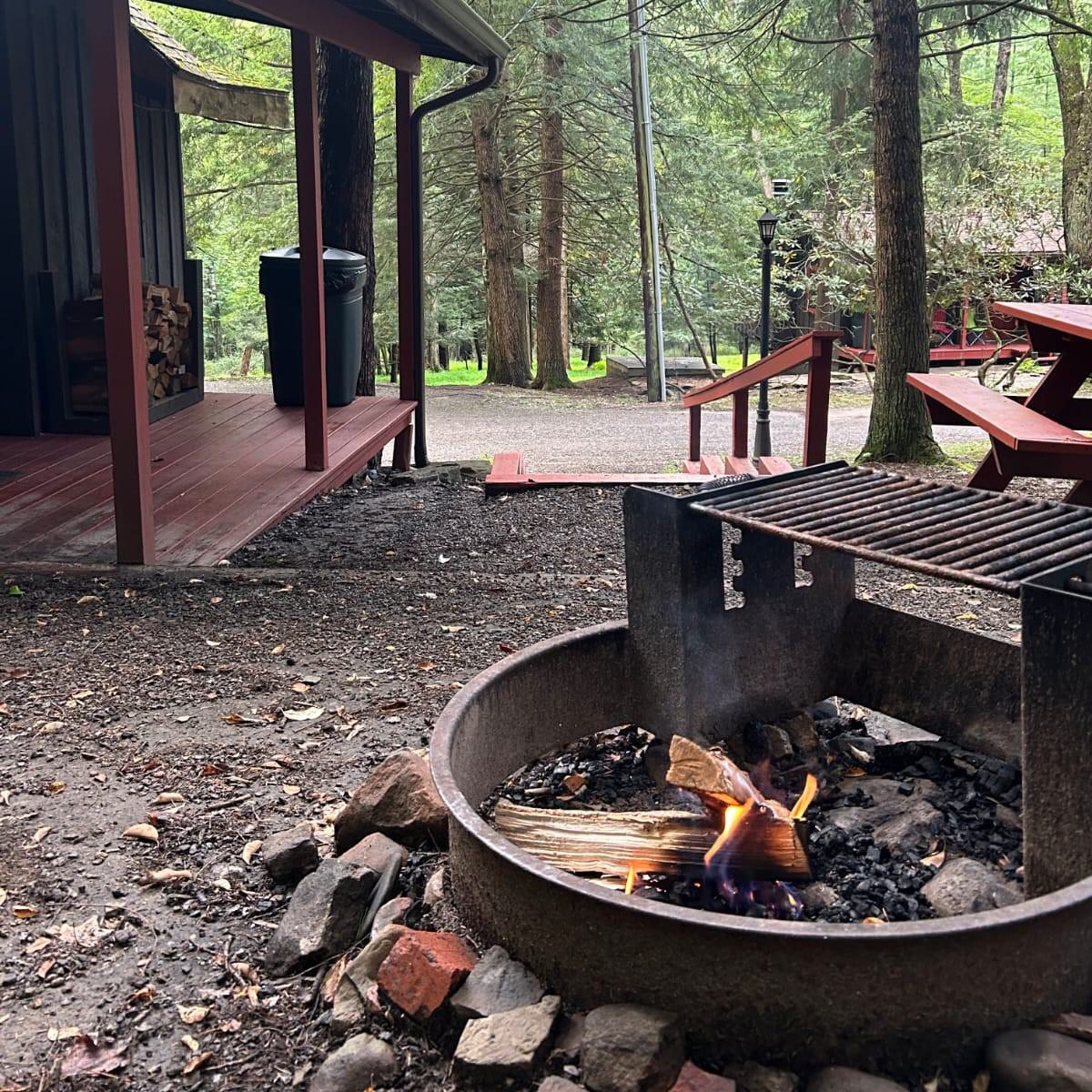} &
        \includegraphics[width=0.205\columnwidth]{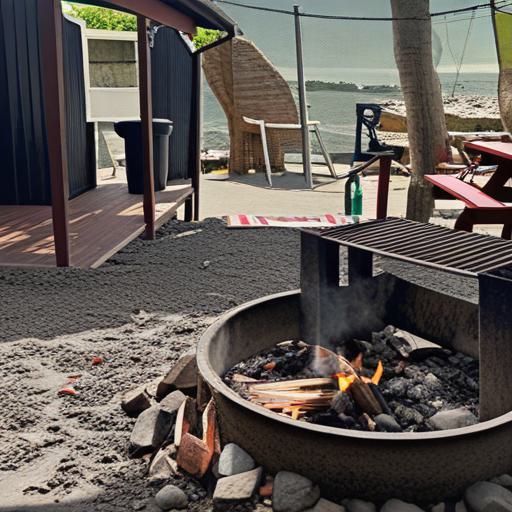} &
        \includegraphics[width=0.205\columnwidth]{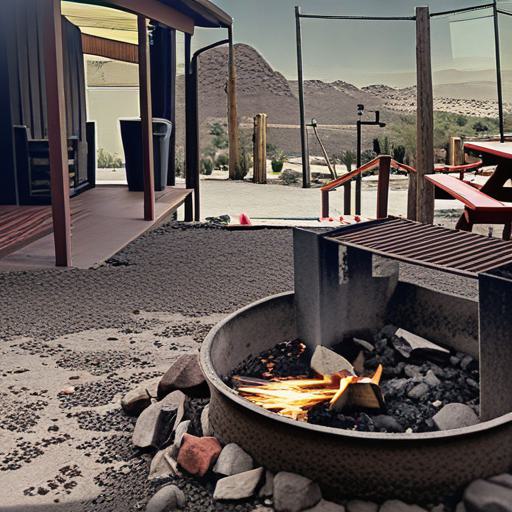} &
        \includegraphics[width=0.205\columnwidth]{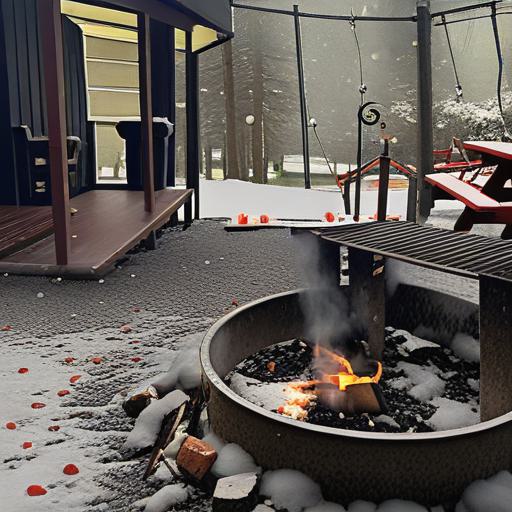} \\

        {``...in the woods''} &
        {``...in the beach''} &
        {``...in the desert''} &
        {``...in the snow''} \\
        
        \includegraphics[width=0.205\columnwidth]{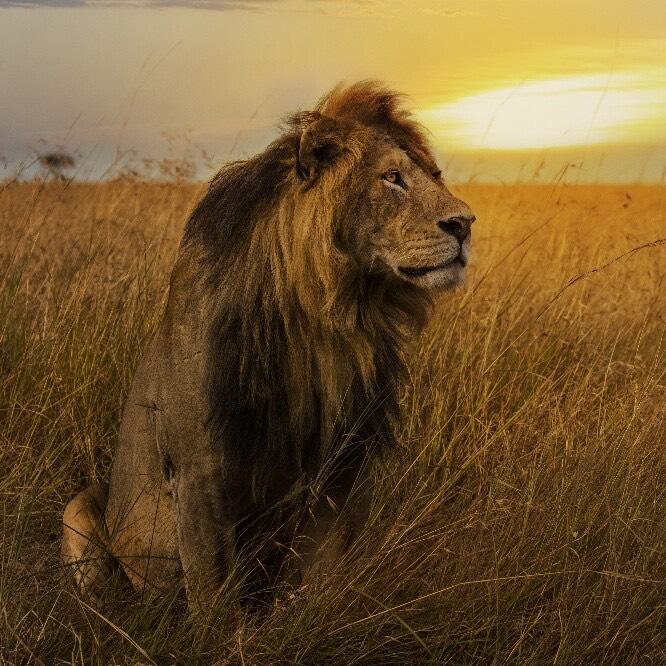} &
        \includegraphics[width=0.205\columnwidth]{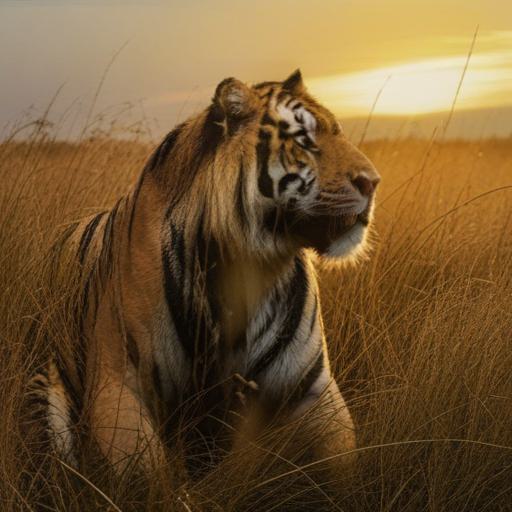} &
        \includegraphics[width=0.205\columnwidth]{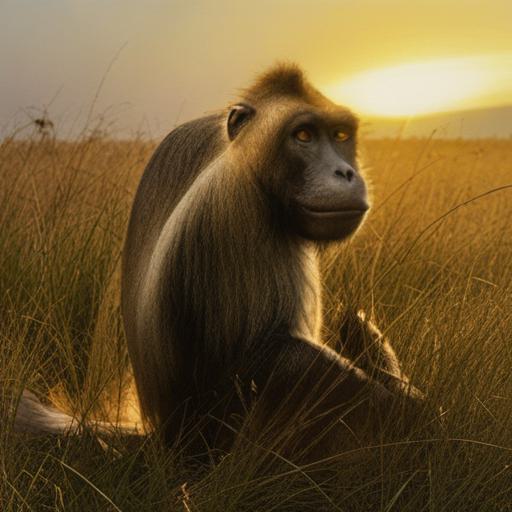} &
        \includegraphics[width=0.205\columnwidth]{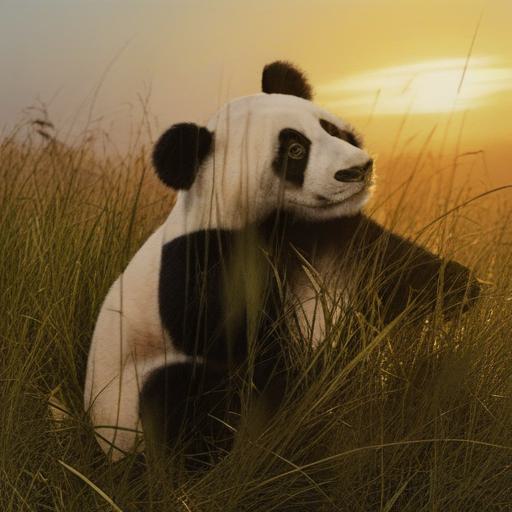} \\

        {``lion''} &
        {``tiger''} &
        {``monkey''} &
        {``panda''}  \\
        
        \includegraphics[width=0.205\columnwidth]{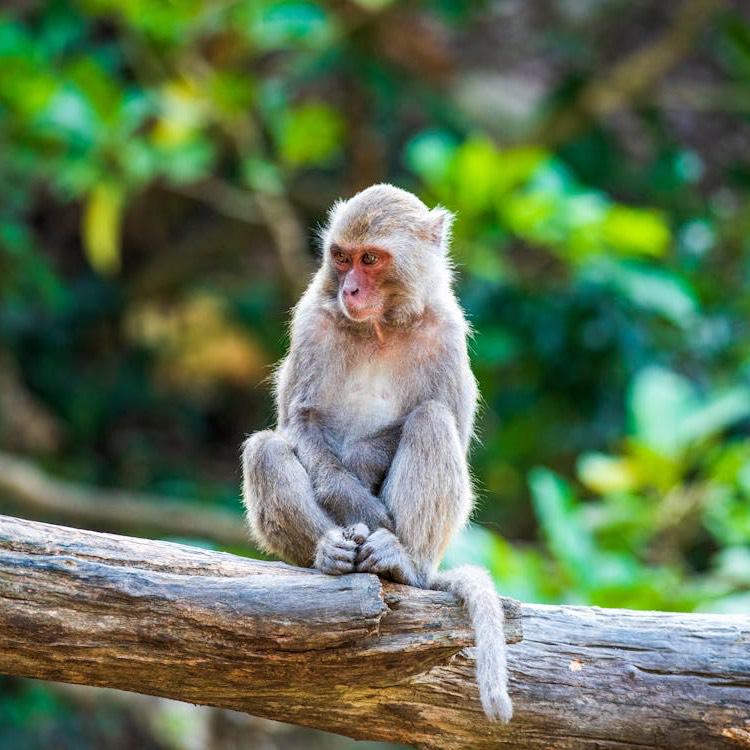} &
        \includegraphics[width=0.205\columnwidth]{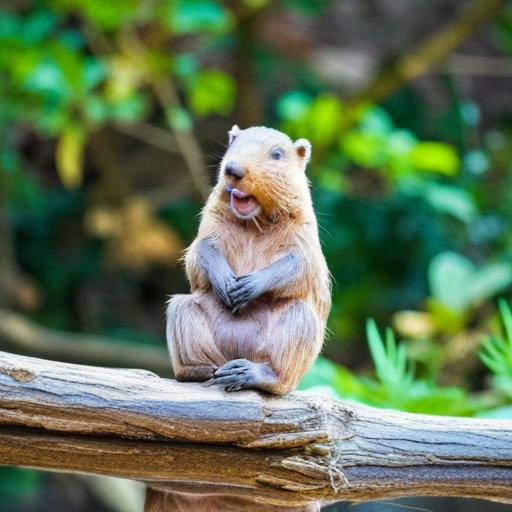} &
        \includegraphics[width=0.205\columnwidth]{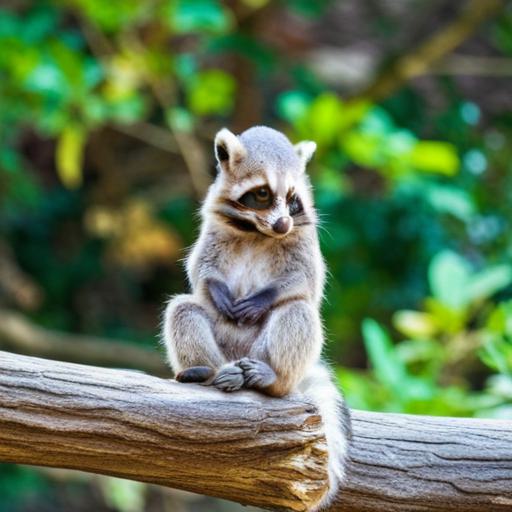} &
        \includegraphics[width=0.205\columnwidth]{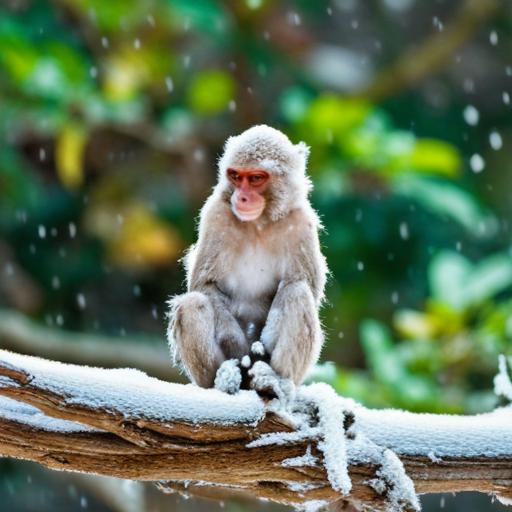} \\

        {``monkey''} &
        {``beaver''} &
        {``raccoon''} &
        {``...in the snow''}  \\
    \end{tabular}

    }
    \vspace{-0.245cm}    
    \caption{  
    SDXL Turbo editing results.
    Each row showcases one image. The leftmost image is the original, followed by three edited versions. 
    The text below each edited image indicates the specific word or phrase replaced or added to the original prompt for that specific edit.
    \ifeccv
    \else
        \vspace{-15pt}
    \fi
    }
    \label{fig:turbo_results}
\end{figure}

%% file: figures/lpips_unet_op_figure.tex
\begin{figure*}
    \centering
    \setlength{\tabcolsep}{1pt}
    \addtolength{\belowcaptionskip}{-10pt}
    {\small
    \centering

    \begin{tabular}{c c c}
        { SDXL } &
        { SDXL Turbo } &
        { LCM } \\
        
        \includegraphics[width=0.33\textwidth]{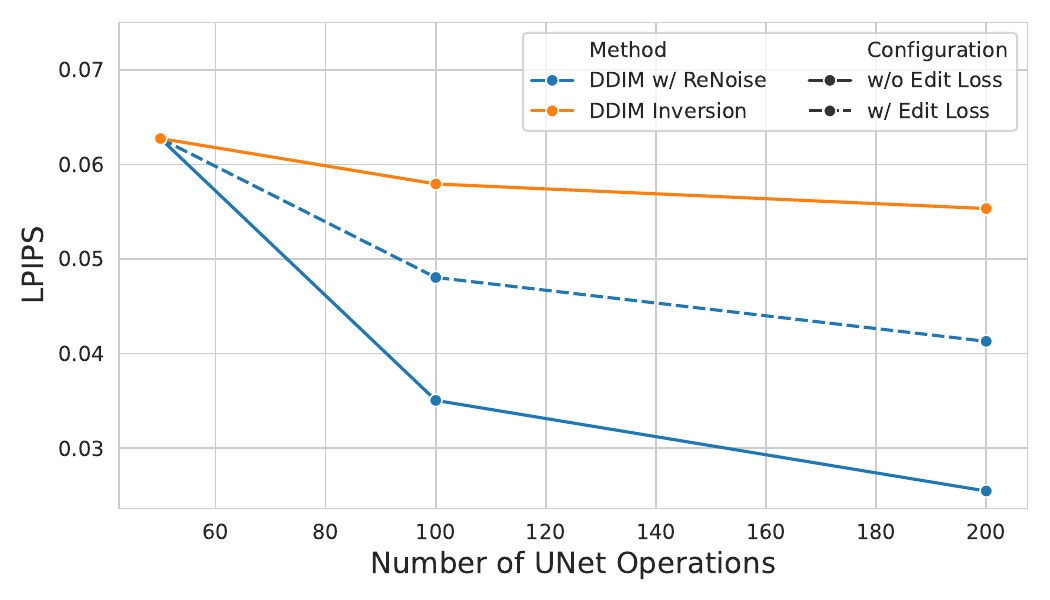} &
        \includegraphics[width=0.33\textwidth]{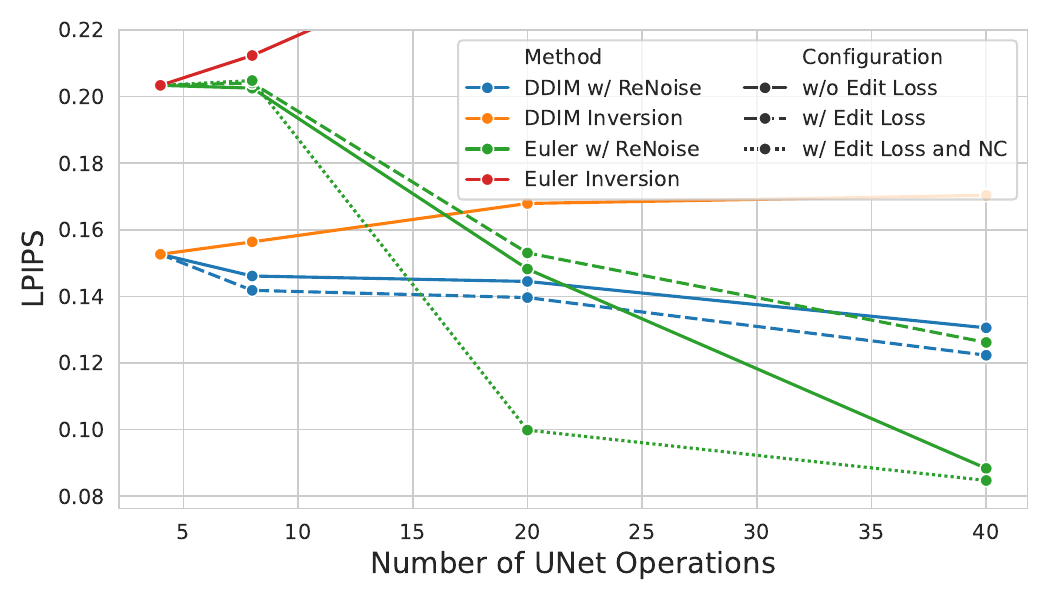} &
        \includegraphics[width=0.33\textwidth]{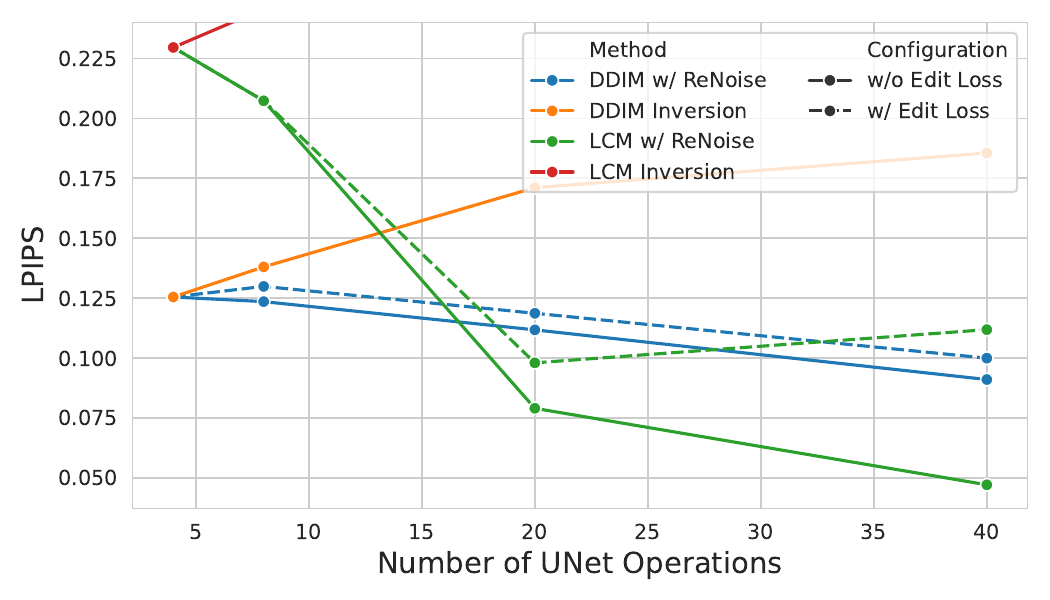} \\      
    \end{tabular}
    
    }
    \vspace{-0.225cm}    
    \caption{
    Image reconstruction results comparing sampler reversing inversion techniques across different samplers (e.g., vanilla DDIM inversion) with our ReNoise method using the same sampler. The number of denoising steps remains fixed. However, the number of UNet passes varies, 
    with the number of inversion steps increasing in the sampler reversing approach, and the number of renoising iterations increasing in our method.
    We present various configuration options for our method, including options with or without edit enhancement loss and Noise Correction (NC).
    \ifeccv
    \else
        \vspace{-14pt}
    \fi
    }
    \label{fig:lpips_unet_op_figure}
\end{figure*}

%% file: 8.2-Sup-experiments.tex
\ifeccv

\input{figures/sup_turbo_results}\input{figures/lpips_unet_op_figure}
\fi

\section{Additional Experiments}

\paragraph{\textbf{Editing Results With SDXL Turbo}}
Figure~\ref{fig:turbo_results} showcases additional image editing examples achieved using our ReNoise inversion method. These edits are accomplished by inverting the image with a source prompt, and then incorporating a target prompt that differs by only a few words during the denoising process.

\input{figures/pix2pix_figure}

\paragraph{\textbf{Reconstruction and Speed}}
\label{sec:sup_reconstruction_and_speed}

We continue our evaluation of the reconstruction-speed tradeoff from Section 5.1 in the main paper.
Figure~\ref{fig:lpips_unet_op_figure} presents quantitative LPIPS results for the same configuration described in the main paper. As expected, LPIPS scores exhibit similar trends to the PSNR metric shown previously.

    \input{figures/edit_ablation}
    \input{figures/sup_turbo_compare_ddpm}

\paragraph{\textbf{Image Editing Ablation}}
\label{sec:sup_img_edit_ablation}
Figure~\ref{fig:edit_ablation} visually illustrates the impact of edit enhancement losses and noise correction when editing inverted images using SDXL Turbo~\cite{sauer2023adversarial}.
While achieving good reconstructions without the $\mathcal{L}_{\text{edit}}$ regularization, the method struggles with editing capabilities (second column).
Although the $\mathcal{L}_{\text{edit}}$ regularization enhances editing capabilities, it comes at the cost of reduced reconstruction accuracy of the original image, as evident in the two middle columns.
In the third column, we use $\mathcal{L}_{\text{KL}}$ as defined in pix2pix-zero~\cite{parmar2023zeroshot}.
While $\mathcal{L}_{\text{patch-KL}}$ surpasses $\mathcal{L}_{\text{KL}}$ in original image preservation, further improvements are necessary.
These improvements are achieved by using the noise correction technique. 
To correct the noise, we can either override the noise $\epsilon_t$ in Equation 1 in the main paper (fifth column), or optimize it (sixth column). As observed, overriding the noise $\epsilon_t$ affects editability, while optimizing it achieves good results in terms of both reconstruction and editability. Therefore, in our full method, we use $\mathcal{L}_{\text{patch-KL}}$, $\mathcal{L}_{\text{pair}}$, and optimization-based noise correction.

\ifeccv
\else
    \vspace{-6pt}
\fi

\paragraph{\textbf{Editing With ReNoise}}
The ReNoise technique provides a drop-in improvement for methods (e.g., editing methods) that rely on inversion methods like DDIM~\cite{dhariwal2021diffusion}, negative prompt inversion~\cite{miyake2023negativeprompt} and more. It seamlessly integrates with these existing approaches, boosting their performance without requiring extensive modifications.
Figure~\ref{fig:pix2pix_figure} showcases image editing examples using Zero-Shot Image-to-Image Translation (pix2pix-zero)~\cite{parmar2023zeroshot}. We compared inversions with both the pix2pix-zero inversion method and our ReNoise method. Our method demonstrably preserves finer details from the original image while improving editability, as exemplified by the dog-to-cat translation.

\ifeccv
\else
    \vspace{-12pt}
\fi

\paragraph{\textbf{Inversion for Non-deterministic Samplers}}
In Figure~\ref{fig:sup_turbo_compare_ddpm} we show more qualitative comparisons with ``an edit-friendly DDPM''~\cite{hubermanspiegelglas2023edit} where we utilize SDXL Turbo~\cite{sauer2023adversarial}.
As can be seen, encoding a significant amount of information within only a few external noise vectors, $\epsilon_t$, limits editability in certain scenarios, such as the ginger cat example. 
It is evident that the edit-friendly DDPM method struggles to deviate significantly from the original image in certain aspects while also failing to faithfully preserve it in others. 
For instance, it encounters difficulty in transforming the cat into a ginger cat while omitting the preservation of the decoration in the top left corner.
In addition, the image quality of edits produced by edit-friendly DDPM is lower, as demonstrated in the dog example.

\ifeccv
\else
    \vspace{-12pt}
\fi

\paragraph{\textbf{Improving DDIM Instabilities}}
As mentioned in Section 3.1 of the main paper, DDIM inversion~\cite{dhariwal2021diffusion} can exhibit instabilities depending on the prompt. 
Figure~\ref{fig:ddim_with_renoise_stability} demonstrates this with image reconstruction on SDXL~\cite{podell2023sdxl} using an empty prompt. Notably, incorporating even a single renoising iteration significantly improves inversion stability.

\input{figures/ddim_with_renoise_stability}

%% file: figures/pix2pix_figure.tex
\begin{figure} [t]
    \centering
    \setlength{\tabcolsep}{1pt}
    \addtolength{\belowcaptionskip}{-10pt}
    {\scriptsize
    \centering

    \ifeccv
    \begin{tabular}{c cc c cc}
        {\begin{tabular}{c} Original \\ Image  \end{tabular}} & 
        {\begin{tabular}{c} Pix2Pix zero \\   \end{tabular}} & 
        {\begin{tabular}{c} Pix2Pix zero \\ w/ ReNoise  \end{tabular}} & 
        {\begin{tabular}{c} Original \\ Image  \end{tabular}} & 
        {\begin{tabular}{c} Pix2Pix zero \\   \end{tabular}} & 
        {\begin{tabular}{c} Pix2Pix zero \\ w/ ReNoise  \end{tabular}} \\
        
        \includegraphics[width=0.162\columnwidth]{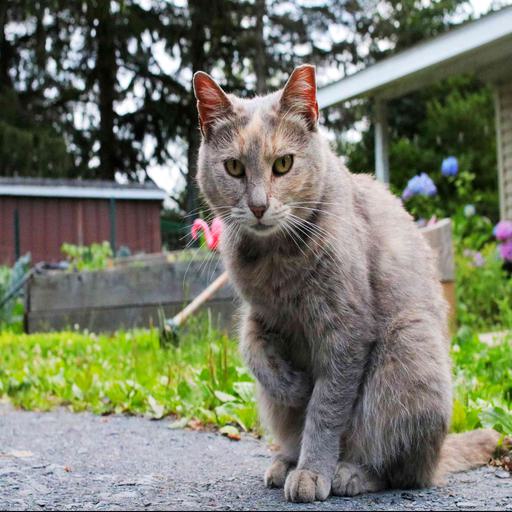} &
        \includegraphics[width=0.162\columnwidth]{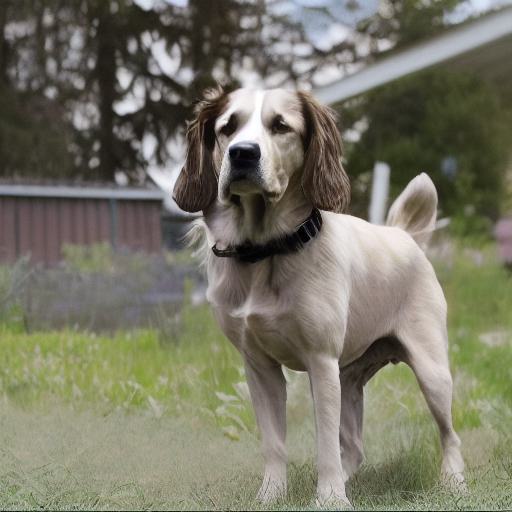} &
        \includegraphics[width=0.162\columnwidth]{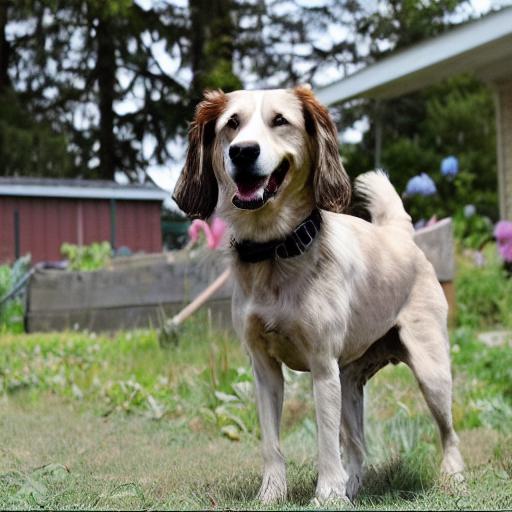} &
        
        \includegraphics[width=0.162\columnwidth]{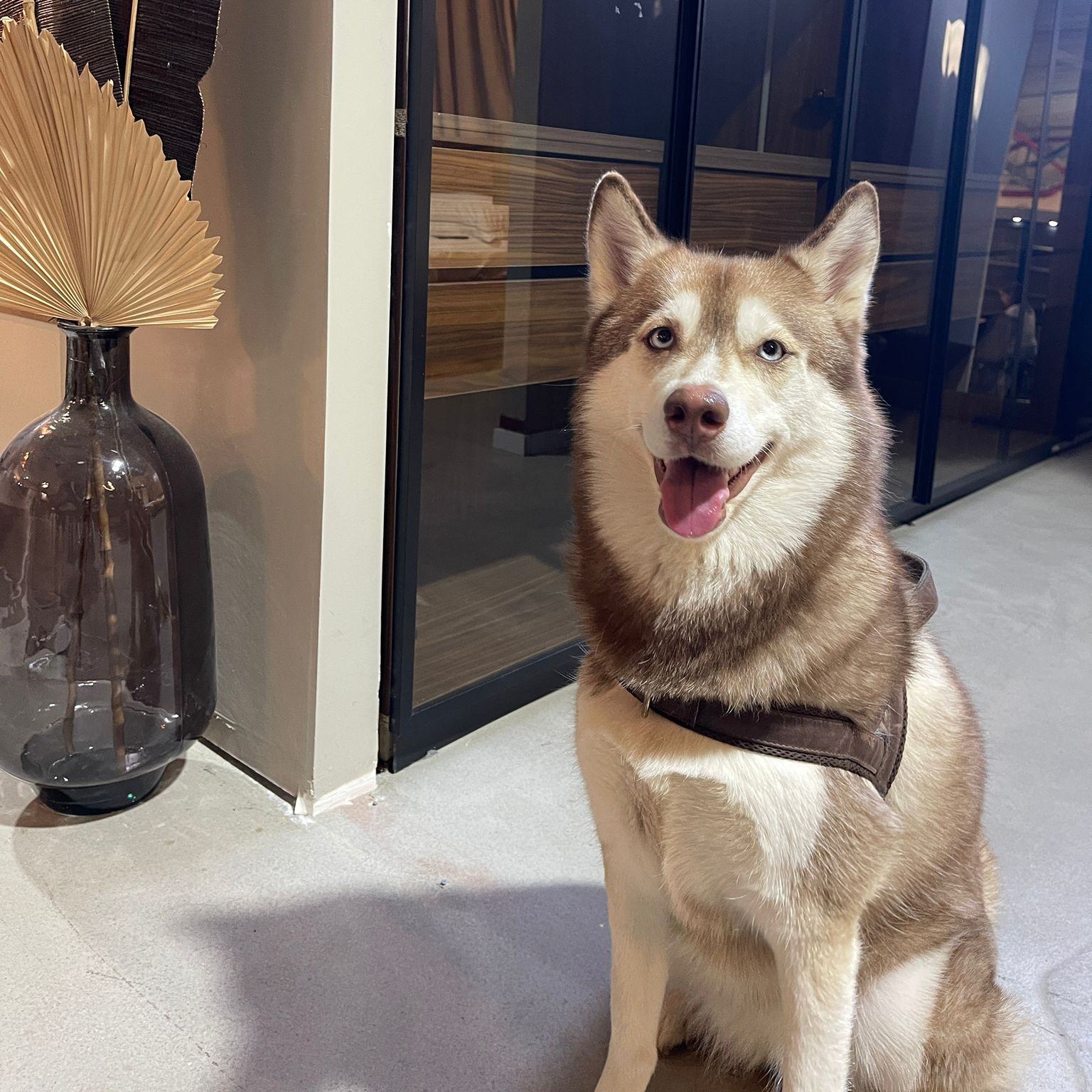} &
        \includegraphics[width=0.162\columnwidth]{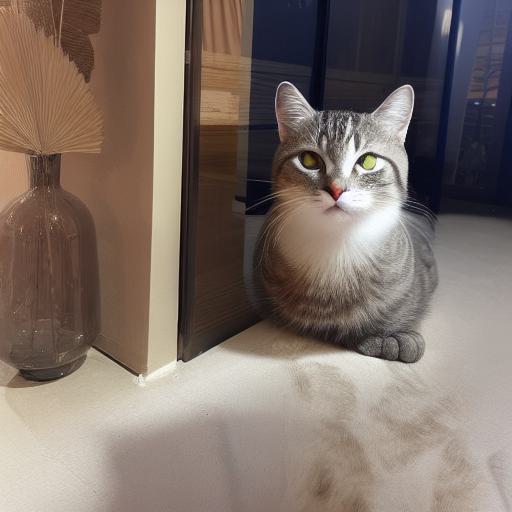} &
        \includegraphics[width=0.162\columnwidth]{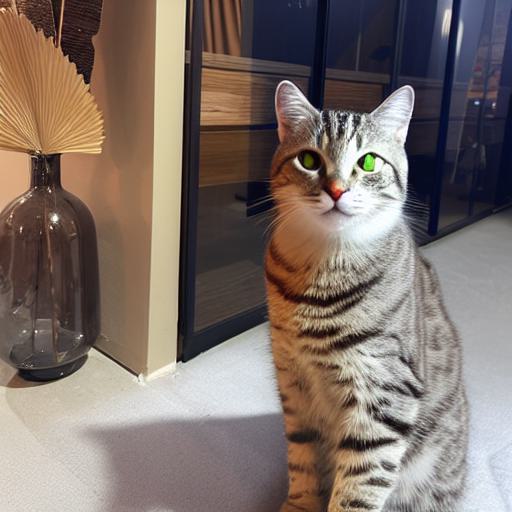} \\

        {} &
        \multicolumn{2}{c}{``Cat''  $\longrightarrow$ ``Dog''} &
        
        {} &
        \multicolumn{2}{c}{``Dog''  $\longrightarrow$ ``Cat''} \\
    \end{tabular}
    \else
    \begin{tabular}{c cc}
        {\begin{tabular}{c} Original \\ Image  \end{tabular}} & 
        {\begin{tabular}{c} Pix2Pix zero \\   \end{tabular}} & 
        {\begin{tabular}{c} Pix2Pix zero \\ w/ ReNoise  \end{tabular}} \\
        
        \includegraphics[width=0.33\columnwidth]{images/pix2pix/origin_1.jpg} &
        \includegraphics[width=0.33\columnwidth]{images/pix2pix/pix2pix_1.jpg} &
        \includegraphics[width=0.33\columnwidth]{images/pix2pix/our_1.jpg} \\
        
        {} &
        \multicolumn{2}{c}{``Cat''  $\longrightarrow$ ``Dog''} \\
        
        \includegraphics[width=0.33\columnwidth]{images/pix2pix/origin_2.jpg} &
        \includegraphics[width=0.33\columnwidth]{images/pix2pix/pix2pix_2.jpg} &
        \includegraphics[width=0.33\columnwidth]{images/pix2pix/our_2.jpg} \\
        
        {} &
        \multicolumn{2}{c}{``Dog''  $\longrightarrow$ ``Cat''} \\
    \end{tabular}
    
    \fi

    }
    \vspace{-0.245cm}    
    \caption{  
    Zero-Shot Image-to-Image Translation editing results.
    This figure compares editing results with Stable Diffusion~\cite{rombach2022highresolution} achieved using two inversion methods: pix2pix-zero~\cite{parmar2023zeroshot} and our proposed ReNoise inversion. As observed, ReNoise inversion preserves image details while effectively incorporating the desired edits.
    \ifeccv
    \else
        \vspace{-15pt}
    \fi
    }
    \label{fig:pix2pix_figure}
    \ifeccv
        \vspace{-10pt}
    \fi
\end{figure}

%% file: figures/edit_ablation.tex
\begin{figure*}
    \centering
    \setlength{\tabcolsep}{1pt}
    \addtolength{\belowcaptionskip}{-10pt}
    {\small
    \centering

    \ifeccv
        \hspace{-15pt}
    \fi
    \begin{tabular}{c cccccc}
            &
        \ifeccv
            {\begin{tabular}{c} Original \\ Image  \end{tabular}} &
            {\begin{tabular}{c} Our w/o  \\ Edit Losses  \end{tabular}} &
            {\begin{tabular}{c} w/ $\mathcal{L}_{\text{KL}}$ \\  and $\mathcal{L}_{\text{pair}}$ \end{tabular}} &
            {\begin{tabular}{c} w/ $\mathcal{L}_{\text{patch-KL}}$ \\  and $\mathcal{L}_{\text{pair}}$ \end{tabular}} &
            {\begin{tabular}{c} w/ NC  \\ Override \end{tabular}} &
            {\begin{tabular}{c} w/ NC \\ Optimize \end{tabular}} \\
        \else
            {\begin{tabular}{c} Original \\ Image  \end{tabular}} &
            {\begin{tabular}{c} Our w/o  \\ Edit Losses  \end{tabular}} &
            {\begin{tabular}{c} Our w/  \\ $\mathcal{L}_{\text{KL}}$ and $\mathcal{L}_{\text{pair}}$ \end{tabular}} &
            {\begin{tabular}{c} Our w/  \\ $\mathcal{L}_{\text{patch-KL}}$ and $\mathcal{L}_{\text{pair}}$ \end{tabular}} &
            {\begin{tabular}{c} Our w/  \\ Noise Correction \\ Override \end{tabular}} &
            {\begin{tabular}{c} Our w/  \\ Noise Correction \\ Optimize \end{tabular}} \\
        \fi
        
        \ifeccv
            \raisebox{20pt}{\rotatebox{90}{Rec.}} &
        \else
            \raisebox{15pt}{\rotatebox{90}{Rec.}} &
        \fi
        \includegraphics[width=0.16\linewidth]{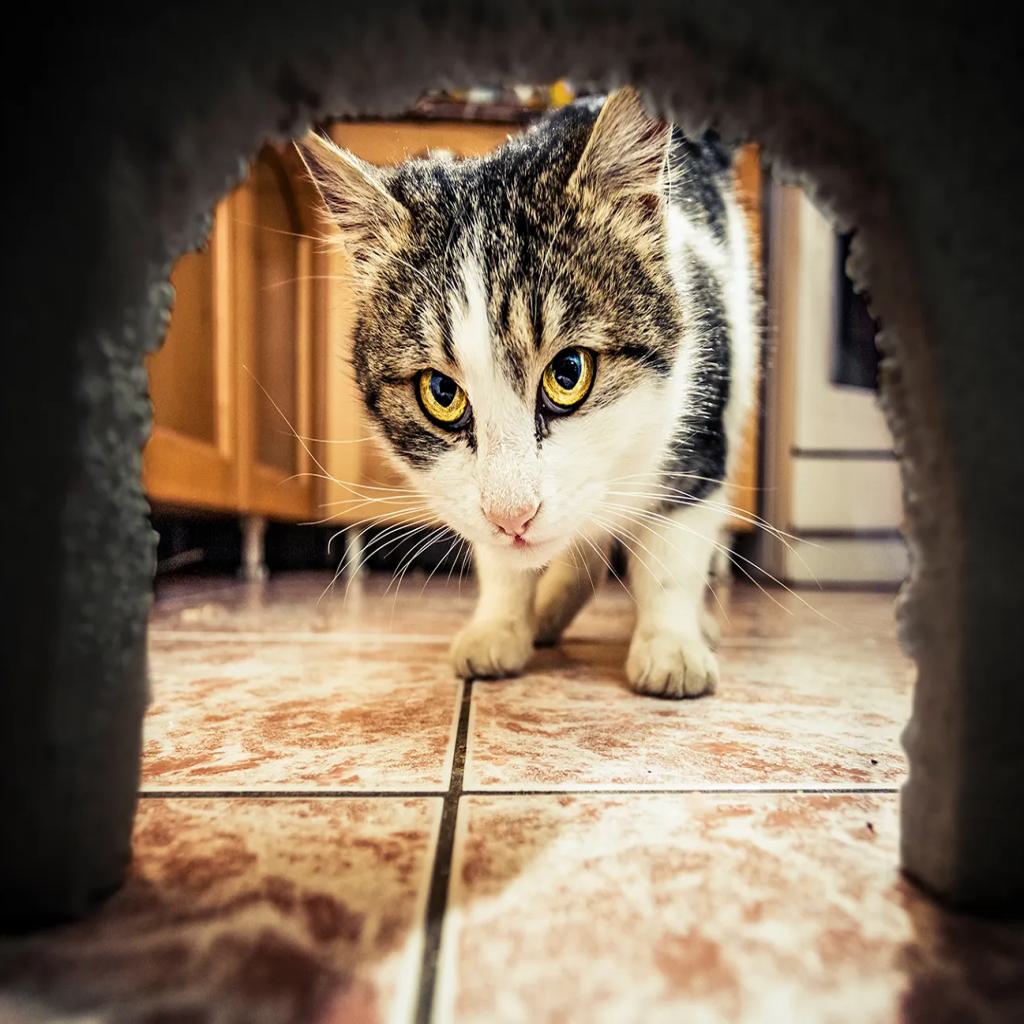} &
        \includegraphics[width=0.16\linewidth]{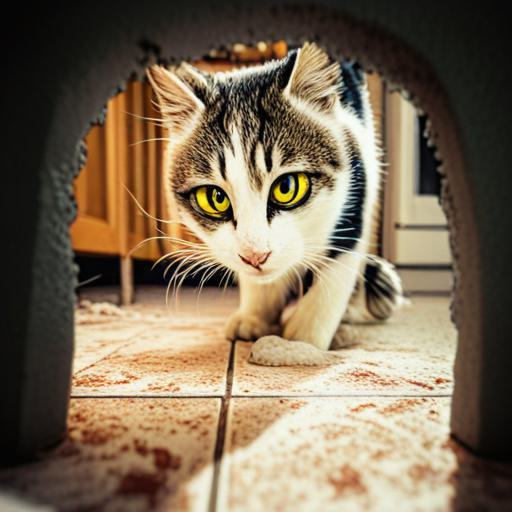} &
        \includegraphics[width=0.16\linewidth]{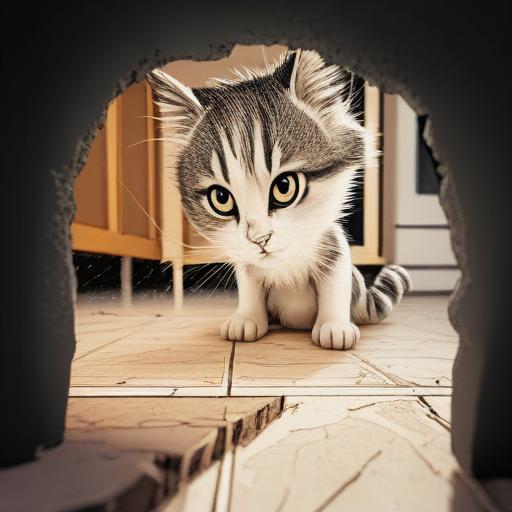} &
        \includegraphics[width=0.16\linewidth]{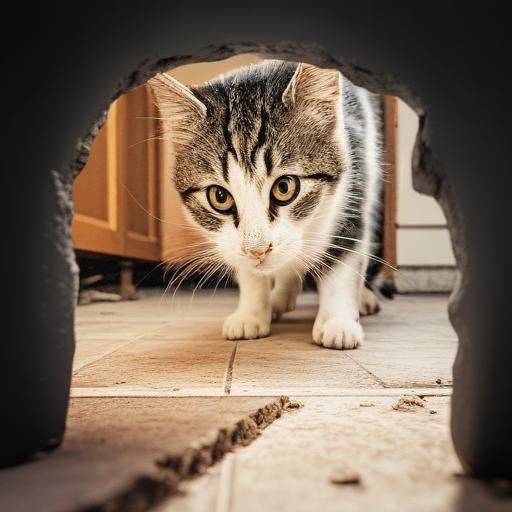} &
        \includegraphics[width=0.16\linewidth]{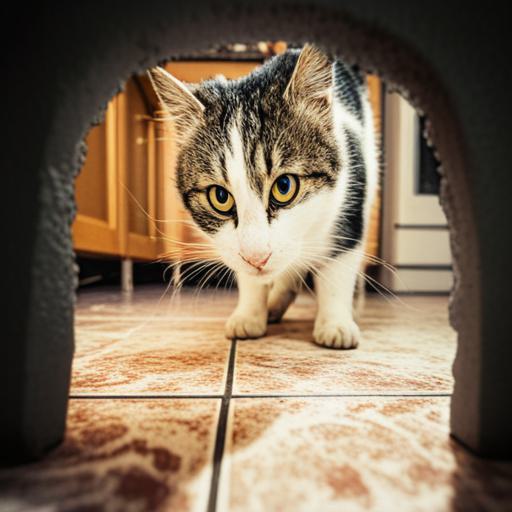} &
        \includegraphics[width=0.16\linewidth]{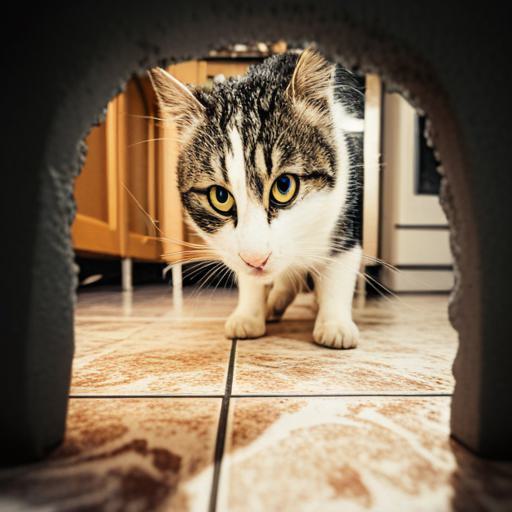} \\

        \ifeccv
            \raisebox{20pt}{\rotatebox{90}{Edit}} &
        \else
            \raisebox{15pt}{\rotatebox{90}{Edit}} &
        \fi
        \includegraphics[width=0.16\linewidth]{images/dataset-2/20-cat.jpg} &
        \includegraphics[width=0.16\linewidth]{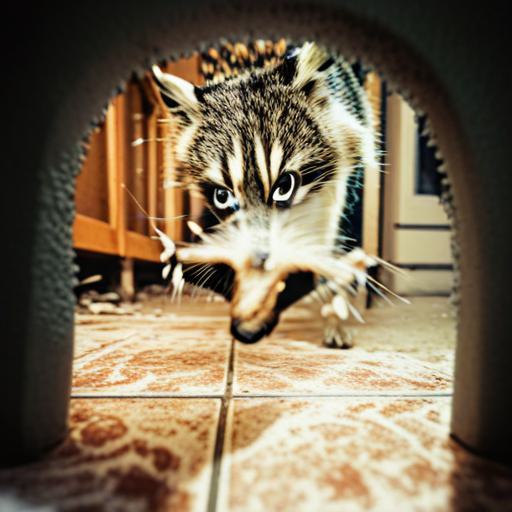} &
        \includegraphics[width=0.16\linewidth]{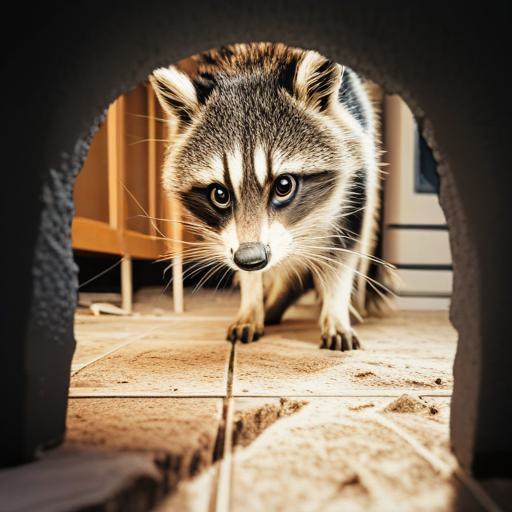} &
        \includegraphics[width=0.16\linewidth]{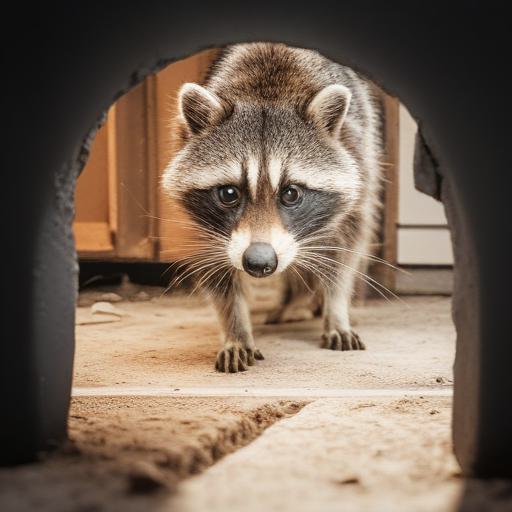} &
        \includegraphics[width=0.16\linewidth]{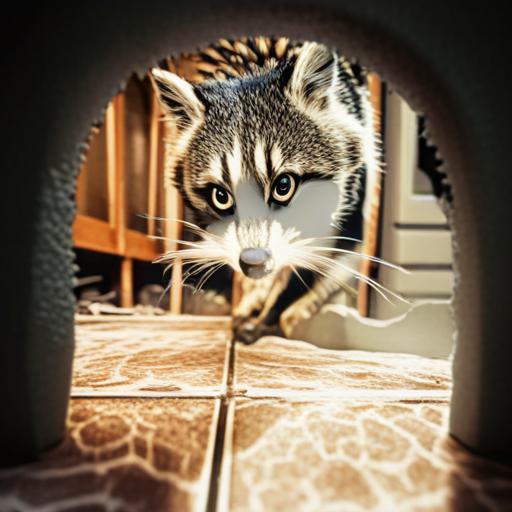} &
        \includegraphics[width=0.16\linewidth]{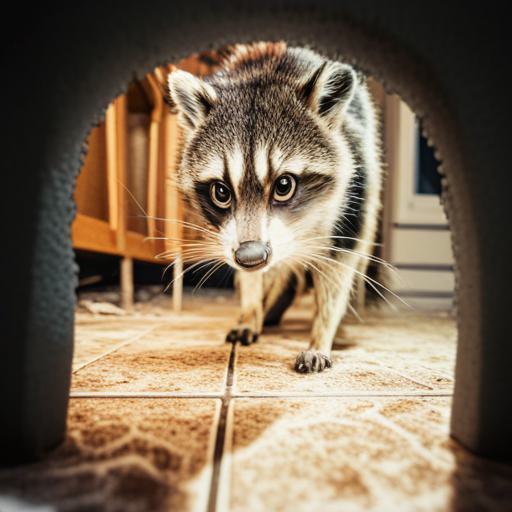} \\
    \end{tabular}
    
    }
    \vspace{-0.165cm}    
    \caption{
    Ablation study on SDXL Turbo for image editing. The first row displays reconstructed images, while the second row showcases edits replacing ``cat'' with ``raccoon''. Each column represents a different inversion configuration. From left to right, the second column demonstrates results with renoising iterations and estimations averaging. 
    In the following two columns we employ different edit enhancement losses. Finally, the last two columns present results using both noise correction (NC) and $\mathcal{L}_{\text{edit}}$. In the fifth column, NC overrides the noise $\epsilon_t$, while in the sixth column, we present the results of our full method, where NC optimizes $\epsilon_t$.
    \ifeccv
    \else
        \vspace{-14pt}
    \fi
    }
    \label{fig:edit_ablation}
\end{figure*}

%% file: figures/sup_turbo_compare_ddpm.tex
\begin{figure} [t]
    \centering
    \setlength{\tabcolsep}{1pt}
    \addtolength{\belowcaptionskip}{-10pt}
    {\scriptsize
    \begin{tabular}{c c c c c}
        \raisebox{23pt}{\rotatebox{90}{Ours}} &
        \includegraphics[width=0.23\columnwidth]{images/dataset-2/21-cat-in-bed.jpg} &
        \includegraphics[width=0.23\columnwidth]{images/turbo_compare_ddpm/our/recon_cat_bed.jpg} &
        \includegraphics[width=0.23\columnwidth]{images/turbo_compare_ddpm/our/cat_gingercat.jpg} &
        \includegraphics[width=0.23\columnwidth]{images/turbo_compare_ddpm/our/cat_wood2metal.jpg} \\
        \raisebox{11pt}{\rotatebox{90}{Edit Friendly}} &
        \includegraphics[width=0.23\columnwidth]{images/dataset-2/21-cat-in-bed.jpg} &
        \includegraphics[width=0.23\columnwidth]{images/turbo_compare_ddpm/ddpm/friendly_recon_cat_bed.jpg} &
        \includegraphics[width=0.23\columnwidth]{images/turbo_compare_ddpm/ddpm/cat_gingercat-0.75.jpg} &
        \includegraphics[width=0.23\columnwidth]{images/turbo_compare_ddpm/ddpm/cat_wood2metal-0.75.jpg} \\
        & Original & Reconstruction & ``ginger cat'' & ``wood'' $\rightarrow$ ``metal'' \\
        \raisebox{23pt}{\rotatebox{90}{Ours}} &
        \includegraphics[width=0.23\columnwidth]{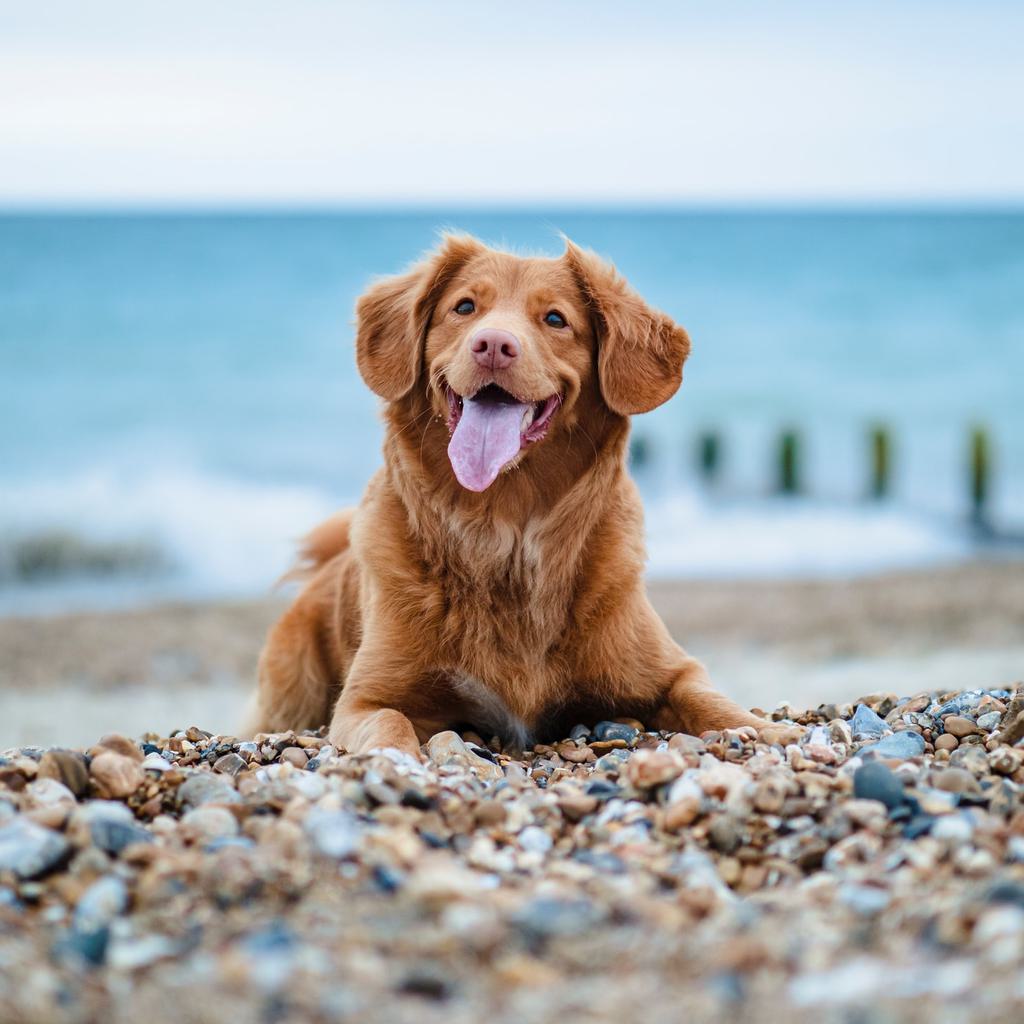} &
        \includegraphics[width=0.23\columnwidth]{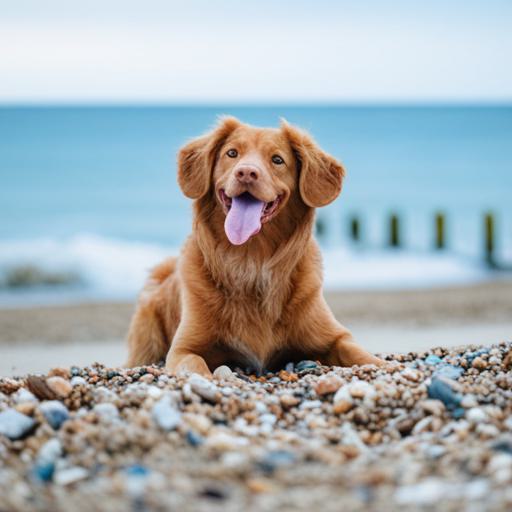} &
        \includegraphics[width=0.23\columnwidth]{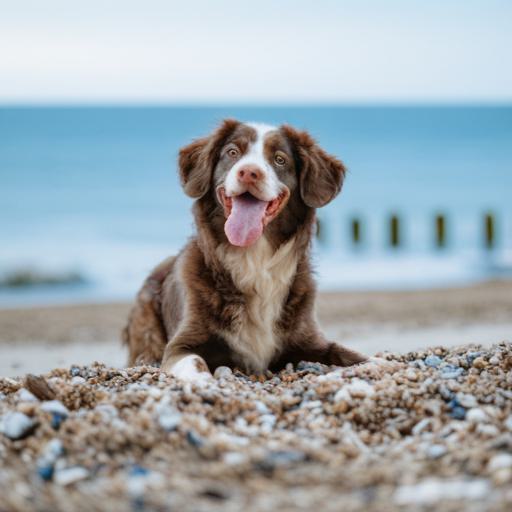} &
        \includegraphics[width=0.23\columnwidth]{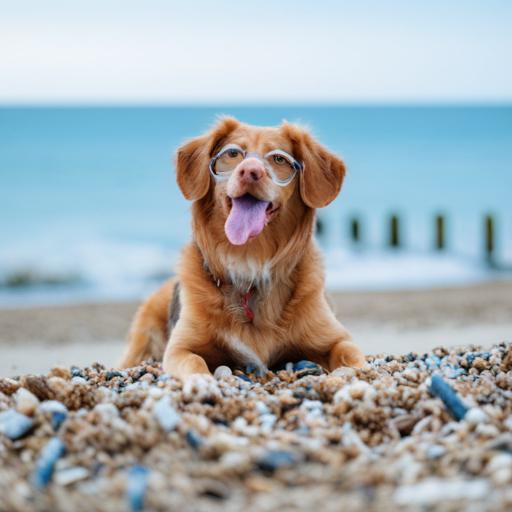} \\
        \raisebox{11pt}{\rotatebox{90}{Edit Friendly}} &
        \includegraphics[width=0.23\columnwidth]{images/dataset-1024/02-dog.jpg} &
        \includegraphics[width=0.23\columnwidth]{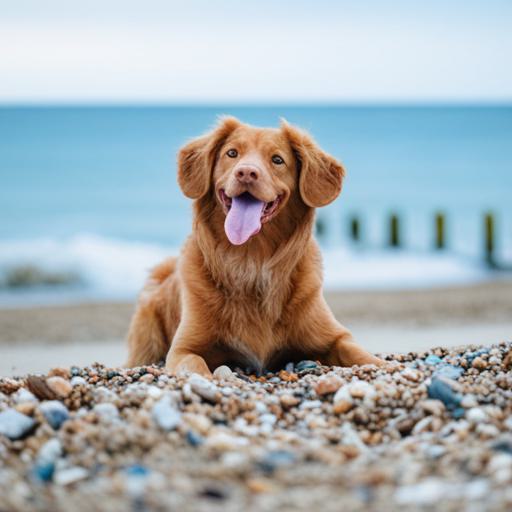} &
        \includegraphics[width=0.23\columnwidth]{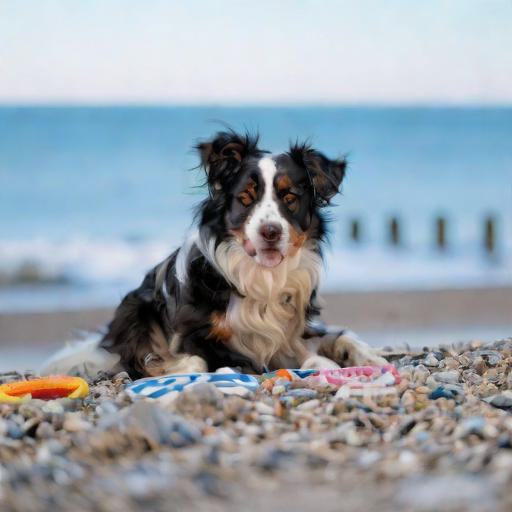} &
        \includegraphics[width=0.23\columnwidth]{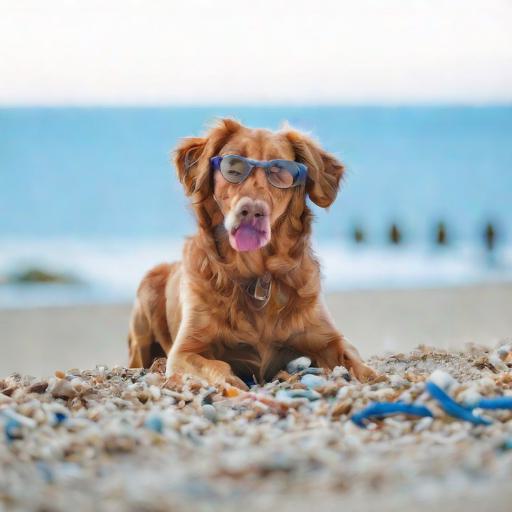} \\
        & Original & Reconstruction & ``black\&white dog'' & ``...with glasses''
    \end{tabular}
  
    }
    \ifeccv
        \vspace{-0.325cm}
    \else
        \vspace{-0.15cm}
    \fi
    \caption{
    Comparison with edit-friendly DDPM Inversion with SDXL Turbo. We invert two images with the prompts: ``a cat laying in a bed made out of wood'' (top) and ``a dog sitting on the beach with its tongue out'' (bottom) and apply two edits to each image.
    \ifeccv
    \else
        \vspace{-16pt}
    \fi
    }
    \label{fig:sup_turbo_compare_ddpm}
    \vspace{-0.15cm}
\end{figure}

%% file: figures/ddim_with_renoise_stability.tex
\begin{figure*}
    \centering
    \setlength{\tabcolsep}{1pt}
    \addtolength{\belowcaptionskip}{-10pt}
    {\small
    \centering
    
    \begin{tabular}{ccc ccc}
    \ifeccv
        {\begin{tabular}{c} Original \\ Image  \end{tabular}} &
        {\begin{tabular}{c} DDIM \\ Inversion  \end{tabular}} &
        {\begin{tabular}{c} {+1 ReNoise} \\ Step  \end{tabular}} &
        {\begin{tabular}{c} Original \\ Image  \end{tabular}} &
        {\begin{tabular}{c} DDIM \\ Inversion  \end{tabular}} &
        {\begin{tabular}{c} {+1 ReNoise} \\ Step  \end{tabular}} \\
    \else
        {Original Image} &
        {DDIM  Inversion} &
        {+1 ReNoise Step} &
        {Original Image } &
        {DDIM Inversion } &
        {+1 ReNoise Step } \\
    \fi
            
        \includegraphics[width=0.15\columnwidth]{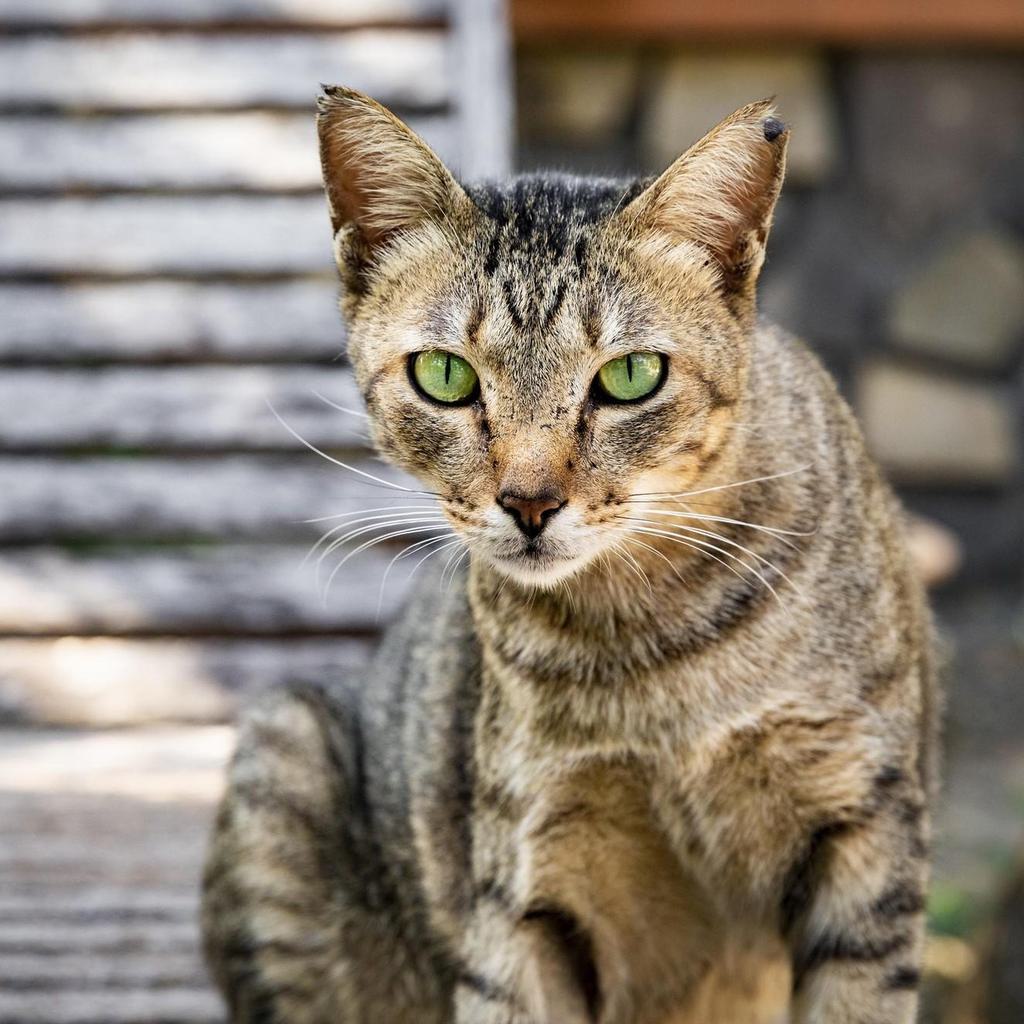} &
        \includegraphics[width=0.15\columnwidth]{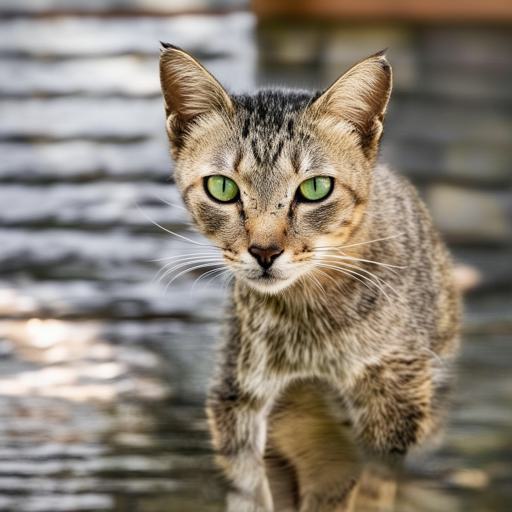} &
        \includegraphics[width=0.15\columnwidth]{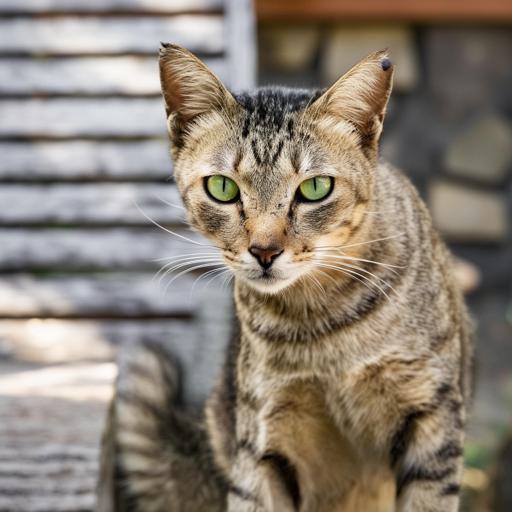} & 
        
        \includegraphics[width=0.15\columnwidth]{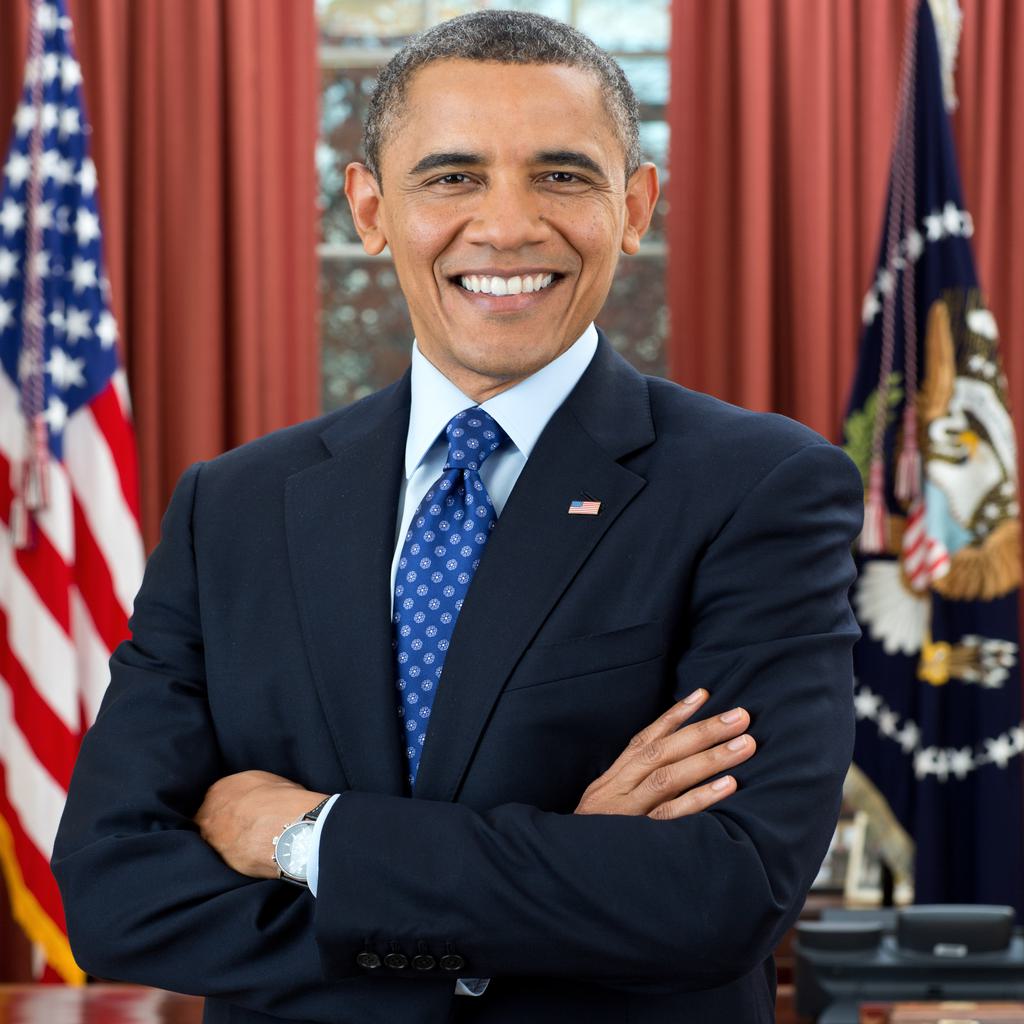} &
        \includegraphics[width=0.15\columnwidth]{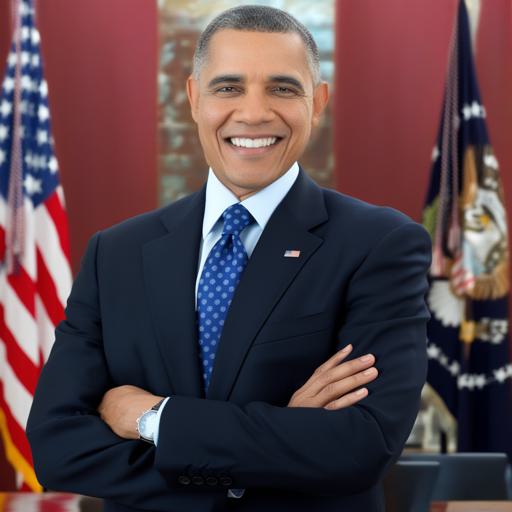} &
        \includegraphics[width=0.15\columnwidth]{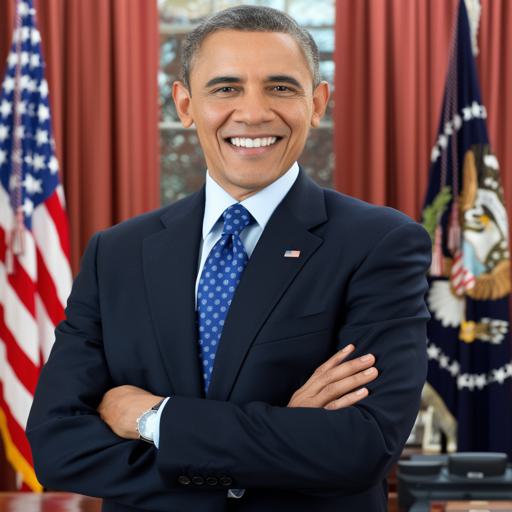}  \\
            
        \includegraphics[width=0.15\columnwidth]{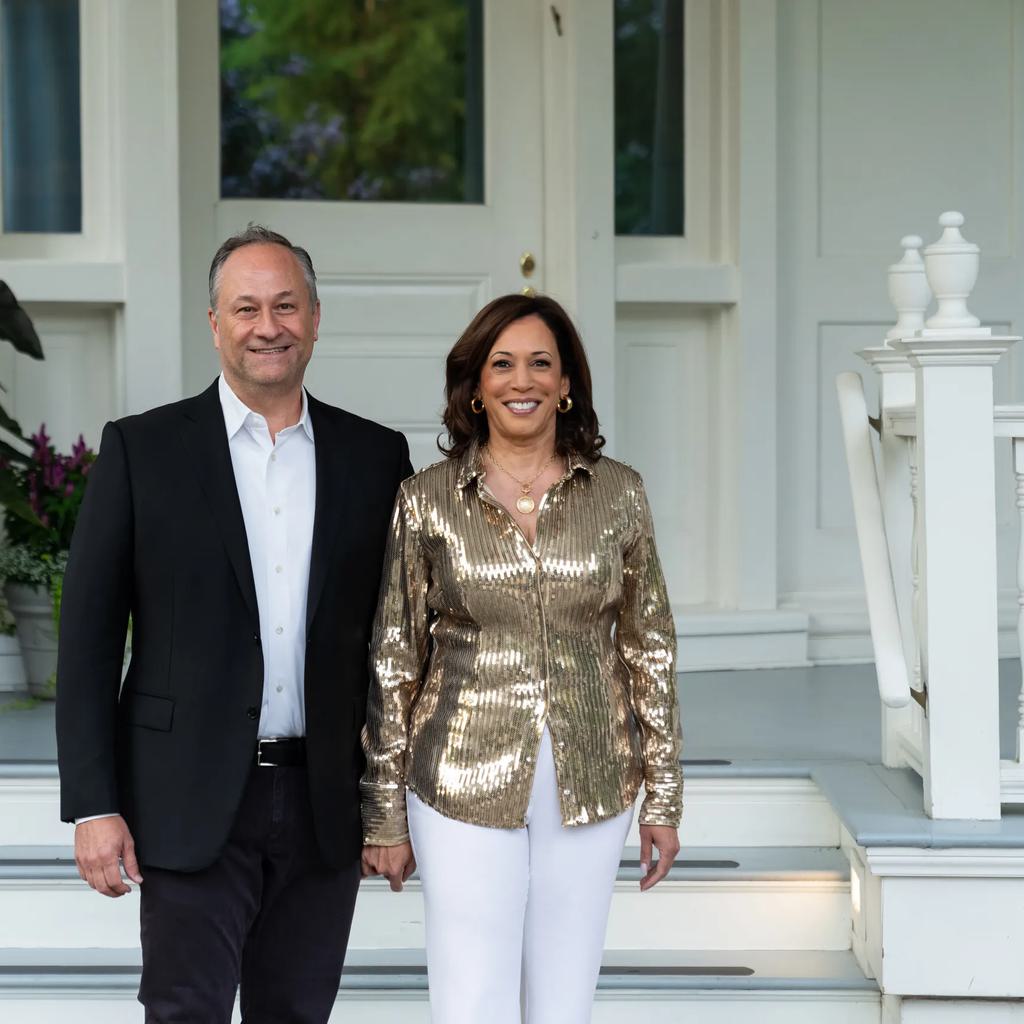} &
        \includegraphics[width=0.15\columnwidth]{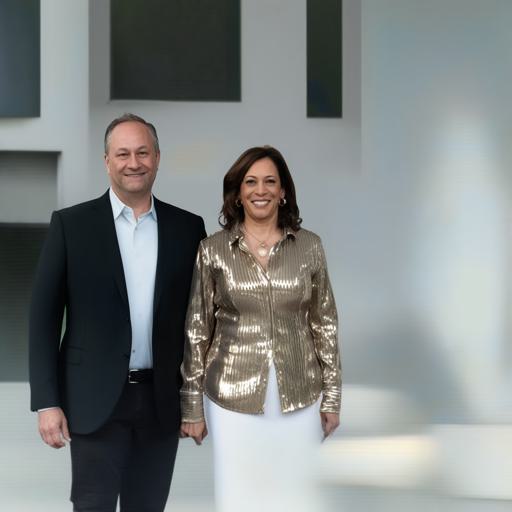} &
        \includegraphics[width=0.15\columnwidth]{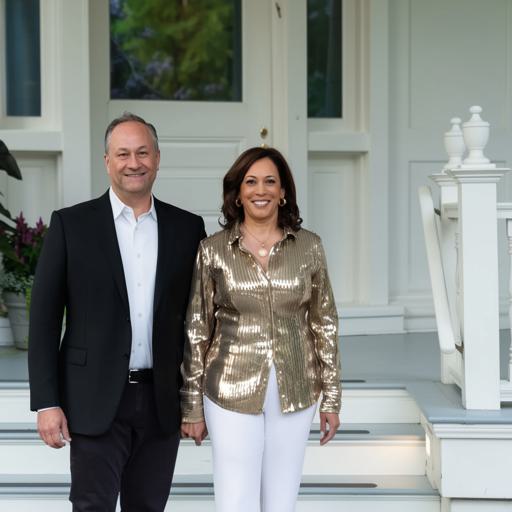} & 
        
        \includegraphics[width=0.15\columnwidth]{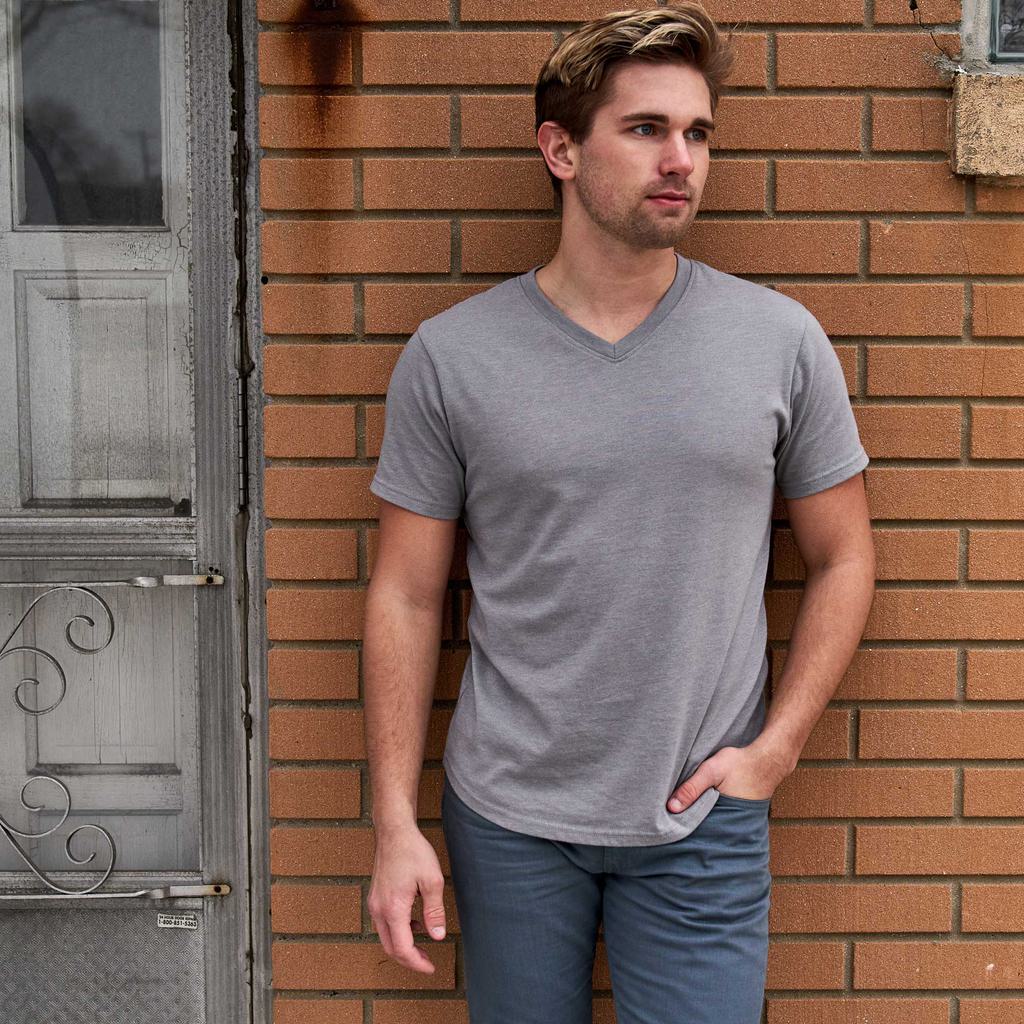} &
        \includegraphics[width=0.15\columnwidth]{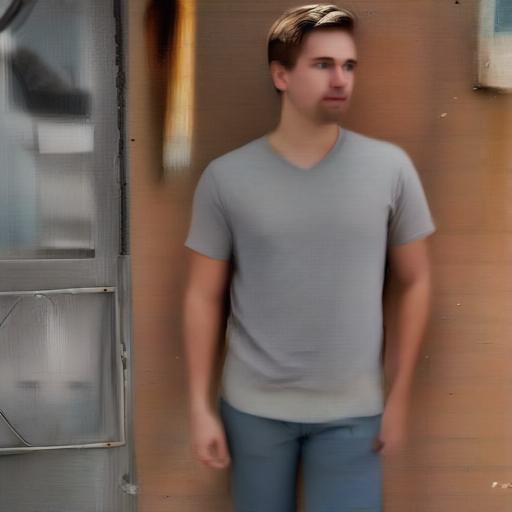} &
        \includegraphics[width=0.15\columnwidth]{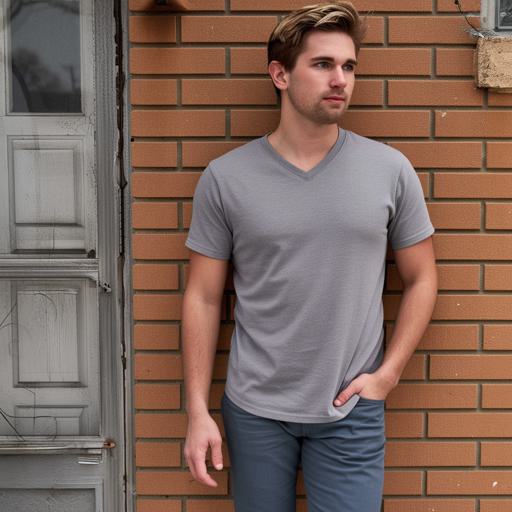}  \\
            
        \includegraphics[width=0.15\columnwidth]{images/dataset-2/11-cat.jpg} &
        \includegraphics[width=0.15\columnwidth]{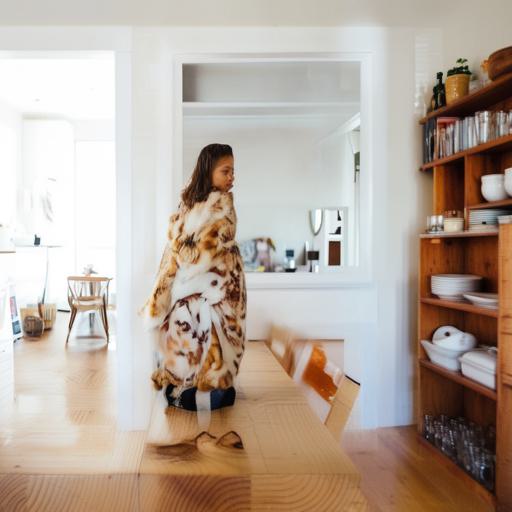} &
        \includegraphics[width=0.15\columnwidth]{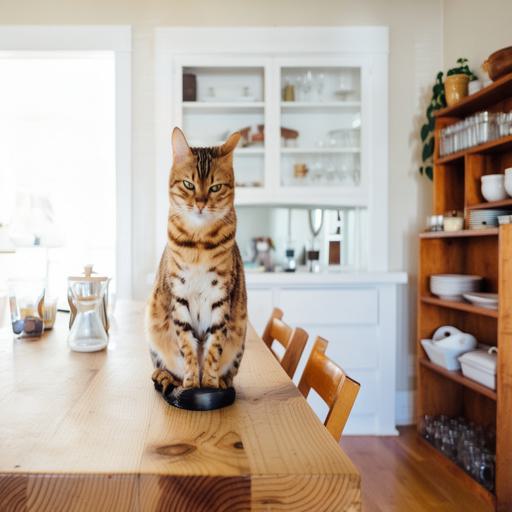} & 
        
        \includegraphics[width=0.15\columnwidth]{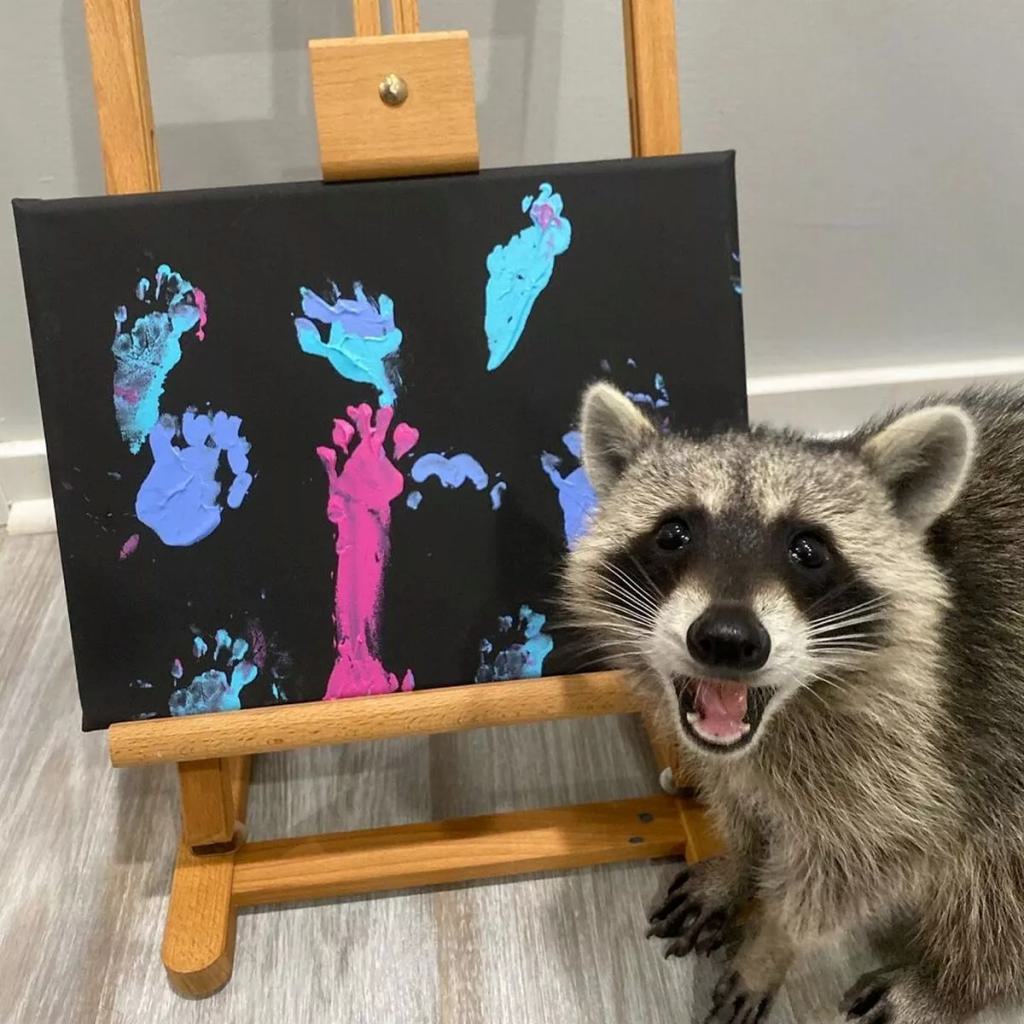} &
        \includegraphics[width=0.15\columnwidth]{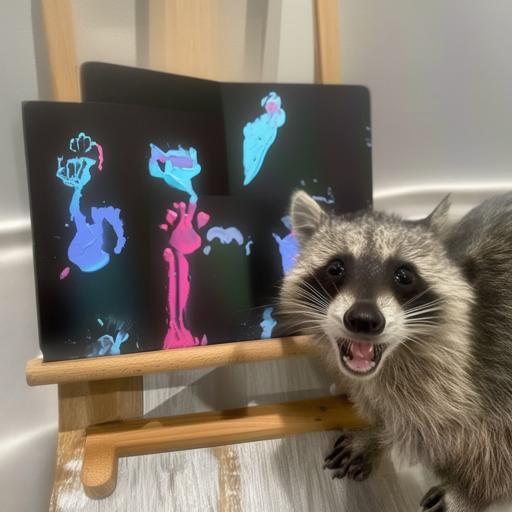} &
        \includegraphics[width=0.15\columnwidth]{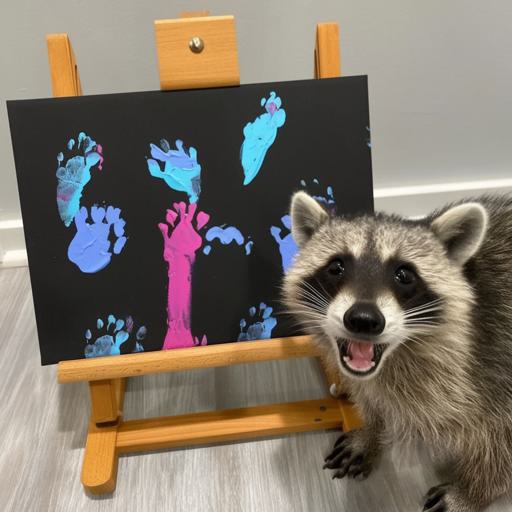}  \\
            
        \includegraphics[width=0.15\columnwidth]{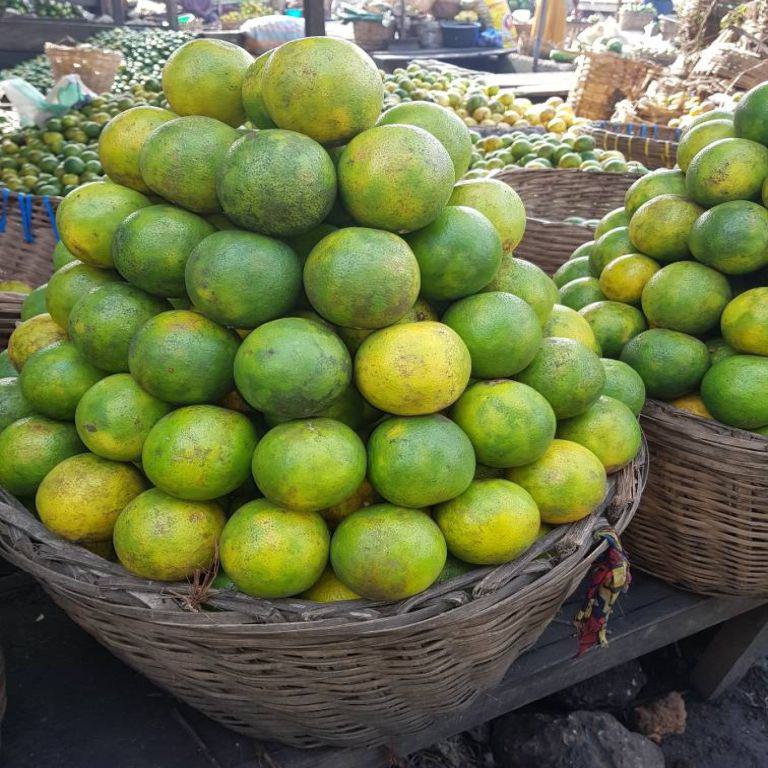} &
        \includegraphics[width=0.15\columnwidth]{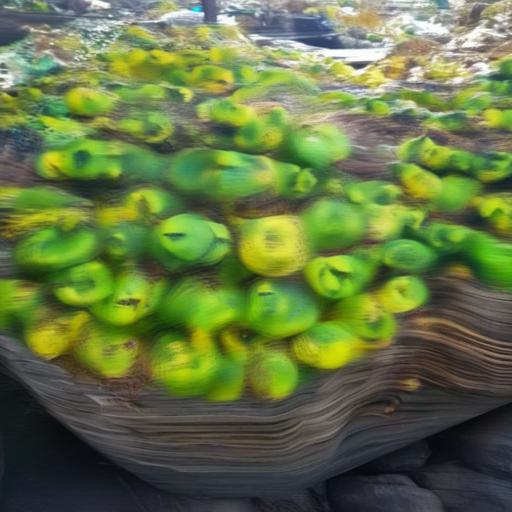} &
        \includegraphics[width=0.15\columnwidth]{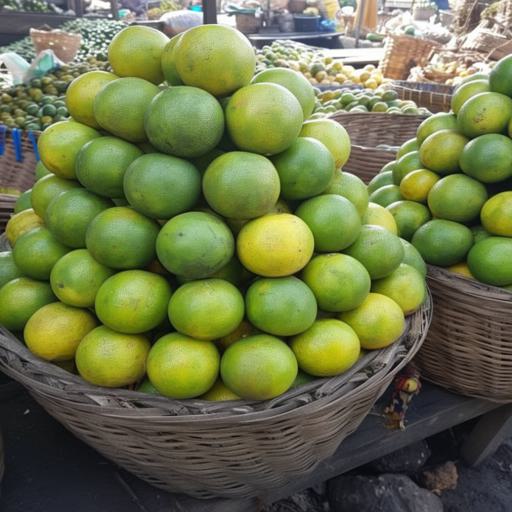} & 
        
        \includegraphics[width=0.15\columnwidth]{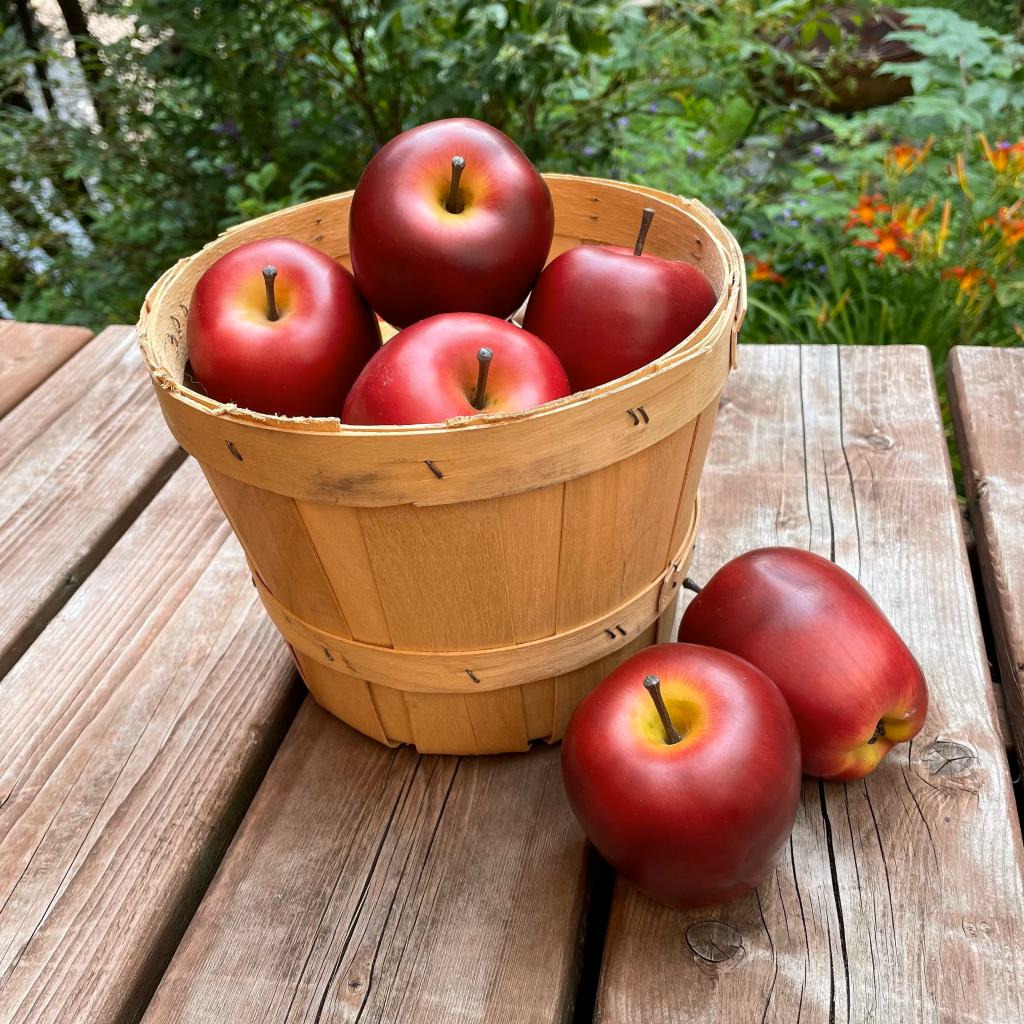} &
        \includegraphics[width=0.15\columnwidth]{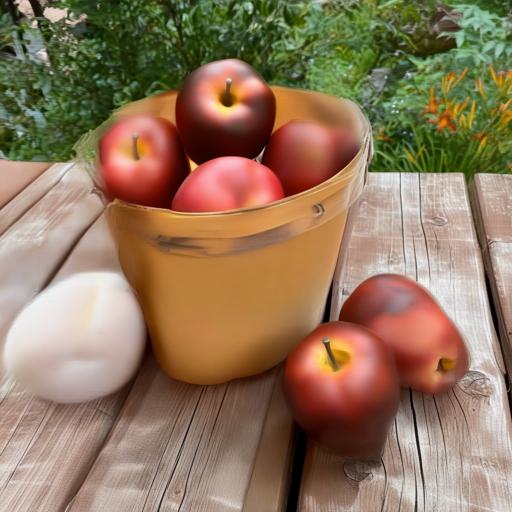} &
        \includegraphics[width=0.15\columnwidth]{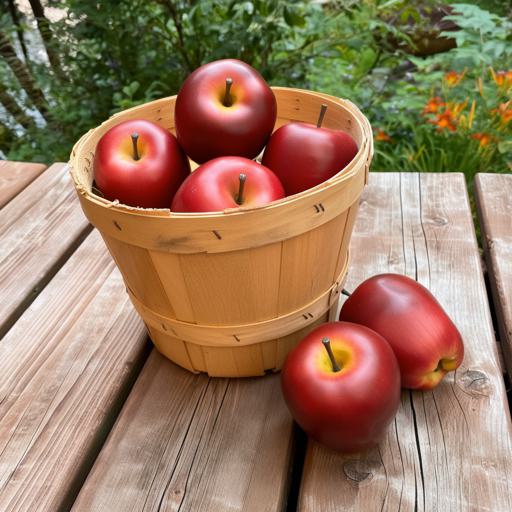}  \\
            
        \includegraphics[width=0.15\columnwidth]{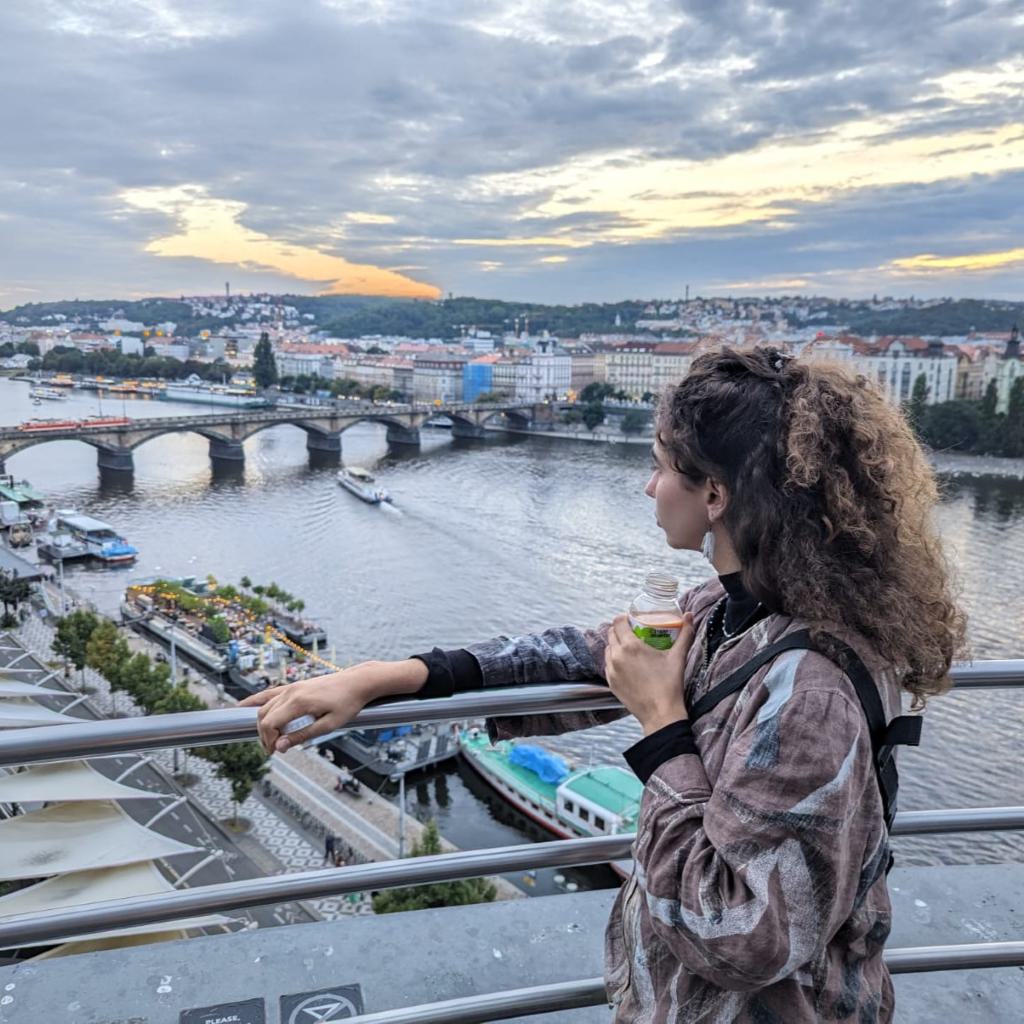} &
        \includegraphics[width=0.15\columnwidth]{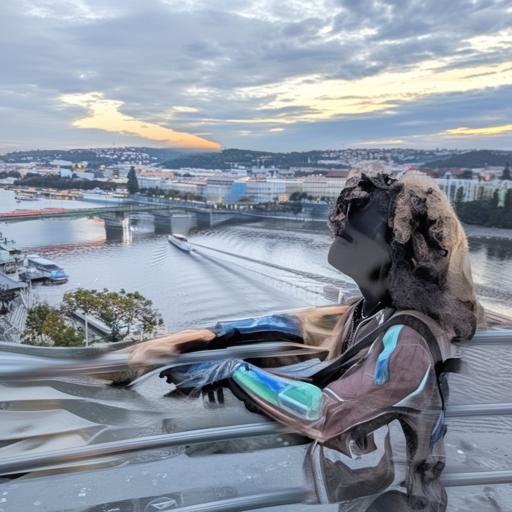} &
        \includegraphics[width=0.15\columnwidth]{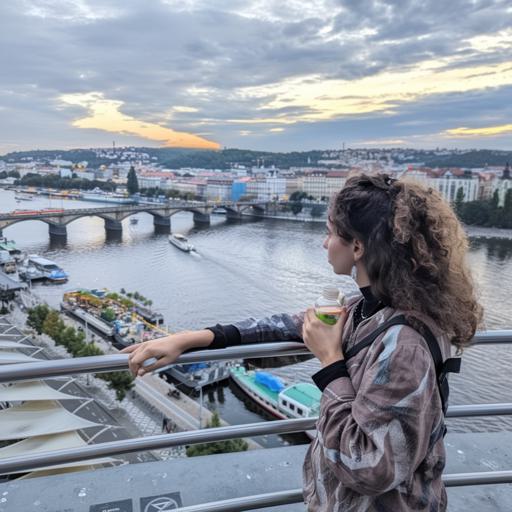} & 
        
        \includegraphics[width=0.15\columnwidth]{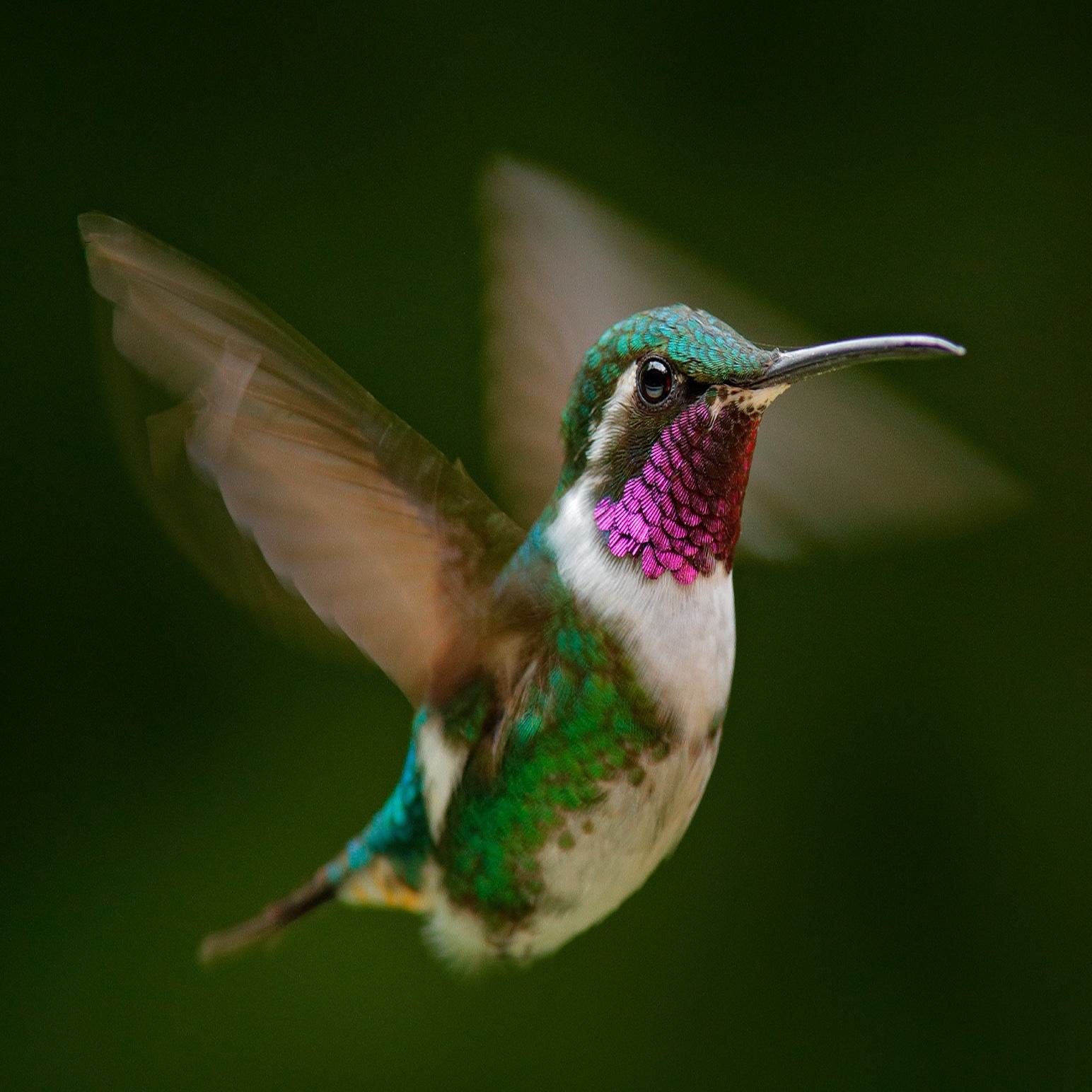} &
        \includegraphics[width=0.15\columnwidth]{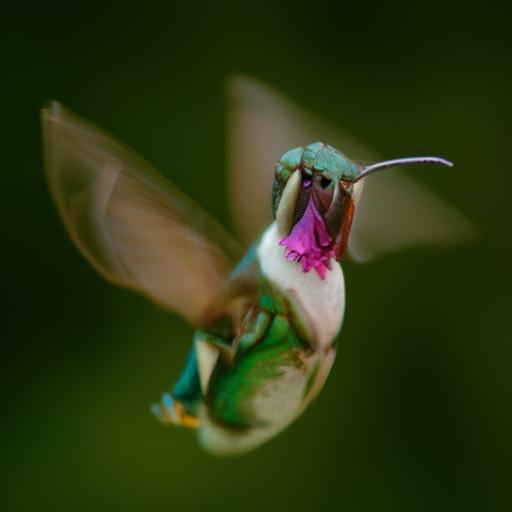} &
        \includegraphics[width=0.15\columnwidth]{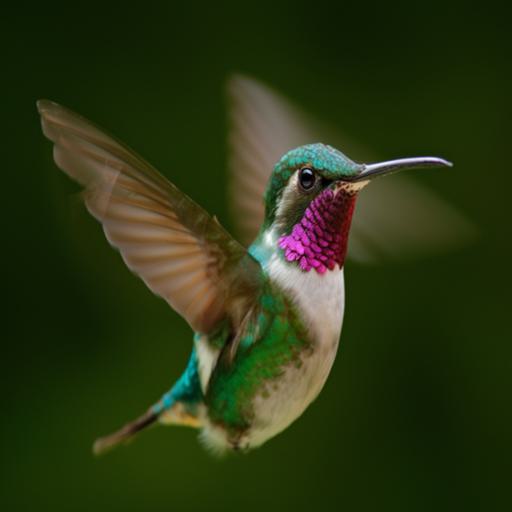}  \\
            
        \includegraphics[width=0.15\columnwidth]{images/dataset/bird_3.jpg} &
        \includegraphics[width=0.15\columnwidth]{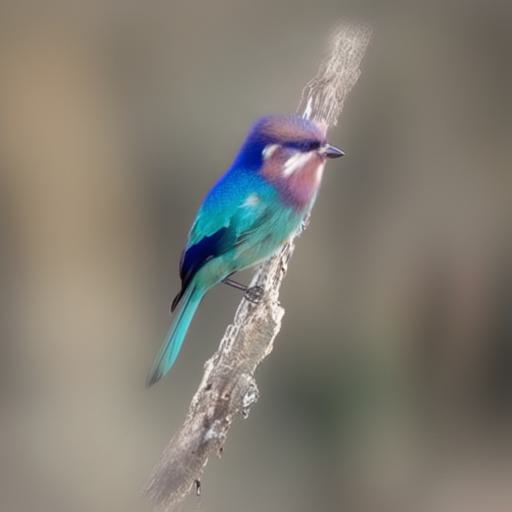} &
        \includegraphics[width=0.15\columnwidth]{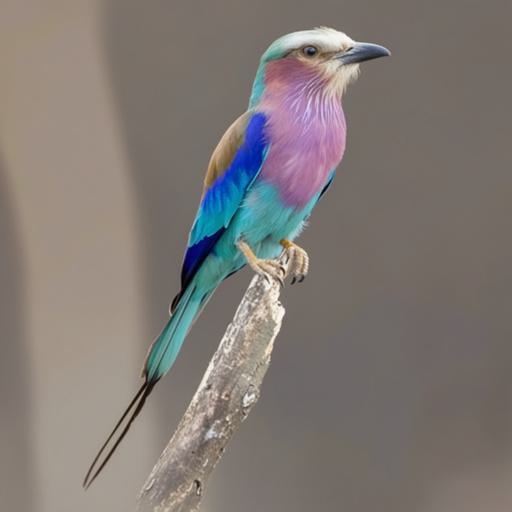} & 
        
        \includegraphics[width=0.15\columnwidth]{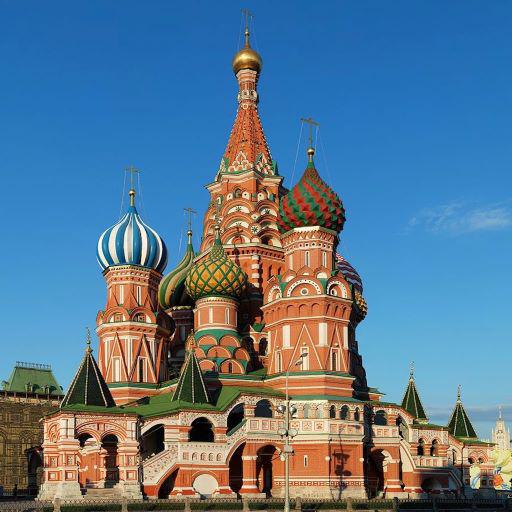} &
        \includegraphics[width=0.15\columnwidth]{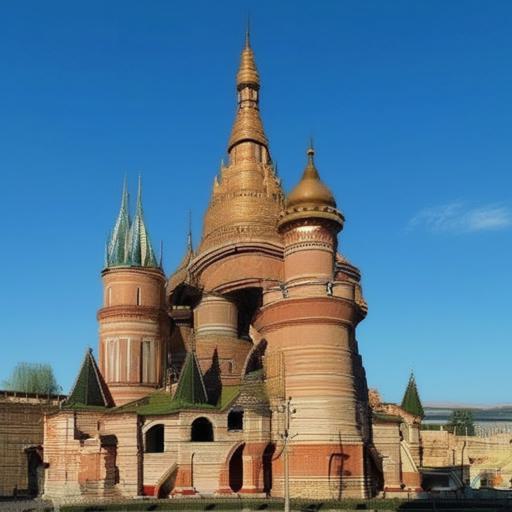} &
        \includegraphics[width=0.15\columnwidth]{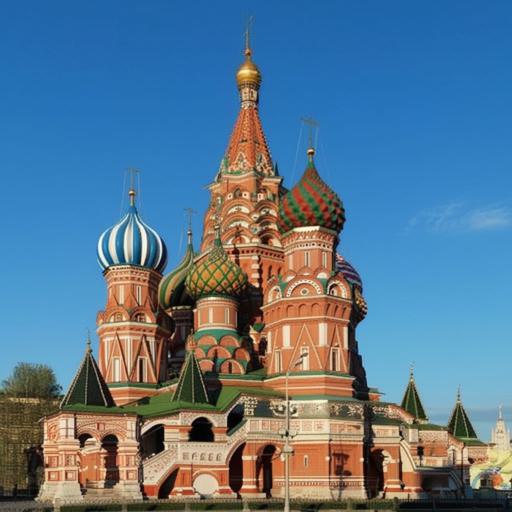}  \\
            
        \includegraphics[width=0.15\columnwidth]{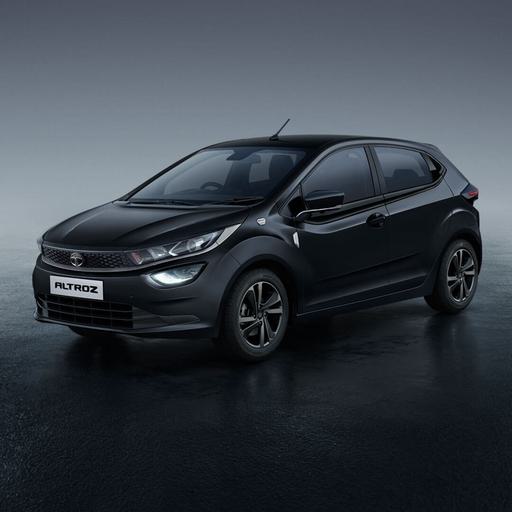} &
        \includegraphics[width=0.15\columnwidth]{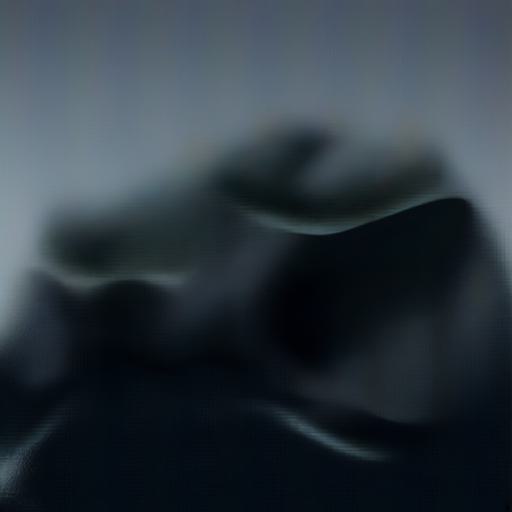} &
        \includegraphics[width=0.15\columnwidth]{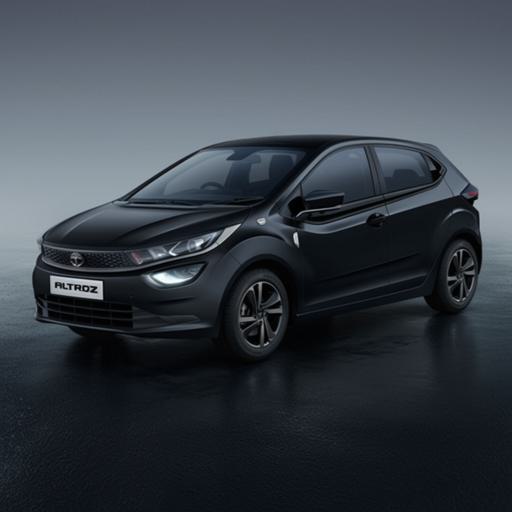} & 

        \ifeccv
        \includegraphics[width=0.15\columnwidth]{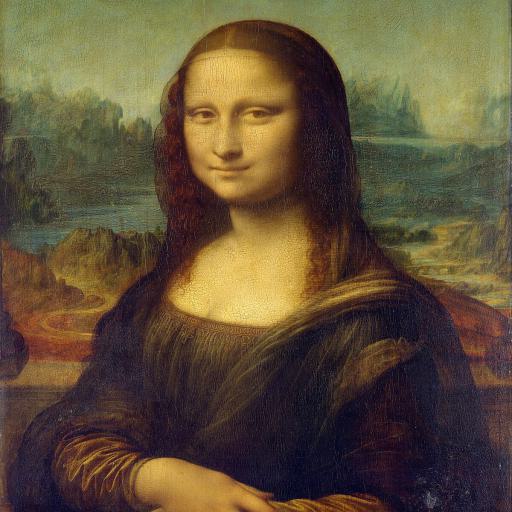} &
        \includegraphics[width=0.15\columnwidth]{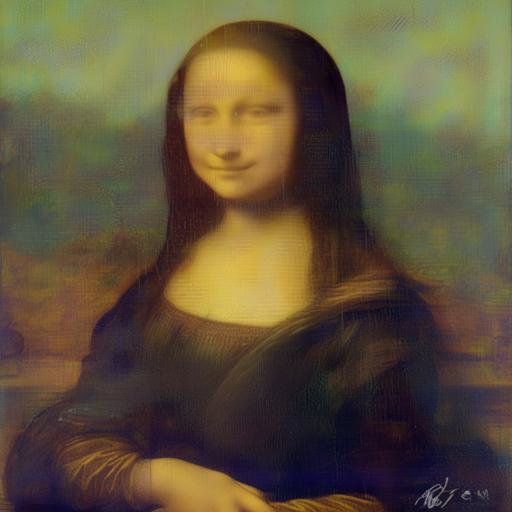} &
        \includegraphics[width=0.15\columnwidth]{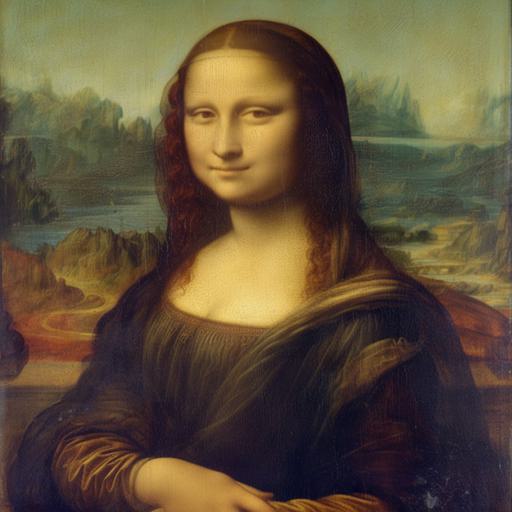}  \\
            
        \includegraphics[width=0.15\columnwidth]{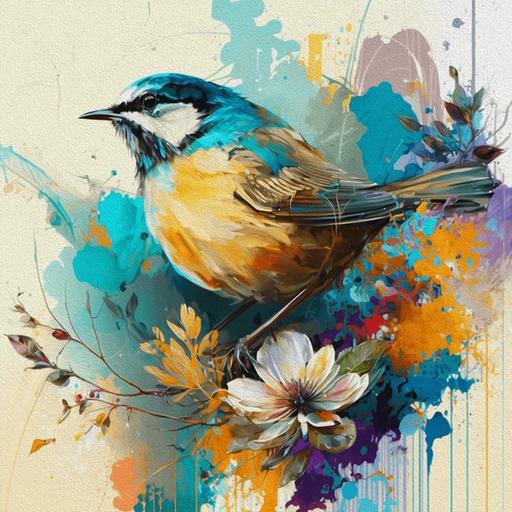} &
        \includegraphics[width=0.15\columnwidth]{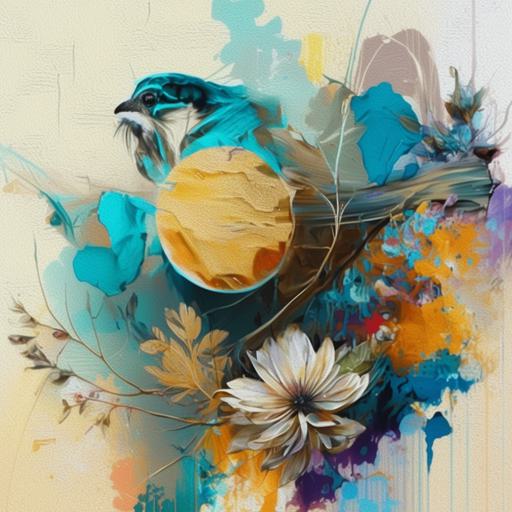} &
        \includegraphics[width=0.15\columnwidth]{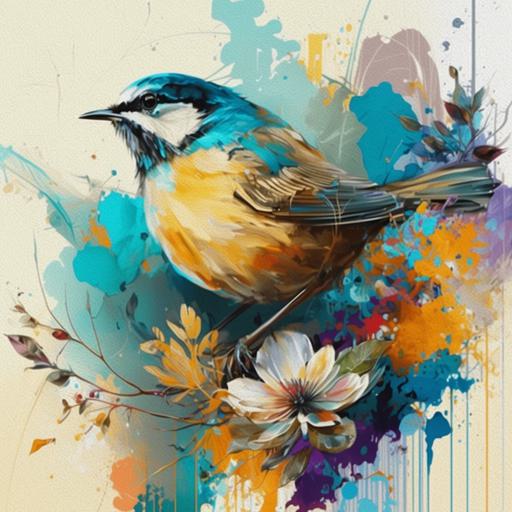} & 
        \fi
        
        \includegraphics[width=0.15\columnwidth]{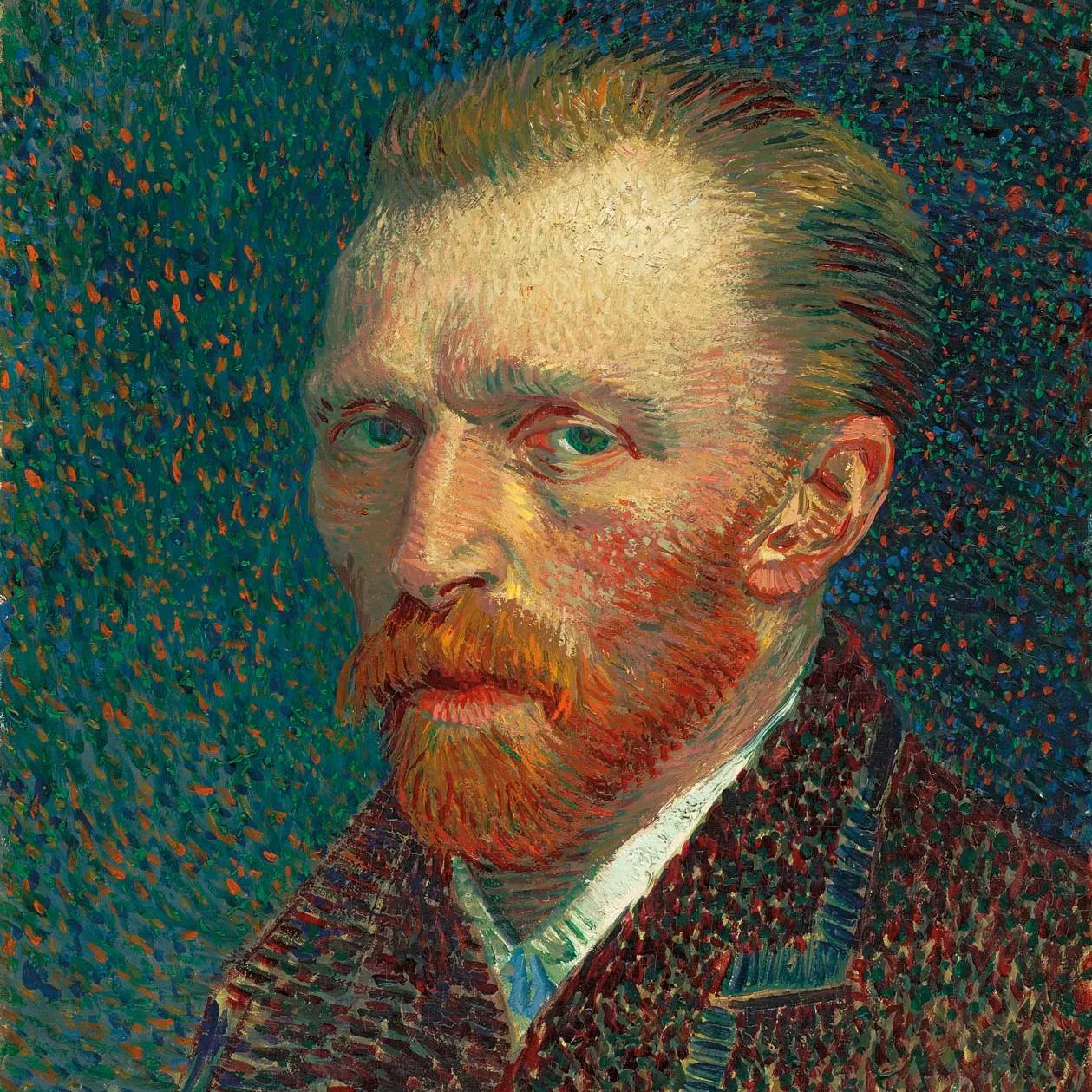} &
        \includegraphics[width=0.15\columnwidth]{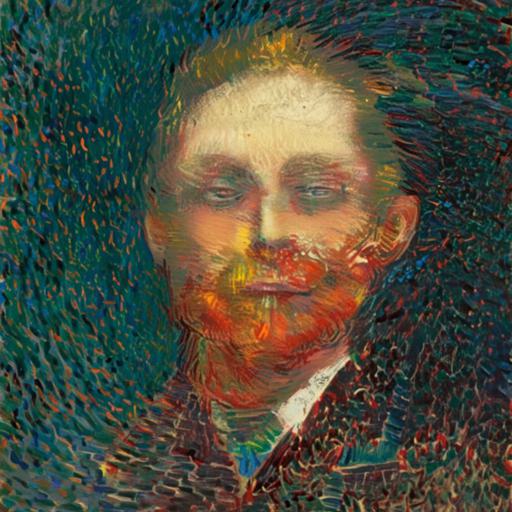} &
        \includegraphics[width=0.15\columnwidth]{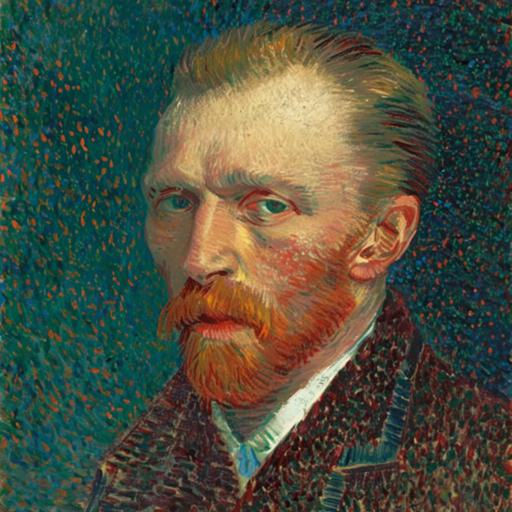}  \\
            
        \includegraphics[width=0.15\columnwidth]{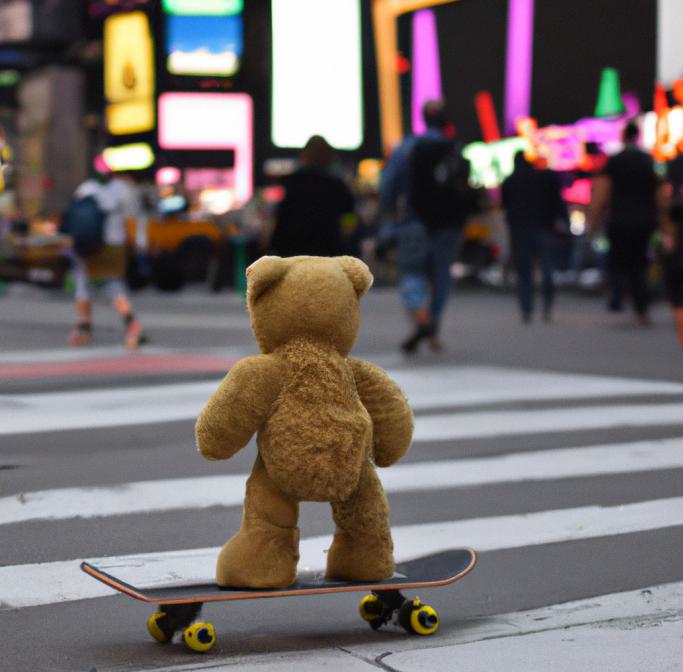} &
        \includegraphics[width=0.15\columnwidth]{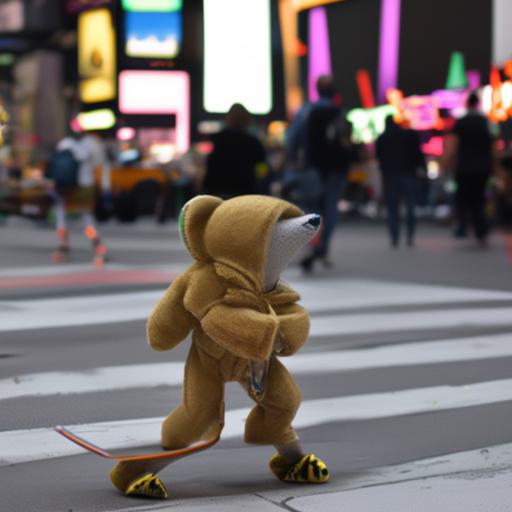} &
        \includegraphics[width=0.15\columnwidth]{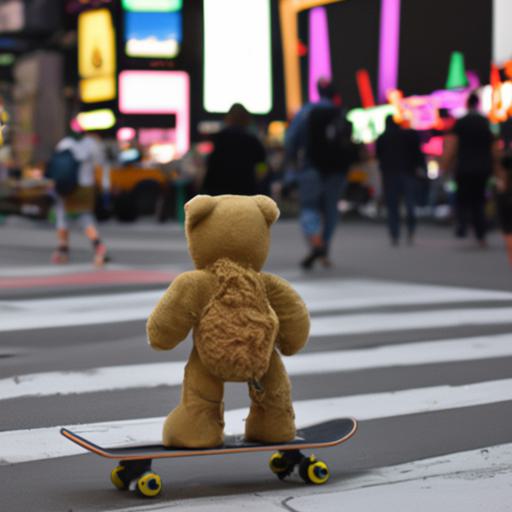} & 
        
        \includegraphics[width=0.15\columnwidth]{images/dataset/person_2.jpg} &
        \includegraphics[width=0.15\columnwidth]{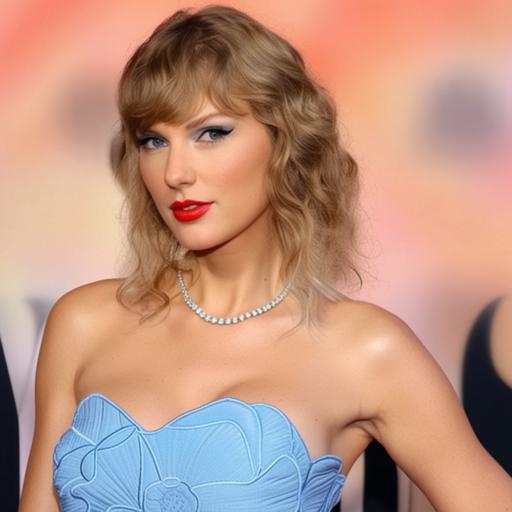} &
        \includegraphics[width=0.15\columnwidth]{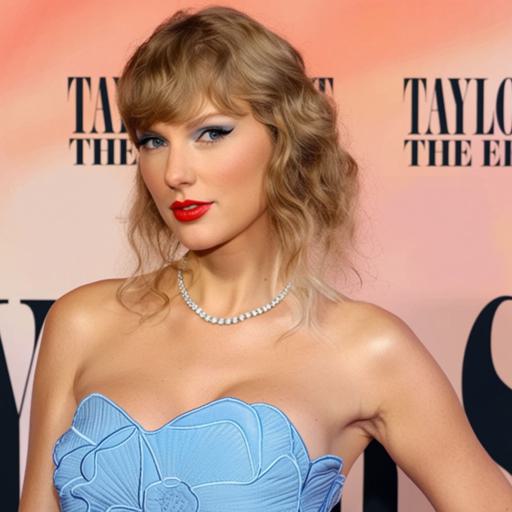}  \\
        
    \end{tabular}
    
    }
    \vspace{-0.325cm}
    \caption{
        Comparing reconstruction results with an empty prompt of plain DDIM inversion on SDXL to DDIM inversion with one ReNoise iteration.
    }
    \label{fig:ddim_with_renoise_stability}

\end{figure*}